%% file: emnlp2023.tex
\pdfoutput=1

\documentclass[11pt]{article}

\usepackage{EMNLP2023}

\usepackage{times}
\usepackage{latexsym}
\usepackage{amsmath}
\usepackage{amssymb}
\usepackage{booktabs}
\usepackage{multicol, multirow}
\usepackage[T1]{fontenc}

\usepackage[utf8]{inputenc}

\usepackage{microtype}

\usepackage{inconsolata}
\usepackage{mdframed}
\usepackage{xcolor}
\usepackage{url}
\usepackage{graphicx}
\usepackage{caption}
\usepackage{subcaption}
\usepackage{tcolorbox}
\usepackage{soul}

\usepackage[shortlabels]{enumitem}

\newtcolorbox{highlight}{
  colback=lightgray!20, 
  colframe=gray!60, 
  boxrule=1pt, 
  arc=0pt, 
  boxsep=0pt, 
  left=6pt, 
  right=6pt, 
  top=6pt, 
  bottom=6pt 
}
\newtcolorbox{example}{
  colback=blue!20, 
  colframe=blue!60, 
  boxrule=1pt, 
  arc=0pt, 
  boxsep=0pt, 
  left=6pt, 
  right=6pt, 
  top=6pt, 
  bottom=6pt 
}

\newcommand{\ourbenchmark}{\textsc{CRoW}}

\newif\ifcomments
\commentstrue
\ifcomments
\usepackage[normalem]{ulem}
\definecolor{ABpurple}{rgb}{0.8,0.0,0.8}
\newcommand\ab[1]{\textcolor{ABpurple}{\textsf{\scriptsize[\textbf{AB\@:} #1]}}} 
\newcommand\abi[1]{\textcolor{ABpurple}{#1}} 
\newcommand\abm[1]{\marginpar{\raggedright\tiny\textcolor{ABpurple}{\textsf{{\bfseries AB\@:} #1}}}} 
\newcommand\abs{\bgroup\markoverwith{\textcolor{ABpurple}{\rule[.4ex]{2pt}{0.8pt}}}\ULon} 
\else
\newcommand\ab[1]{}
\newcommand\abi[1]{\ignorespaces}
\newcommand\abm[1]{}
\newcommand\abs[1]{#1}
\fi

\newcommand\crow{\raisebox{-8pt}{\includegraphics[width=1.8em]{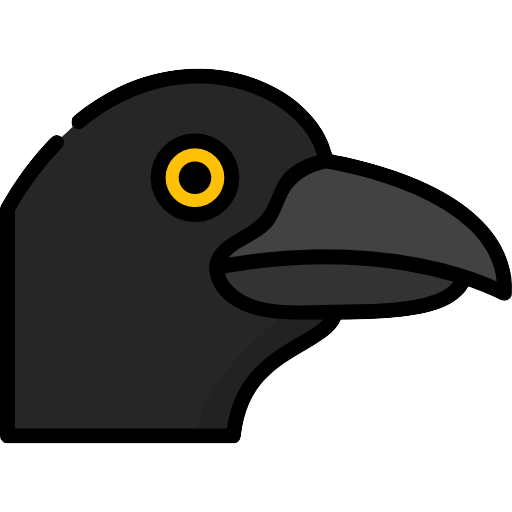}}}

\newcommand\ie{\textit{i.e.}}
\newcommand\eg{\textit{e.g.}}

\title{\crow ~\ourbenchmark: Benchmarking Commonsense Reasoning in Real-World Tasks}
\author{Mete Ismayilzada~~~~ Debjit Paul$^*$~~~~  Syrielle Montariol$^*$~~~~ \textbf{Mor Geva$^\Diamond$}~~~~ Antoine Bosselut  \\
  EPFL, Switzerland \\ 
  $^\Diamond$ Google DeepMind\\ 
  \texttt{mahammad.ismayilzada@epfl.ch}}

\begin{document}
\maketitle

\def\thefootnote{*}\footnotetext{Equal contribution}
\def\thefootnote{\arabic{footnote}}

\begin{abstract}
\input{00_abstract}
\end{abstract}

\section{Introduction}
\input{01_introduction}
\label{sec:introduction}

\section{Related Work}
\input{02_related_work}
\label{sec:related-work}

\input{03_benchmark}
\label{sec:benchmark}

\section{Experimental Setup}

\input{04_experiments}
\label{sec:experiments}

\section{Results}
\input{05_results}
\label{sec:results}

\section{Analysis}

\input{06_analysis}
\label{sec:analysis}

\section{Conclusion}
\input{07_conclusion}
\label{sec:conclusion}

\section*{Limitations}
\input{08_limitations}
\label{sec:limitations}


\section*{Acknowledgements}
\input{10_acknowledgements}
\label{sec:acknowledgements}

\bibliography{anthology,references}
\bibliographystyle{acl_natbib}

\clearpage
\appendix

\input{11_appendix}

\end{document}

%% file: 00_abstract.tex
Recent efforts in natural language processing (NLP) commonsense reasoning research have yielded a considerable number of new datasets and benchmarks. However, most of these datasets formulate commonsense reasoning challenges in artificial scenarios that are not reflective of the tasks which real-world NLP systems are designed to solve. 
In this work, we present \ourbenchmark, a manually-curated, 
multi-task benchmark that evaluates
the ability of models to apply commonsense reasoning in the context of six real-world NLP tasks. \ourbenchmark{} is constructed using a multi-stage data collection pipeline that rewrites examples from existing datasets using commonsense-violating perturbations. We use \ourbenchmark to study how NLP systems perform across different dimensions of commonsense knowledge, such as physical, temporal, and social reasoning. We find a significant performance gap when NLP systems are evaluated on \ourbenchmark compared to humans, showcasing that commonsense reasoning is far from being solved in real-world task settings. We make our dataset and leaderboard available to the research community.\footnote{\url{https://github.com/mismayil/crow}}

%% file: 01_introduction.tex


Commonsense reasoning is a long-standing challenge in artificial intelligence (AI) and natural language processing \cite{McCarthy1960ProgramsWC, Winograd1974UnderstandingNL,Davis2015CR,Choi2022TheCC}, resulting in a large number of datasets and benchmarks designed to evaluate how AI systems reason in commonsense scenarios described in natural language \cite{davis2023benchmarks}.  Recently, large language models, such as GPT-3 \cite{NEURIPS2020_gpt3} and PaLM \cite{chowdhery2022palm}, have demonstrated near-human performance on many of these benchmarks \cite{Lourie2021UNICORNOR}.
However, these models can still be 
brittle in practical deployments, raising questions about how reliably these commonsense benchmarks truly evaluate the commonsense reasoning abilities of models. 

\begin{figure}[t]
\includegraphics[scale=0.6]{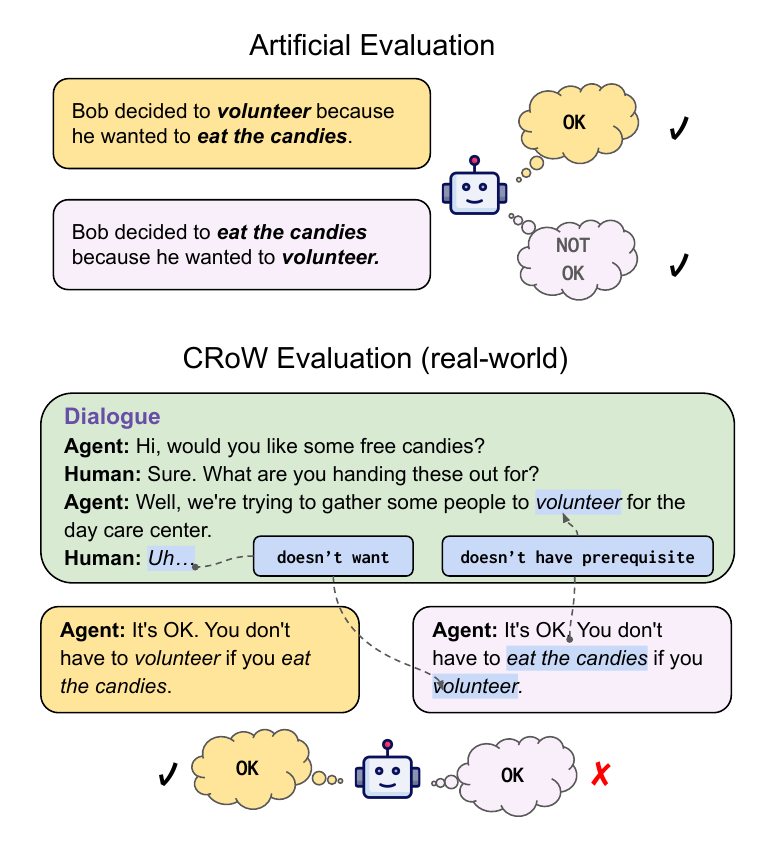}
\centering
\caption{An example from one of the tasks (Dialogue) in our benchmark showcasing the difference between the evaluation of commonsense reasoning in an artificial and real-world setting. \ourbenchmark grounds this evaluation in a real-world context that often requires the use of rich and \textit{implicit} commonsense knowledge to solve a task.}
\label{fig:motivation-example}
\end{figure}



Part of this issue stems from the practice that most commonsense datasets are designed to evaluate reasoning in artificial task settings that are not reflective of the real-world use cases in which NLP systems are deployed. In real-world settings, one almost never directly observes a test of commonsense knowledge in isolation. In this paper, we argue instead that \textit{commonsense reasoning benchmarks should evaluate commonsense reasoning in the tasks in which these abilities are required.} 


 
The necessity of commonsense to solve real-world tasks has been extensively argued since the early stages of AI, notably by \citet{BarHillel1960ADO} in the context of machine translation. However, despite these early arguments, only recently was there an attempt to construct a commonsense reasoning dataset for machine translation \cite{he-etal-2020-box}, an effort which concluded that the commonsense reasoning abilities of modern models were still in their infancy when applied in real NLP tasks.


In this work, we build on these original ideas and introduce 
\textbf{\ourbenchmark}: a \textbf{C}ommonsense \textbf{R}eas\textbf{o}ning Benchmark for Real-\textbf{W}orld Tasks, a multi-task benchmark containing high-quality datasets for six real-world NLP tasks: machine translation (MT), open-domain dialogue (DG), dialogue summarization (DS), intent detection (ID), stance classification (SC), and safety detection (SD). Inspired by Winograd schemas \cite{Levesque2011TheWS}, we build our benchmark by applying commonsense-based minimal perturbations on examples from existing datasets for each task. For each of these tasks, we crowdsource collections of potential target references for the task, each grounded to a particular commonsense violation with respect to the original context (see Figure~\ref{fig:motivation-example} for examples in dialogue response generation). We categorize these commonsense violations across six dimensions --- temporal, causal, attribution, comparison, physical, and social --- ensuring a diverse breakdown of commonsense reasoning types in \ourbenchmark.

Our empirical study across 13 state-of-the-art (SoTA) systems (including GPT-4) shows that \ourbenchmark{} is a challenging commonsense reasoning testbed, with the highest performing model scoring {$\sim$18\%} lower than humans on individual examples and {$\sim$37\%} lower on our more restrictive metric that evaluates situational robustness. 
Consequently, we provide \ourbenchmark{} to the community as the first commonsense benchmark specifically formed to test commonsense knowledge and reasoning abilities in the same contexts as real-world deployments of NLP systems.
\noindent The contributions of our work can be summarized as follows:

\begin{itemize}
[leftmargin=*,topsep=1pt,parsep=1pt]
\item We design a common multi-stage data collection pipeline for generating commonsense-based Winograd-style variations of examples, which can be applied to many tasks. 

\item  We apply our data collection pipeline to construct \ourbenchmark, a multi-task benchmark that evaluates the commonsense reasoning ability of models in solving six diverse real-world NLP tasks.

\item For each task, we evaluate and analyze the performance of state-of-the-art 
    models on our benchmark across different dimensions of commonsense knowledge.
\end{itemize}

%% file: 02_related_work.tex

\paragraph{Commonsense Reasoning Benchmarks}
Many benchmarks measuring the commonsense reasoning abilities of state-of-the-art models have been released in recent years. Starting with the well-known Winograd Schema Challenge (WSC; \citealp{Levesque2011TheWS}), these benchmarks have attempted to test the commonsense reasoning ability of models using different task formats, such as pronoun resolution \cite{Levesque2011TheWS, Rudinger2018GenderBI, eisenschlos-etal-2023-winodict}, question-answering \cite{talmor-etal-2019-commonsenseqa, Zellers2018SWAGAL, chen-etal-2019-codah, reddy-etal-2019-coqa, zellers-etal-2019-hellaswag}, plausible inference \cite{copaGordon2011, Bhagavatula2019AbductiveCR, wang-etal-2019-make, singh-etal-2021-com2sense,gao-etal-2022-comfact} and natural language generation \cite{lin-etal-2020-commongen}. Benchmarks have also been created to evaluate commonsense reasoning across different dimensions of commonsense knowledge, including social \cite{rashkin-etal-2018-modeling,rashkin-etal-2018-event2mind,sap-etal-2019-social}, physical  \cite{Bisk2019PIQARA, dalvi-etal-2018-tracking, storks-etal-2021-tiered-reasoning}, temporal  \cite{qin-etal-2021-timedial, zhou-etal-2019-going} and numerical reasoning \cite{lin-etal-2020-birds}. Additionally, there exist comprehensive multi-task benchmarks that consist of several new or existing datasets for commonsense reasoning \cite{tamari-etal-2022-dyna, srivastava2022imitation, Wang2019SuperGLUEAS}. For a thorough survey in this area, we refer readers to \cite{Storks2019RecentAI, davis2023benchmarks}. 
In contrast to these benchmarks, where the underlying task formulation is centered around a task that is typically not grounded in a real-world setting, we construct \ourbenchmark{} to specifically focus on evaluating commonsense reasoning in real-world tasks for which NLP systems would be deployed.

\paragraph{Commonsense Reasoning in Real-World Contexts} 
A few recent works have explored the role of commonsense knowledge in real-world settings, such as open-ended response generation \cite{zhou-etal-2021-commonsense, ghosal-etal-2021-cider, ghosal-etal-2022-cicero}, machine translation \cite{he-etal-2020-box} and reading comprehension \cite{Zhang2018ReCoRDBT, huang-etal-2019-cosmos} and have proposed new commonsense reasoning tasks and benchmarks. We build on top of these benchmarks and extend them to several other real-world NLP tasks, along with a general data collection methodology for commonsense knowledge annotation and Winograd-style schema generation that can be applied to other tasks in the future.


%% file: 03_benchmark.tex
\begin{figure*}
\includegraphics[width=0.9\textwidth]{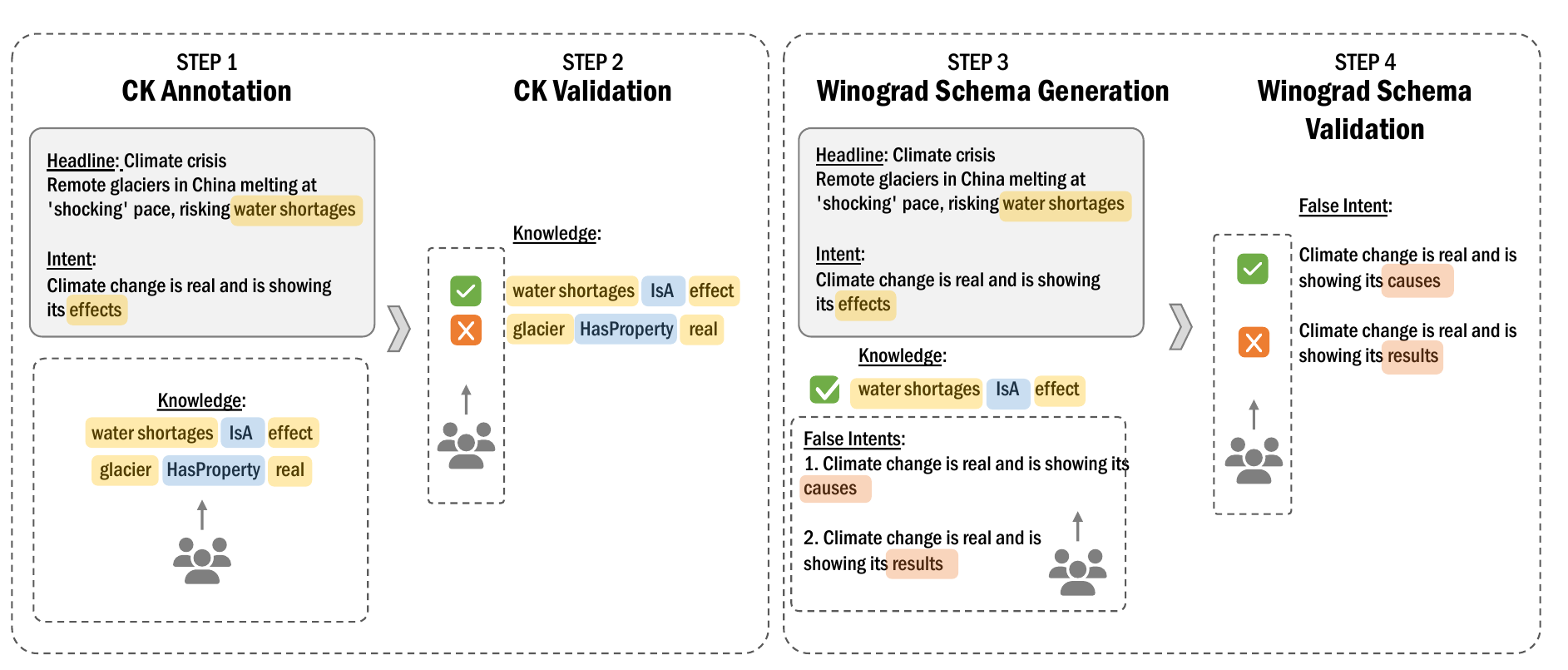}
\centering
  \vspace{-2mm}
\caption{\textbf{CRoW Data Collection Pipeline} (as illustrated for the Intent Detection Task). Given a \textit{context} (news headline) and a \textit{target} (writer's intent behind it), in the first phase of the pipeline, annotators are asked to identify commonsense knowledge about this context. In the second phase, annotators use the commonsense knowledge from the previous phase to minimally perturb the \textit{target} to generate a Winograd-style schema for the given example. Each annotation stage is also followed with its own validation step.}
\label{fig:pipeline-example}
\end{figure*}
\section{Data Collection}
\label{sec:data_collection}

Our goal is to 
assess the ability of NLP systems to apply commonsense reasoning in real-world tasks. 
To this end, we define a general methodology and multi-stage data collection pipeline (Figure \ref{fig:pipeline-example}) for 
generating evaluation examples that require commonsense reasoning in a given real-world task.
In what follows, we outline our general data collection methodology, 
and describe each step in detail. 

\subsection{Overview}
The Winograd Schema Challenge (\citealp{Levesque2011TheWS}), an often-used benchmark to measure commonsense reasoning abilities, tests whether models can distinguish the meaning of pairs of sentences with commonsense-based minimal perturbations that flip their meaning. 
For example, given the sentence, \textit{``The trophy doesn’t fit into the brown suitcase because it’s too large,''} models should identify that the pronoun ``it'' refers to the ``trophy'' (using commonsense knowledge), but distinguish that replacing the word ``large'' by ``small'' would flip this reference to ``suitcase''. 
Winograd-style schemas have been widely adopted for tasks involving pronoun resolution \cite{Rudinger2018GenderBI, eisenschlos-etal-2023-winodict, Thrush2022WinogroundPV}, but also sense-making \cite{wang-etal-2019-make, singh-etal-2021-com2sense} and reasoning about exceptions \cite{do-pavlick-2021-rotten}. 
While these schemas are simple and effective for measuring commonsense robustness of models, they are rarely applied in real-world tasks.

Motivated by this gap, we construct \ourbenchmark, a benchmark of Winograd-style examples for real-world NLP tasks. While the inherent subtlety of commonsense-based minimal perturbations led the original Winograd schemas to be expert-crafted and limited in size, later works developed large-scale sets of Winograd schemas using crowdsourcing and adversarial filtering \cite{Sakaguchi2019WINOGRANDEAA}. In our work, we also employ crowdsourcing to generate Winograd-style perturbed examples, but our approach differs in one key aspect. Instead of asking crowdworkers to perturb the given sentences directly, we design a data collection pipeline that breaks down the schema construction into two independent stages: \textbf{Commonsense Knowledge Annotation (CKA)} and \textbf{Winograd-style Schema Generation (WSG)}, each of which is followed by a complementary validation stage. Figure \ref{fig:pipeline-example} illustrates the pipeline for the intent detection task. 

This multi-stage approach has two key benefits. First, we ground the perturbations to commonsense dimensions, ensuring the Winograd-style schemas differ on commonsense violations. Using these dimensions, we also ensure a diverse set of perturbations across different types of commonsense knowledge, allowing us to stratify our later analysis across these dimensions to more finely understand model failures in commonsense reasoning. Second, a particular stage can be skipped if the data for it is already available, which is the case for several tasks in our benchmark. We use Amazon Mechanical Turk (MTurk) as a crowdsourcing platform. Below, we describe each stage in detail.


\subsection{Methodology}
For a given task example, we define the \textit{context} as the unchanged part of the example and the \textit{target} as the candidate for commonsense-based minimal perturbation. {For example, in intent detection, we designate the headline as the \textit{context} and the intent as the \textit{target}. In Table \ref{tab:benchmark_stats_2} in the Appendix, we list respective mappings for all tasks.}

\paragraph{Commonsense Knowledge Annotation and Validation}
In the first stage of our pipeline, we explicitly annotate implicit commonsense knowledge underlying examples in real-world task datasets. 
In this stage, crowd workers are tasked to identify concepts in the \textit{context} and \textit{target} that could serve as the \textit{head} and \textit{tail} of an implicit commonsense relationship, as well as a pre-existing \textit{relation} that connects them. For example, in Figure \ref{fig:pipeline-example}, for an example from an intent detection task \citep{gabriel-etal-2022-misinfo}, a headline \textit{``Remote glaciers in China melting at shocking pace, risking water shortages''} and an intent \textit{``Climate change is real and is showing its effects''} would be presented to crowdworkers. They might connect these two statements with the knowledge \textit{``water shortage is a type of effect''} which would be represented as \textit{(head: water shortages, relation: IsA, tail: effect)}

\begin{figure*}
     \centering
     \begin{subfigure}[b]{0.15\textwidth}
         \centering
         \includegraphics[width=\textwidth]{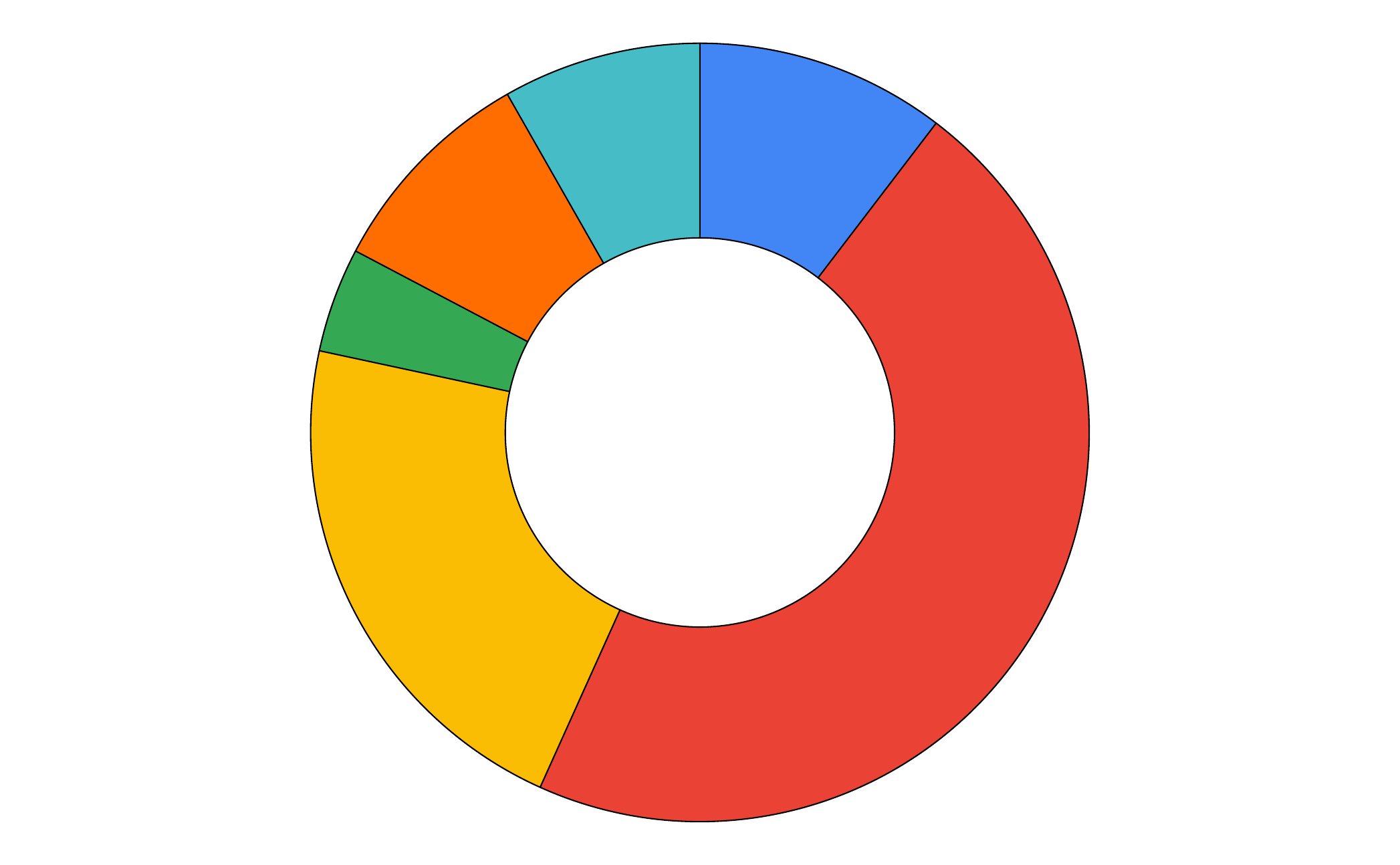}
         \caption{Dialogue}
         \label{fig:dialogue}
     \end{subfigure}
     \hfill
     \begin{subfigure}[b]{0.15\textwidth}
         \centering
         \includegraphics[width=\textwidth]{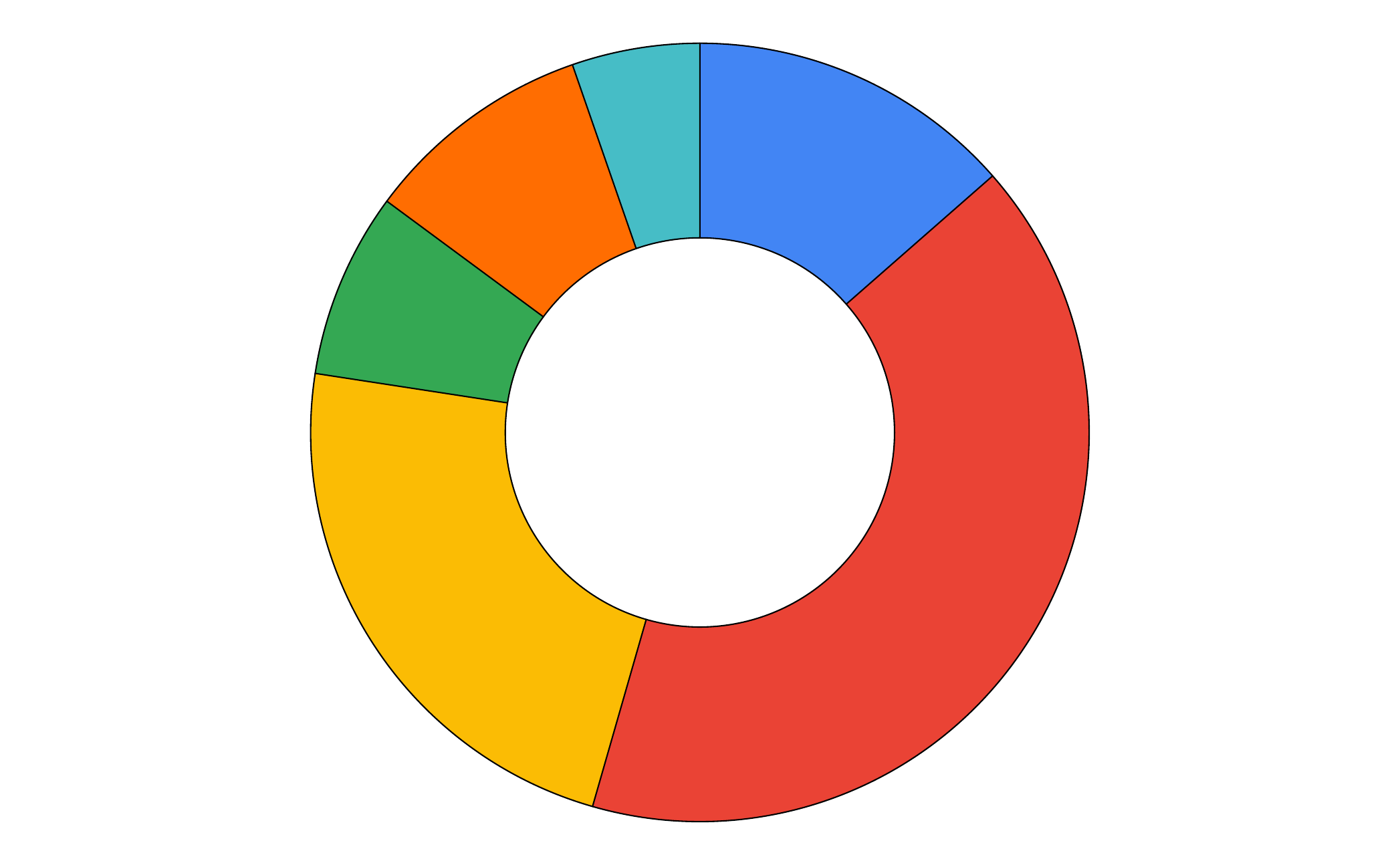}
         \caption{Dialogue Sum.}
         \label{fig:dialogsum}
     \end{subfigure}
     \hfill
     \begin{subfigure}[b]{0.15\textwidth}
         \centering
         \includegraphics[width=\textwidth]{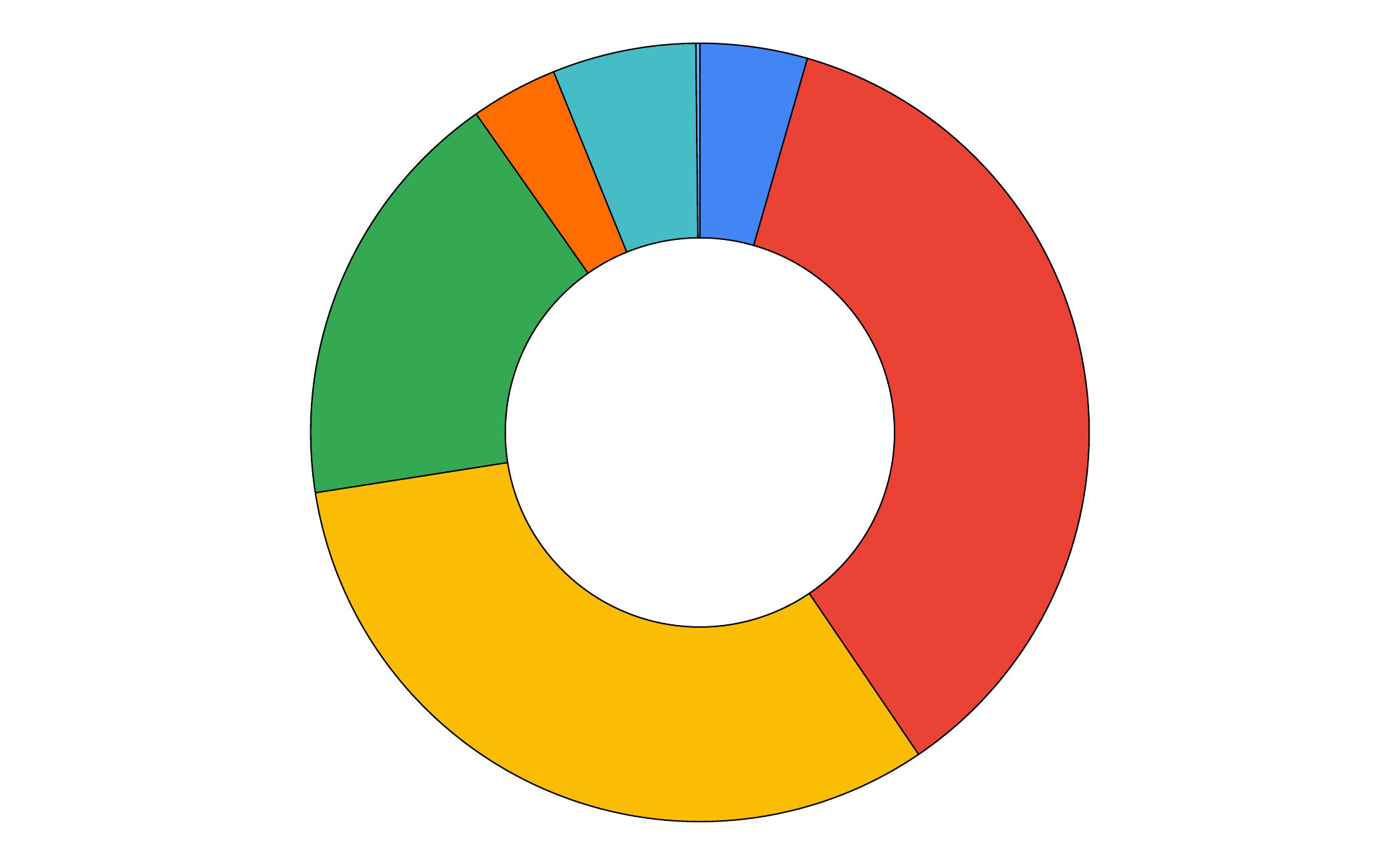}
         \caption{Intent Det.}
         \label{fig:intent-detection}
     \end{subfigure}
     \begin{subfigure}[b]{0.15\textwidth}
         \centering
         \includegraphics[width=\textwidth]{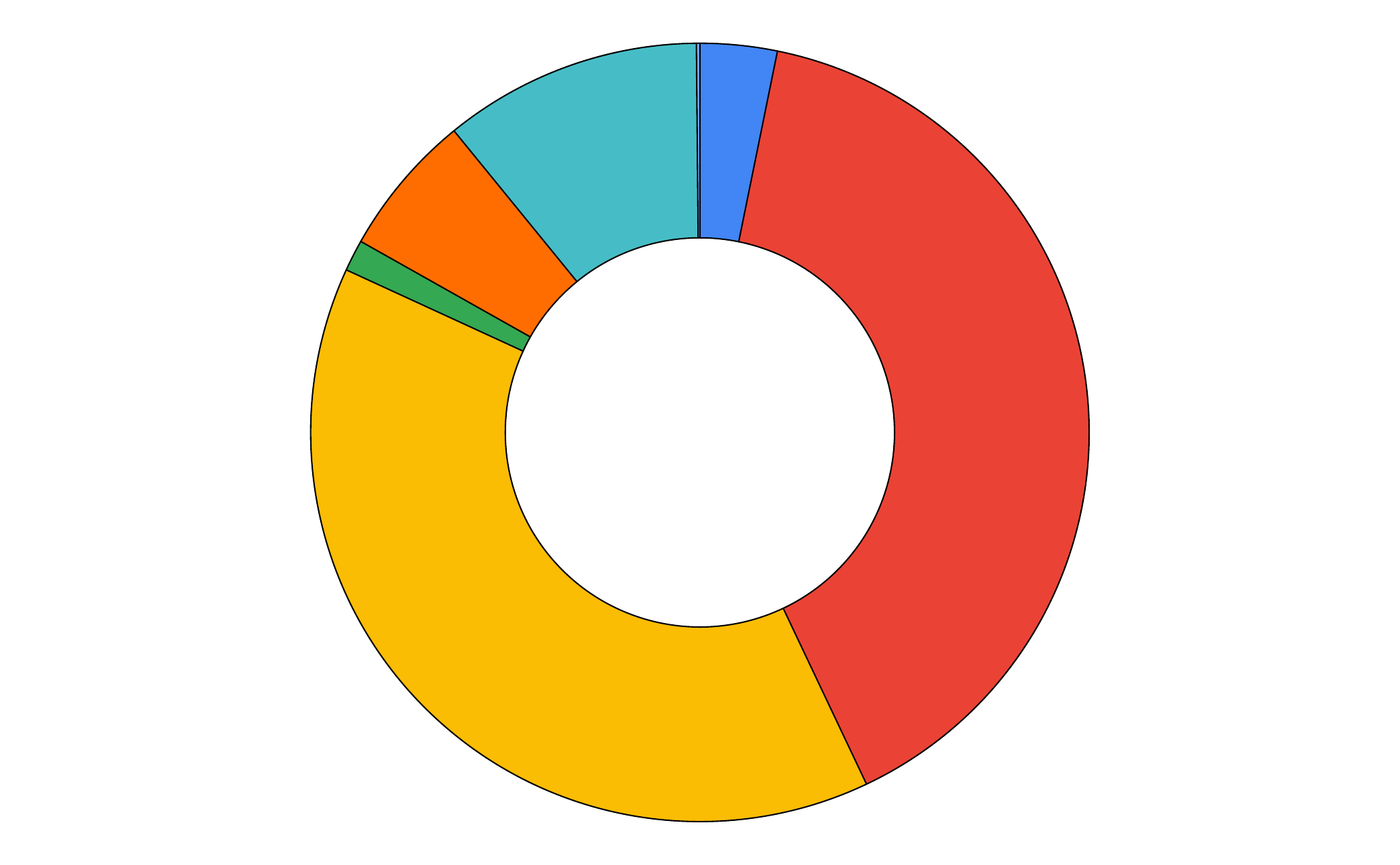}
         \caption{Safety Det.}
         \label{fig:safety-detection}
     \end{subfigure}
     \begin{subfigure}[b]{0.15\textwidth}
         \centering
         \includegraphics[width=\textwidth]{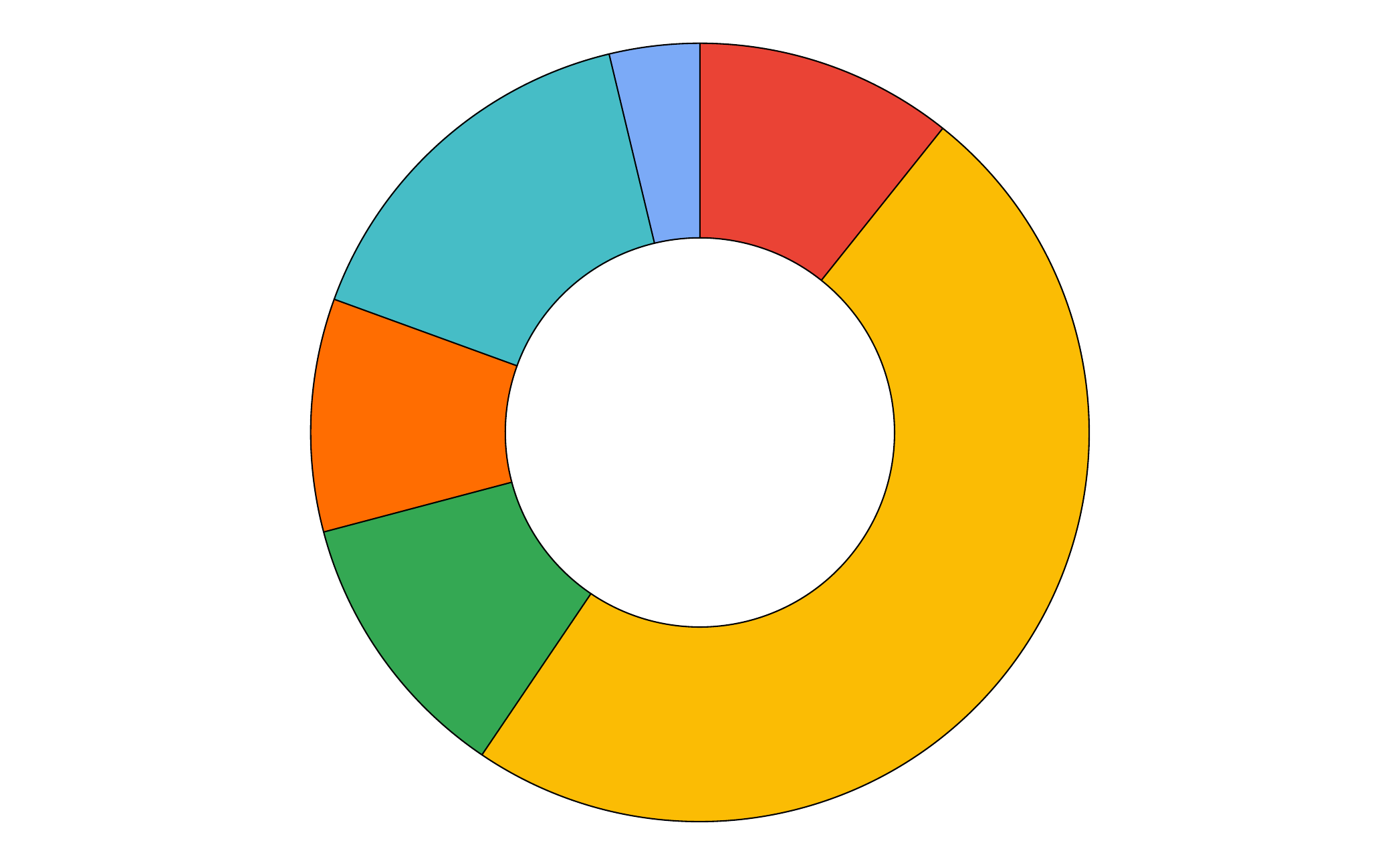}
         \caption{Stance Class.}
         \label{fig:stance-detection}
     \end{subfigure}
    \begin{subfigure}[b]{0.15\textwidth}
         \centering
         \includegraphics[width=\textwidth]{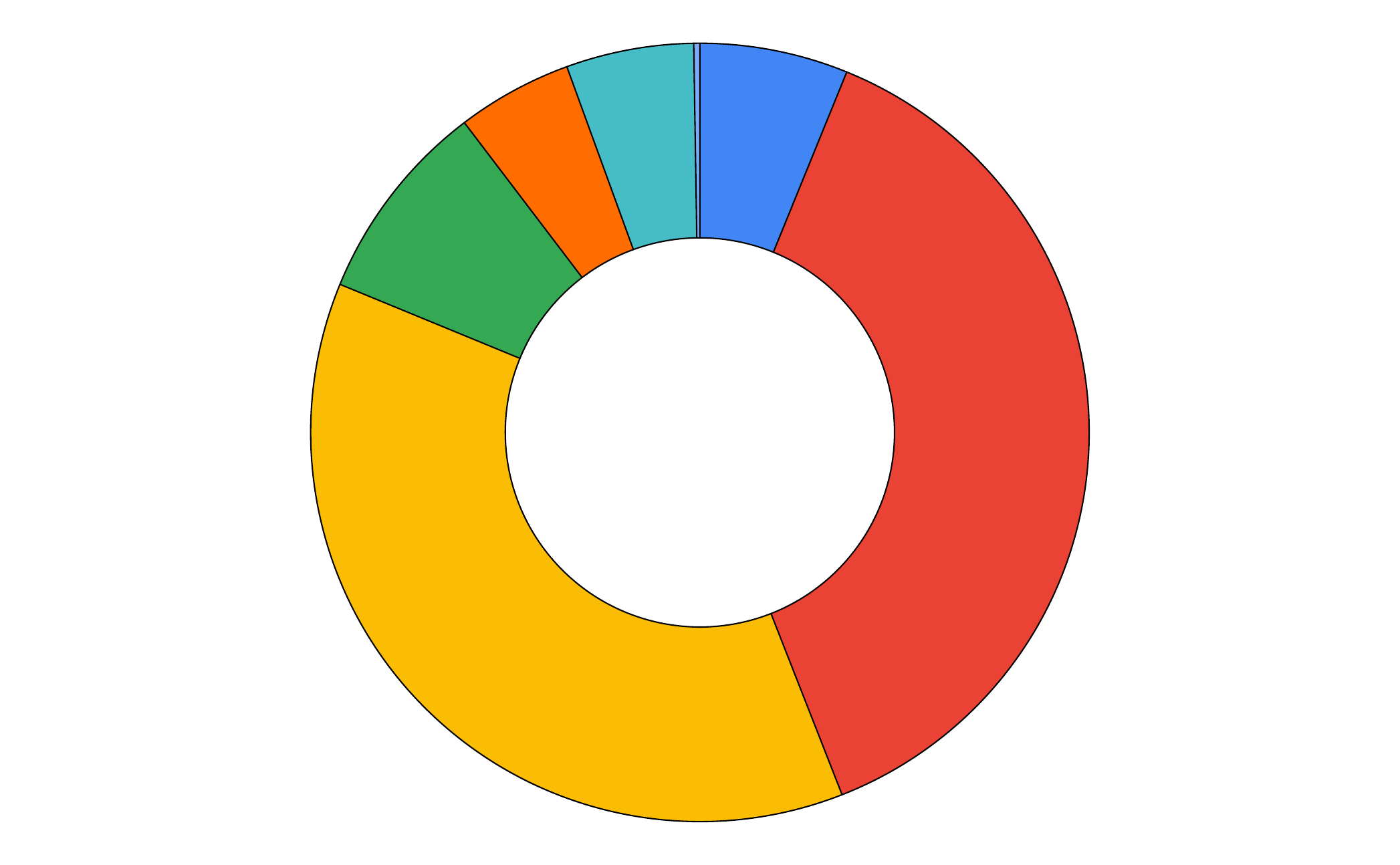}
         \caption{Translation}
         \label{fig:machine-translation}
     \end{subfigure}
    \begin{subfigure}[b]{\textwidth}
         \centering
         \includegraphics[width=0.75\textwidth]{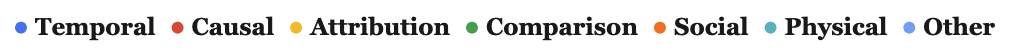}
     \end{subfigure}
       \vspace{-3mm}
        \caption{Distribution of commonsense knowledge dimensions across tasks}
        \label{fig:dim-distribution}
\end{figure*}

Based on earlier work \cite{ilievski2021dimensions, Speer2016ConceptNet5A, Sap2019ATOMICAA, ghosal-etal-2021-cider}, we also categorize relations into six dimensions of commonsense knowledge: \textit{Attributional, Physical/Spatial, Temporal, Causal, Social and Comparative}. Figure \ref{fig:dim-distribution} shows the distribution of dimensions per task.\footnote{Appendix \ref{sec:app-cka} provides more details on the selection and categorization of the relations.} The dimensions serve as support for a fine-grained analysis of the commonsense reasoning abilities of models when tackling tasks. Following the CKA stage, we apply a validation phase to filter out low-quality annotations. {For example, in Figure \ref{fig:pipeline-example}, the knowledge \textit{``glacier HasProperty real''} would be filtered by crowd workers as it is not helpful for the task in the given context.} Each annotation is verified by three unique workers, and we take the majority vote as the qualifying threshold for the next stage.

\paragraph{Winograd Schema Generation and Validation}
In this stage, we present workers with a \textit{context}, a \textit{target}, and the associated commonsense knowledge from the previous stage, and ask them to rewrite the \textit{target} such that it satisfies the following four conditions.\footnote{Additional details on the generation instructions can be found in Appendix \ref{sec:app-instructions}.} The new \textit{target} must (1) minimally differ from the original target (\ie, by edit distance of at most five words), 
(2) directly violate the given commonsense knowledge,  
(3) be an incorrect answer for the given context, 
and (4) be contextually relevant. Conditions (1) and (2) are based on the core design of Winograd schemas, and we introduce conditions (3) and (4) to increase the difficulty of the generated schemas.
Each annotated schema is further validated by three unique workers with respect to the conditions above, and those with at least two valid votes proceed to the final expert validation stage. For example, in Figure \ref{fig:pipeline-example}, given the knowledge \textit{``water shortages IsA effect''}, annotators might produce Winograd-style schemas where the word \textit{``effect''} in the given intent is replaced with related concepts such as \textit{``causes''} or \textit{``results''}. However, as \textit{``results''} would not change the underlying intent of the example, the schema based on this replacement would not satisfy condition (3) above, and hence would be filtered in the validation stage. In Appendix \ref{sec:app-wsg}, we provide more examples of violations of each condition.

\subsection{Data Quality Verification}

\paragraph{Qualification}
In order to collect high-quality annotations, we design a qualification test consisting of multiple-choice and open-ended questions. 
Following earlier work that identified the importance of a large pool of annotators for data diversity \cite{geva-etal-2019-modeling}, we qualify $58$ workers located in the US based on a precision threshold of $0.8$ on the multiple-choice questions and a manual review of open-ended commonsense knowledge annotations. Based on the best practices for an effective crowdsourcing protocol \cite{nangia-etal-2021-ingredients}, we further train the annotators on a small sample of examples from our tasks, regularly engaging with them and sending feedback during the whole data collection process. Instruction templates and details about this test can be found in Appendix \ref{sec:app-qual}


\begin{table}[t]
\centering
\small 
\resizebox{\linewidth}{!}{
    
    \begin{tabular}{lcccccc}
    \toprule
   \textbf{Task}  & \# \textbf{Contexts} & \# \textbf{Examples} \\ 
    \midrule
    Dialogue &  1,169 & 3,548 \\
    Dialogue Summarization &  453 & 1,805 \\
    Machine Translation (zh-en) &  600 & 1200 \\
    Machine Translation (en-de) &  500 & 1000 \\
    Machine Translation (en-fr) &  500 & 1000 \\
    Machine Translation (en-ru) & 500 & 1000 \\
    Intent Detection &  589 & 2,440 \\
    Stance Classification &  397 & 1,722 \\
    Safety Detection &  366 & 2,826 \\
   \midrule
   Total & 5,074 & 16,541 \\
   \bottomrule
\end{tabular}}
    \caption{Statistics of the \ourbenchmark{} benchmark.}
    \label{tab:benchmark_stats}
\end{table}

\section{\ourbenchmark}
\ourbenchmark{} consists of six real-world NLP tasks where commonsense reasoning ability is implicitly required to solve the task correctly. 
Initially, to select tasks that could serve as good testbeds for \ourbenchmark, we
followed \citet{davis2023benchmarks}, and identified the following desiderata of tasks in the benchmark: (1) tasks should represent real-world applications of natural language technologies (\eg, machine translation), (2) tasks should involve rich commonsense knowledge use and reasoning, and (3) tasks should be easy for humans.
Our final benchmark contains {$\sim$5K} unique contexts with {$\sim$500} unique contexts per task (on average) {and $\sim$16K examples (\ie, context-target pairs) in total}. Table \ref{tab:benchmark_stats} provides statistics about our benchmark (additional statistics can be found in Table \ref{tab:benchmark_stats_2} in the Appendix). In this section, we outline the methodology for selecting of real-world tasks that require commonsense reasoning,
as well as a brief overview of each task included in our benchmark. 

\subsection{Task Selection}
To identify NLP tasks that satisfy our desiderata above, we first crawl papers from the whole ACL anthology published since the year 2000 (approximated $94$K papers). Next, we 
select the papers that have done an error analysis and mention \textit{commonsense} or \textit{world knowledge} in their categories of errors.\footnote{Many papers report an analysis of error types, and often identify commonsense reasoning errors as a typical category.} This step results in around $200$ papers.\footnote{This number is an underestimation, as some papers were not considered due to parsing failures.} A further manual review of these papers to filter out false positives reduces this number to $82$, and we categorize and group the resulting papers by tasks which yield around $25$ potential tasks. 

Out of these discovered tasks, we select three \textit{classic} NLP tasks -- machine translation, summarization, and dialogue response generation -- that are also often used to evaluate the abilities of general generative language models. 
In addition, we select three tasks that are more applied and specialized -- intent detection, stance classification, and safety detection.
Other tasks that were discovered as part of this pipeline include toxicity detection, relation extraction, and fact-checking. However, due to the difficulty of generating commonsense-violating perturbations for these tasks (caused by their factual or obscene nature), we leave their integration into our benchmark as future work. 

\subsection{\ourbenchmark{} Tasks}
\label{sec:tasks}
We apply our pipeline (\S\ref{sec:data_collection}) to the six real-world tasks identified in the selection phase. 
For each task, we select a recent existing dataset that contains contexts rich with the use of commonsense knowledge. Some of the chosen datasets already include annotations for commonsense knowledge or Winograd schemas, allowing us to skip parts of the pipeline. Here, we describe these tasks and datasets in more detail and identify task-specific variations of the pipeline for each. 

\noindent \textbf{Machine translation (MT)} is known to require commonsense knowledge \cite{BarHillel1960ADO} to resolve translation errors.
We select the test suite constructed by \citealp{he-etal-2020-box} for Chinese-English translation and the Wino-X dataset \cite{emelin-sennrich-2021-wino} for English to German, French, and Russian translation. Both datasets consist of Winograd-style examples containing a source sentence and two translations that minimally differ from each other, but only one of which is correct due to underlying commonsense knowledge.

\noindent \textbf{Open-domain Dialogue (DG)} is a core real-world NLP task requiring systems to produce chat responses to a given conversational context. 
The important role of commonsense knowledge and reasoning in open-domain dialogue systems has been well-documented \cite{richardson2023commonsense}. For this task, we choose the CIDER dataset \cite{ghosal-etal-2021-cider}, which already contains expert-annotated commonsense knowledge that connects utterances in different turns of the dialogue.

\noindent \textbf{Dialogue summarization (DS)} is another NLP task with real-world applications (\eg, meeting, email summarization). Also, enhancing summarization models with commonsense knowledge has been shown to generate more informative and abstractive summaries \cite{Kim2022MindTG}.
For this task, we choose the test split of the DialogSum dataset \cite{chen-etal-2021-dialogsum}, which contains real-life dialogues along with their abstractive summaries.

\noindent \textbf{Intent detection (ID)} is the task of 
identifying the underlying intent of the author of the text. As the intent is typically implicit, it involves significant use of commonsense knowledge. For this task, we choose the Misinformation Reaction Frames dataset proposed by \citet{gabriel-etal-2022-misinfo}, which contains news headlines along with news writers' intents behind them and readers' reactions to them.

\noindent \textbf{Stance classification (SC)} involves inferring the \textit{stance} (either supporting or opposing) of an argument given a belief. Such a task typically requires understanding social, cultural or ontological commonsense knowledge. We use the ExplaGraphs dataset \cite{saha-etal-2021-explagraphs}, which provides, for each argument-belief pair, a crowd-sourced commonsense explanation graph that explains the stance between the two sentences through a set of commonsense knowledge triplets. 

\noindent \textbf{Safety detection (SD)}, detecting safe actions in a given scenario, has real-world applications, especially in the deployment of autonomous robots and systems capable of giving advice. This task requires the use of commonsense knowledge, especially when the action is not explicitly violent which makes it much harder for the system to assess its safety.
For this task, we use the SafeText dataset \cite{levy-etal-2022-safetext}, where each sample consists of a sentence describing a real-life scenario and a list of safe and unsafe actions that could be taken in these situations. 

%% file: 04_experiments.tex
\input{results-table}

\textbf{Task Formulation.} All tasks in \ourbenchmark{} are treated as binary classification tasks. Given a context, a model must predict whether a provided \textit{target} is a suitable response for the corresponding real-world task. For instance, in machine translation, given an English sentence and a translated sentence in French, the model must predict whether the translation is valid or not. 

\paragraph{Evaluation Metrics.} 
We evaluate models on CRoW using two scores: \textbf{Macro-F1} of predicting valid and invalid \textit{targets}, and \textbf{Situational Accuracy}, a stringent metric that reports whether the model correctly identifies the validity (or invalidity) of \textbf{all} \textit{targets} for a given \textit{context} (similar to \citealp{storks-chai-2021-beyond-tip}'s strict coherence score). 
A single mistake on any \textit{target} results in a score of $0$ for that context. We design this metric to account for the fact that robust commonsense reasoning would provide the model with a full situational understanding of the \textit{context}. The \ourbenchmark{} score is computed as a macro-average of the task scores.

\paragraph{Models.}
We evaluate a series of language models that are diverse in terms of scale, training, and data:
\begin{itemize}
[leftmargin=*,topsep=1pt,parsep=1pt]
    \item \textbf{LLaMA} \cite{touvron2023llama}, an open-source decoder-only model with various sizes (7B, 13B, 33B parameters) and \textbf{PaLM-1-540B} \cite{chowdhery2022palm}, a closed-source decoder-only model with 540B parameters. Both models are pretrained using only a language modeling loss.
    \item \textbf{GPT-3.5} \cite{NEURIPS2020_gpt3} and \textbf{GPT-4} \cite{openai2023gpt4}: two closed-source decoder-only models that were trained with instruction-tuning. For GPT-3.5, we use the \texttt{text-davinci-003} model with 175B parameters. 
    \item \textbf{Alpaca} \cite{alpaca}, \textbf{Vicuna} \cite{vicuna2023} and \textbf{Stable-Vicuna}: 
     three open-source decoder-only models based on LLaMA. Alpaca has 7B parameters, while Vicuna and Stable-Vicuna have 13B. They are instruction-tuned using different instructions-following datasets; Stable-Vicuna is further fine-tuned with RLHF.
    \item \textbf{Flan-T5-XXL} (\citealp{chung2022scaling}, 11B parameters) and \textbf{Flan-Alpaca} (\citealp{chia2023instructeval,peng2023instruction}; 3B), two open-source encoder-decoder models based on T5 \cite{raffel2020exploring} and trained on instruction-following datasets.
    \item \textbf{BloomZ-7B}\ and \textbf{mT0-xxl} \cite{muennighoff2022crosslingual}, two open-source instruction-following multilingual language models of 7.1B and 13B parameters, respectively. The former is a decoder-only model fine-tuned from BLOOM \cite{scao2022bloom} while the latter is an encoder-decoder fine-tuned from mT5 \cite{xue2020mt5}.
\end{itemize}
 



\noindent All models are evaluated using one-shot in-context learning and greedy decoding.\footnote{Further results with varying temperature values are in Appendix \ref{sec:app-results}.} We use the same task-specific prompt templates for all models.\footnote{More details on prompt templates are in Appendix \ref{sec:app-tasks}.} 
We also report the performance of a \textit{random} baseline that randomly chooses whether a context and target pair is valid, and a \textit{majority} baseline, which selects the most frequent label for each task. 

\paragraph{Human Evaluation.}
{We evaluate the human performance on each task of the benchmark using two expert annotators who evaluate $100$ random samples from the task. Our experts are NLP researchers from our lab who were not involved in the original data collection. As a result, they are more experienced, can clarify misunderstandings in the annotation guideline with us, and generally produce more careful annotations than crowd workers. Following \citet{amidei-etal-2018-rethinking} and \citet{oortwijn-etal-2021-interrater}, we intentionally allow evaluators to discuss and reach a final answer in cases of disagreement, which reduces variance and yields a robust upper bound for our task. In Appendix \ref{sec:app-human-eval}, we provide further details on the number of resolved disagreements, the human performance before and after the discussion, and the statistical significance of the human evaluation results.} 

%% file: results-table.tex
\begin{table*}[t!]
\footnotesize
\scalebox{0.72}{
\centering
\begin{tabular}{lrrrrrrrrrrr}
\toprule 
   \multirow{2}{*}{\textbf{Models}}
 & \multicolumn{4}{c}{{\textbf{MT}}} 
 & \multicolumn{1}{c}{{\textbf{DG}}}
 & \multicolumn{1}{c}{{\textbf{DS}}}
 & \multicolumn{1}{c}{{\textbf{SC}}}
 & \multicolumn{1}{c}{{\textbf{SD}}}
 & \multicolumn{1}{c}{{\textbf{ID}}}
 & \multicolumn{1}{c}{{\textbf{CROW}}}
 & \multicolumn{1}{c}{{\textbf{CROW}}}  \\
 
 & \multicolumn{1}{c}{Zh-En}& \multicolumn{1}{c}{En-Fr}  & \multicolumn{1}{c}{En-De} & \multicolumn{1}{c}{En-Ru} & & & & & & \multicolumn{1}{c}{\textbf{Score (-MT)}} & \multicolumn{1}{c}{\textbf{Score}} \\
 \midrule
\textbf{Majority} & 33.3 / 0.0 & 33.3 / 0.0 & 33.3 / 0.0 & 33.3 / 0.0 & 40.1 / 0.0 & 42.8 / 0.0 & 33.6 / 0.0 & 36.5 / 0.0 & 41.3 / 0.0 & 38.9 / 0.0 & 36.4 / 0.0 \\
\textbf{Random} & 49.5 / 25.7 & 50.8 / 25.3 & 51.7 / 25.9 & 47.7 / 22.5 & 47.3 / 13.9 & 45.5 / 9.6 & 51.3 / 6.6 & 50.6 / 0.8 & 48.8 / 10.6 & 48.7 / 8.3 & 49.3 / 15.6 \\
\midrule
\textbf{LLaMA-7B} & 49.9 / 0.0 & -- & -- & -- & 48.7 / 0.7 & 53.2 / 4.9 & 57.6 / 0.8 & 29.9 / 0.0 & 41.3 / 0.0 & 46.1 / 1.3 & 46.8 / 1.1 \\
\textbf{LLaMA-13B} & 50.7 / 1.7 & -- & -- & -- & 50.6 / 7.9 & 40.5 / 2.0 & 57.6 / 1.8 & 32.7 / 0.5 & 41.5 / 0.0 & 44.6 / 2.4 & 45.6 / 2.3 \\
\textbf{LLaMA-33B} & 50.5 / 1.2 & -- & -- & -- & 50.5 / 2.6 & 48.2 / 7.8 & 57.1 / 0.0 & 44.1 / 4.1 & 42.4 / 1.2 & 48.5 / 3.1 & 48.8 / 2.8 \\
\textbf{Flan-T5-11B} & 45.5 / 10.1 & -- & -- & -- & 70.4 / 42.0 & 66.9 / 33.1 & 76.5 / 51.6 & 83.8 / 34.9 & \textbf{84.3 / 57.7} & 76.4 / 43.9 & 71.2 / 38.2 \\
\textbf{Alpaca} & 56.0 / 13.4 & -- & -- & -- & 55.2 / 15.3 & 48.5 / 9.6 & 55.9 / 14.4 & 55.6 / 6.6 & 60.1 / 17.7 & 55.1 / 12.7 & 55.2 / 12.8 \\
\textbf{Flan-Alpaca} & 60.5 / 25.5 & -- & -- & -- & 62.3 / 26.4 & 52.3 / 18.7 & 72.2 / 43.8 & 75.0 / 21.4 & 78.2 / 45.7 & 68.0 / 31.2 & 66.7 / 30.3 \\
\textbf{Vicuna} & 61.3 / 26.8 & -- & -- & -- & 60.6 / 20.4 & 64.6 / 22.2 & 64.5 / 24.3 & 65.4 / 14.0 & 68.5 / 28.8 & 64.7 / 22.0 & 64.1 / 22.8 \\
\textbf{Stable-Vicuna} & 53.5 / 8.9 & -- & -- & -- & 52.0 / 11.5 & 38.6 / 7.1 & 59.6 / 8.9 & 72.8 / 20.9 & 59.9 / 17.4 & 56.6 / 13.1 & 56.1 / 12.4 \\
\midrule
\textbf{mT0} & 53.8 / 11.2 & 39.5 / 1.8 & 44.7 / 1.8 & 44.2 / 1.6 & 40.8 / 0.4 & 47.8 / 3.8 & 49.2 / 12.9 & 45.2 / 2.5 & 63.3 / 21.0 & 49.3 / 8.1 & 47.6 / 6.3 \\
\textbf{BloomZ-7B} & 45.4 / 8.2 & 45.0 / 3.4 & 46.2 / 1.2 & 49.9 / 2.4 & 49.8 / 7.5 & 41.4 / 6.7 & 58.8 / 15.2 & 67.6 / 8.5 & 64.7 / 21.2 & 56.5 / 11.8 & 52.1 / 8.3 \\
\midrule
\textbf{PaLM-1-540B} & 52.7 / 5.7 & 50.2 / 0.4 & 50.0 / 0.0 & 50.0 / 0.0 & 63.4 / 24.7 & 61.2 / 20.2 & 51.3 / 19.1 & 49.5 / 7.7 & 70.4 / 32.3 & 59.2 / 20.8 & 55.4 / 12.2 \\
\textbf{GPT-3.5} & 66.6 / 38.7 & 50.1 / 18.2 & 50.6 / 18.0 & 48.9 / 13.2 & 67.6 / 36.5 & 68.7 / 31.9 & 67.7 / 36.0 & 85.6 / 40.0 & 76.4 / 41.7 & 73.2 / 37.2 & 64.7 / 30.5 \\
\textbf{GPT-4} & \textbf{75.9 / 57.9} & 54.5 / 21.5 & 54.4 / 20.5 & 54.1 / 19.7 & \textbf{72.4 / 46.5} & \textbf{89.6 / 75.3} & 79.6 / 54.7 & \textbf{89.7 / 51.9} & 84.0 / 57.2 & \textbf{83.1 / 57.1} & \textbf{72.7 / 45.0} \\
\textbf{GPT-4-CoT} & 71.6 / 52.2 & \textbf{64.7 / 42.6} & \textbf{57.1 / 34.2} & \textbf{57.3 / 30.0} & \underline{55.3 / 22.8} & 88.6 / 70.6 & \textbf{84.3 / 60.7} & 87.8 / 47.3 & 84.0 / 57.0 & 80.0 / 51.7 & 72.3 / 46.4 \\

\midrule 
 \textbf{Human$^*$}   &  87.9  /  78.0  & 83.0  /  82.9  &  89.9  /  82.0 & 89.9  /  86.0  &  87.0  /  86.9 & 98.9  /  96.4 & 88.1  /  69.6 & 97.8  /  93.9 & 93.9  /  80.7 & 93.1  /  85.5  & 90.7  /  84.0 \\
\bottomrule
\end{tabular}}
\caption{\textbf{Macro-F1}  /  \textbf{Situational Accuracy} (\ie, results aggregated per \textit{context} instead of per \textit{sample}) for all examined models across \ourbenchmark{} tasks. The performance of the highest scoring model is \textbf{bolded} for each task. $^*$Due to the cost of expert evaluation, our \textbf{Human} study is only evaluated on 100 instances per task.}
\label{tab:main_results-greedy}
\end{table*}

%% file: 05_results.tex

Table \ref{tab:main_results-greedy} reports the results for all models across all tasks. In general, we observe that models vary in their ability to correctly identify the correct responses in the tasks. As expected, GPT-4 outperforms most other models, many of which actually perform worse than the random baseline (\eg, all LLaMA variants). Even among stronger models, though, while performance is higher for individual examples (as measured by Macro-F1), the situational accuracy is significantly lower, often below $50\%$. This gap suggests that these models are not robust and fail to grasp a full situational understanding of the contexts with which they are presented (even as they may correctly classify some individual cases). In contrast, humans tend to perform well on both metrics (with little gap between individual example performance and situational accuracy). Perhaps most surprisingly, our results show that chain-of-thought harms the performance of GPT-4 on some of the tasks, {particularly on the Dialogue task (DG) where the performance drops by $-17.1\%$ in Macro-F1 and $-23.7\%$ in Situational Accuracy (underlined in Table \ref{tab:main_results-greedy}). This behavior} perhaps hints that chain-of-thought decoding is less useful in commonsense tasks requiring implicit, intuitive inferences, rather than complex, multi-step reasoning. {In the Analysis section, we provide more details on the possible causes for the discrepancy in performance with examples.}


\begin{table}[t]
\scalebox{0.55}{
    \centering
    \begin{tabular}{lrrrrrr}
    \toprule
    \textbf{Model} & \multicolumn{6}{c}{{\textbf{CK Dimensions}}} \\
    \midrule
    & Attribution & Physical & Temporal & Causal & Social & Comparison \\
    \midrule
\textbf{Flan-Alpaca$^\clubsuit$} & 70.2 & 72.2 & \underline{68.0} & 70.5 & 72.3 & 73.5 \\
\textbf{Flan-T5-11B$^\clubsuit$} & \underline{77.2} & 78.3 & 78.4 & 78.3 & 79.5 & 79.7 \\
\textbf{LLaMa-33B$^\clubsuit$} & 46.9 & 46.4 & 48.0 & 46.8 & 46.9 & \underline{45.8} \\
\textbf{Stable-Vicuna$^\clubsuit$} & 55.6 & 57.5 & 56.9 & \underline{55.2} & 58.9 & 55.4 \\
\midrule
\textbf{BloomZ-7B} & 53.0 & 54.0 & 51.4 & 52.4 & \underline{51.0} & 51.1 \\
\textbf{PaLM-1-540B} & 56.0 & \underline{53.9} & 57.9 & 54.3 & 57.9 & 55.2 \\
\textbf{GPT-3.5} & 65.3 & 64.1 & \underline{56.6} & 64.3 & 65.8 & 70.1 \\
\textbf{GPT-4} & 74.4 & 73.1 & 71.2 & 73.2 & 72.6 & \underline{70.6} \\
\textbf{GPT-4-CoT} & 73.4 & 72.0 & \underline{69.0} & 71.7 & 73.1 & 74.0 \\
   \bottomrule
    \end{tabular}}
    \caption{\textbf{Macro-F1} scores averaged across commonsense dimensions. 
    ($^\clubsuit$all tasks except for MT)}
    \label{tab:results_commonsense_dimension}
\end{table}

\paragraph{Instruction-tuning.} Models that were trained only with language modeling objectives (\eg, LLaMA and PaLM) obtain lower scores compared to instruction-tuned models of similar size. For example, {Alpaca}, which is an instruction-tuned version of LLaMA-7B, achieves an {average $\sim$10\%} improvement compared to LLaMA-7B across most tasks for both metrics. Also, smaller instruction-tuned models can perform similarly or exceed the performance of much larger models (\eg, GPT-3.5 outperforms PaLM). 
Finally, we find that Stable-Vicuna surprisingly performs worse than Vicuna, 
suggesting that while instruction-tuning improves performance on \ourbenchmark{}, training with RLHF does not necessarily amplify the commonsense reasoning abilities required for these tasks. 

\begin{table}[t]
\centering
\scalebox{0.7}{
    \begin{tabular}{lrr}
    \toprule
    \textbf{Model} & \textbf{Oracle Knowledge} & \textbf{No Knowledge} \\
     \midrule
      \textbf{Flan-T5-11B$^\clubsuit$}  & 77.9 / 48.8 & 76.4 / 43.9\\ 
     \textbf{BloomZ-7B}  & 52.0 / 8.9 & 52.1 / 8.3\\ 
     \textbf{GPT-4}  & 74.5 / 47.6 & 72.7 / 45.0  \\
     \textbf{GPT-4-CoT}  & 76.9 / 53.1 &  72.3 / 46.4 \\
   \bottomrule
    \end{tabular}}
    \caption{\textbf{Macro-F1} / \textbf{Situational Accuracy} scores averaged over all tasks ($^\clubsuit$: all tasks except MT), with and without providing oracle commonsense knowledge as part of the prompt.}
    \label{tab:results_oracle}
\end{table}

\paragraph{Scale.} When we compare the same model with different scales, we do not find a consistent benefit to increasing the size of the model, except on the safety detection task, where LLaMA-33B achieves a $14.2$\% and $11.4$\% improvement score over LLaMA 7B and 13B, respectively.

\begin{figure*}[t]
    \centering    
    \includegraphics[width=0.9\textwidth]{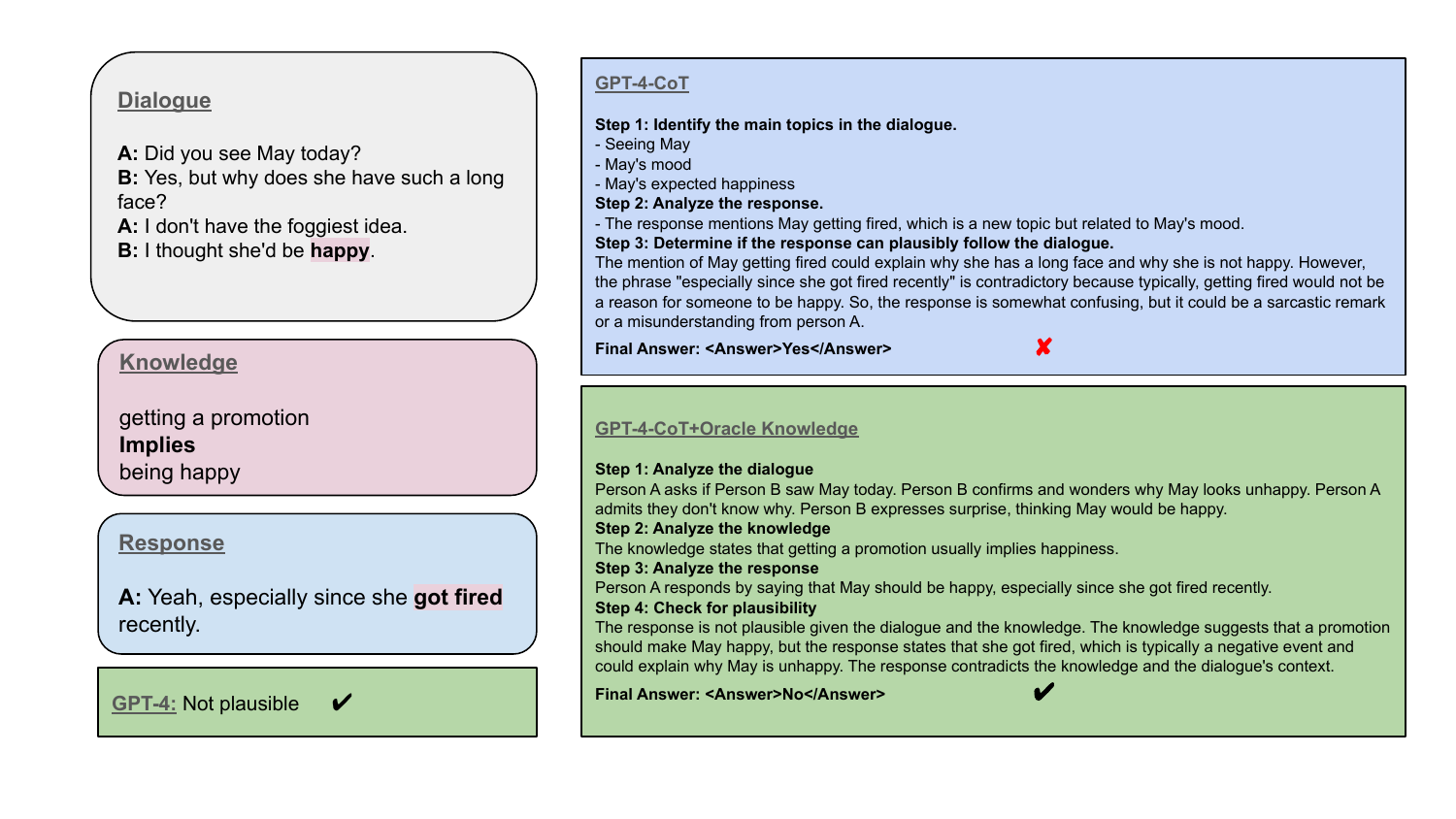}
    \caption{\textbf{Answers generated by GPT-4 in various settings for an example from the Dialogue task.} While chain-of-thought reasoning yields an incorrect answer, adding the commonsense knowledge guides the model toward the correct answer. Note that since the knowledge is annotated for the original, plausible response, \textit{``getting a promotion''} is no longer relevant to the annotated, implausible response, which was modified with \textit{``got fired''}.}
    \label{fig:analysis-ex}
\end{figure*}

\paragraph{Multilinguality.} Most of these models are officially monolingual, though they may have been pretrained on some non-English data. Since one of our testbed tasks centers on machine translation, we evaluate multilingual models on our benchmark. BloomZ performs better than mT0 across most tasks. Certain monolingual models outperform BloomZ on translation tasks (\ie, those with >100B parameters), suggesting these models have seen multilingual data during their pretraining phase. 

%% file: 06_analysis.tex

\paragraph{Dimensions of Commonsense Knowledge.}
Table \ref{tab:results_commonsense_dimension} reports the performance of different models across different commonsense knowledge dimensions.  
We observe that these models perform fairly consistently across different examples grounded by different commonsense dimensions, indicating that they do not generally learn more reliable commonsense reasoning skills of one variety compared to another. 
Part of this uniformity is due to conceptual overlap between commonsense dimensions (\eg, certain social commonsense relations\footnote{We used relations from existing KGs, see Appendix~\ref{sec:app-kg-relations}} may also reflect causal commonsense), a nuance that is not captured by our methodology that requires annotation of a single relation for commonsense knowledge.  
Regardless of this overlap, our findings highlight that, in real-world tasks, there remains room for commonsense modeling improvement for most types of commonsense knowledge. 



\paragraph{Can models leverage explicit commonsense knowledge?}

When constructing \ourbenchmark, we annotate the implicit commonsense relationship required to understand whether a target is valid or invalid given a context. To test whether models can leverage this implicit commonsense relationship for making a correct prediction, we perform an oracle experiment where we augment the prompt with each example's annotated commonsense knowledge triplet. 
Table \ref{tab:results_oracle} shows the impact of adding commonsense knowledge to the prompt for the best closed-source (GPT-4) and open-source (Flan-T5 and BloomZ-7B) models. Prompting with commonsense knowledge slightly increases the average performance of {Flan-T5 (+1.5\% Macro-F1, +4.9\% Sit. Accuracy) and GPT-4 (+1.8\% Macro-F1, +2.6\% Sit. Accuracy), indicating that even with partial\footnote{We can not assume that the annotated relationships are the \textit{only} pieces of commonsense knowledge needed to reason about the solution.} commonsense knowledge, models still fall short of robustly reasoning about the presented situations. 
However, the largest improvement is observed for GPT-4-CoT, indicating that the model can more accurately ground its own reasoning traces when provided with hints.}

\paragraph{Qualitative Analysis.} 

Following up on the lower performance of GPT-4 with CoT prompting compared to GPT-4, we qualitatively analyze the errors made by the model in the CoT setting for potential patterns in the dialogue generation task. In many cases, the reasoning process of the model either focuses solely on the relevance of the response (rather than its sensibility), or, in some cases, follows a less plausible reasoning path, such as imagining a sarcastic response. In Figure \ref{fig:analysis-ex}, we show an example where GPT-4 correctly answers without chain-of-thought, but fails when prompted to ``think step by step,'' arguing that the response is sarcastically plausible (blue box). While such a response could technically be sarcastic, it violates our commonsense idea of what would be a reasonable response to a helpful query. On the other hand, we also observe the direct effect of providing the oracle commonsense knowledge (green box) on the same example where GPT-4 leverages the given knowledge and makes a distinction between sarcastic possibility and commonsensical plausibility. In Appendix Figure \ref{fig:app-analysis-ex}, we provide another example where GPT-4 with chain-of-thought reasoning simply ignores the inherent contradiction created by the commonsense knowledge violation and focuses on the surface-level relevance of the response.

%% file: 07_conclusion.tex
In this work, we propose \ourbenchmark{}, a multi-task commonsense reasoning benchmark consisting of six real-world tasks. To construct our benchmark, we design a data collection pipeline to systematically crowdsource Winograd-style schemas based on commonsense-violating minimal perturbations. Our evaluation of recent large language models on our benchmark shows that the performance of state-of-the-art models still falls far below human performance with respect to commonsense reasoning in real-world contexts. 


%% file: 08_limitations.tex
 Despite our efforts to build a comprehensive benchmark, \ourbenchmark{} faces several limitations. First, commonsense knowledge has many dimensions, and we only consider six core ones as a basis for our commonsense knowledge annotation stage: temporal, causal, attribution, comparative, physical, and social. Second, as we employ crowdsourcing for generating final Winograd schemas, our benchmark is susceptible to data quality issues, annotation artifacts and biases. Lastly, in our experiments, we do not perform prompt tuning. As GPT-3/4 have been found to be sensitive to prompt construction, performance may vary when using other prompts for the same task.



%% file: 10_acknowledgements.tex
We thank Yonatan Bitton for his feedback and constructive suggestions. Access to PaLM was granted through the Google-mediated academic LLM collaboration program, and we thank Shruti Sheth for her support. We also gratefully acknowledge the support of the Swiss National Science Foundation (No. 215390), Innosuisse (PFFS-21-29), the EPFL Science Seed Fund, the EPFL Center for Imaging, Sony Group Corporation, and the Allen Institute for AI.  Moreover, we thank the following participants in our expert human evaluation task: Beatriz Borges, Maria Glarou, Fawzia Zeitoun, Spiros Chalkias, Badr Alkhamissi, Karina Halevy, Khanh Nguyen, Ghali Chraïbi, Li Mi, Hnagyu Yu, Julian Schnitzler, Alex Rubahn, Aurelio Noca, Arina Rak, Nikita Andreev, Sepideh Mamooler, Soyoung Oh, Angelika Romanou, Silin Gao, Khai Loong Aw. 

%% file: 11_appendix.tex
\section{Commonsense Knowledge Dimensions}
\label{sec:app-kg-relations}
{We consider widely used commonsense knowledge bases such as ConceptNet \cite{Speer2016ConceptNet5A} and ATOMIC \cite{Sap2019ATOMICAA}, as well as recent works such as ATOMIC2020 \cite{Hwang2020COMETATOMIC2O} and CIDER \cite{ghosal-etal-2021-cider} for selecting the commonsense relations. As an initial step, we manually categorize the kind of knowledge relations that appear for each task. Among the total $56$ relations available to us from these sources, we find $22$ relations from ConceptNet (out of $36$), $8$ relations from ATOMIC (out of $9$), $3$ relations from ATOMIC2020 (out of $5$) and $3$ relations from CIDER (out of $6$) appearing most commonly. Following \cite{ghosal-etal-2021-cider, ilievski2021dimensions}, we further categorize these $36$ relations into $6$ commonsense knowledge dimensions. In Table \ref{tab:commonsense_relations}, we list the available relations for each dimension with a brief description and an example. Each relationship is represented as \textit{(A, Relation, B)} where A and B refer to phrases from the context.}

\begin{table*}[ht]
    \centering
    \footnotesize
    \begin{tabular}{ll}
     \toprule
         Name & Description \\
         \midrule
         \multicolumn{2}{l}{\textbf{Attributional Relations}} \\
        \midrule
HasProperty & A has B as a property; A can be described as B. \\
CapableOf & Something that A can typically do is B. \\
HasA & B belongs to A, either as an inherent part or due to a social construct of possession. \\
HasSubEvent & A and B are events, and B happens as a subevent of A. \\
IsA & A is a subtype or a specific instance of B; every A is a B. \\
MannerOf & A is a specific way to do B. Similar to ``Is A'', but for verbs. \\
DependsOn & A depends on B. \\
CreatedBy & A is created by B. \\
    \midrule
     \multicolumn{2}{l}{\textbf{Physical/Spatial Relations}} \\
    \midrule
UsedFor & A is used for B. The purpose of A is B. \\
PartOf & A is part of B. \\
MadeOf & A is made up of B. \\
AtLocation & A happens at location B, or B is a typical location for A. \\
LocatedNear & A and B are typically found near each other. \\
    \midrule
     \multicolumn{2}{l}{\textbf{Temporal Relations}} \\
    \midrule
IsAfter & A happens after B. \\
IsBefore & A happense before B. \\
HappensIn & A happens during B. \\
IsSimultaneous & A and B happens at the same time. \\
HasPrerequisite & In order for A to happen, B needs to happen.\\
    \midrule
     \multicolumn{2}{l}{\textbf{Causal Relations}} \\
    \midrule
Causes & A causes B to happen. \\
Implies & A implies B. \\
HinderedBy & A is less likely to happen because of B. \\
    \midrule
     \multicolumn{2}{l}{\textbf{Social Relations}} \\
    \midrule
xIntent & Person in event A intends to do B. \\
xReact & Person in event A reacts as in B. \\
xNeed & Person in event A needs to do B before doing A. \\
xWant & Person in event A wants to do B. \\
xEffect & Event A will have the effect B on the Person in event A. \\
oReact & Others will react to event A as B. \\
oWant & Others will want to do B for A. \\
oEffect & Event A will have effect B on others. \\
MotivatedByGoal & Someone does A because they want result B. \\
    \midrule
     \multicolumn{2}{l}{\textbf{Comparative Relations}} \\
    \midrule
Antonym & A and B are opposites in some relevant way. \\
Synonym & A and B have very similar meanings.\\
SimilarTo & A is similar to B. \\
DistinctFrom & Something that is A is not B. \\
RelatedTo & A is related to B. \\
DefinedAs & A is defined as B.\\

         \bottomrule
    \end{tabular}
    \caption{List of commonsense relations}
    \label{tab:commonsense_relations}
\end{table*}

\section{Data Collection}
\label{sec:app-data-collection}

At a high level, in the CKA stage, the given example is annotated with relevant commonsense knowledge, and in the WSG stage, the example is rewritten with a Winograd-style minimal perturbation based on the commonsense knowledge from the previous stage. 

\subsection{Qualification}
\label{sec:app-qual}
\begin{figure*}[h]
    \centering    
    \includegraphics[width=\textwidth]{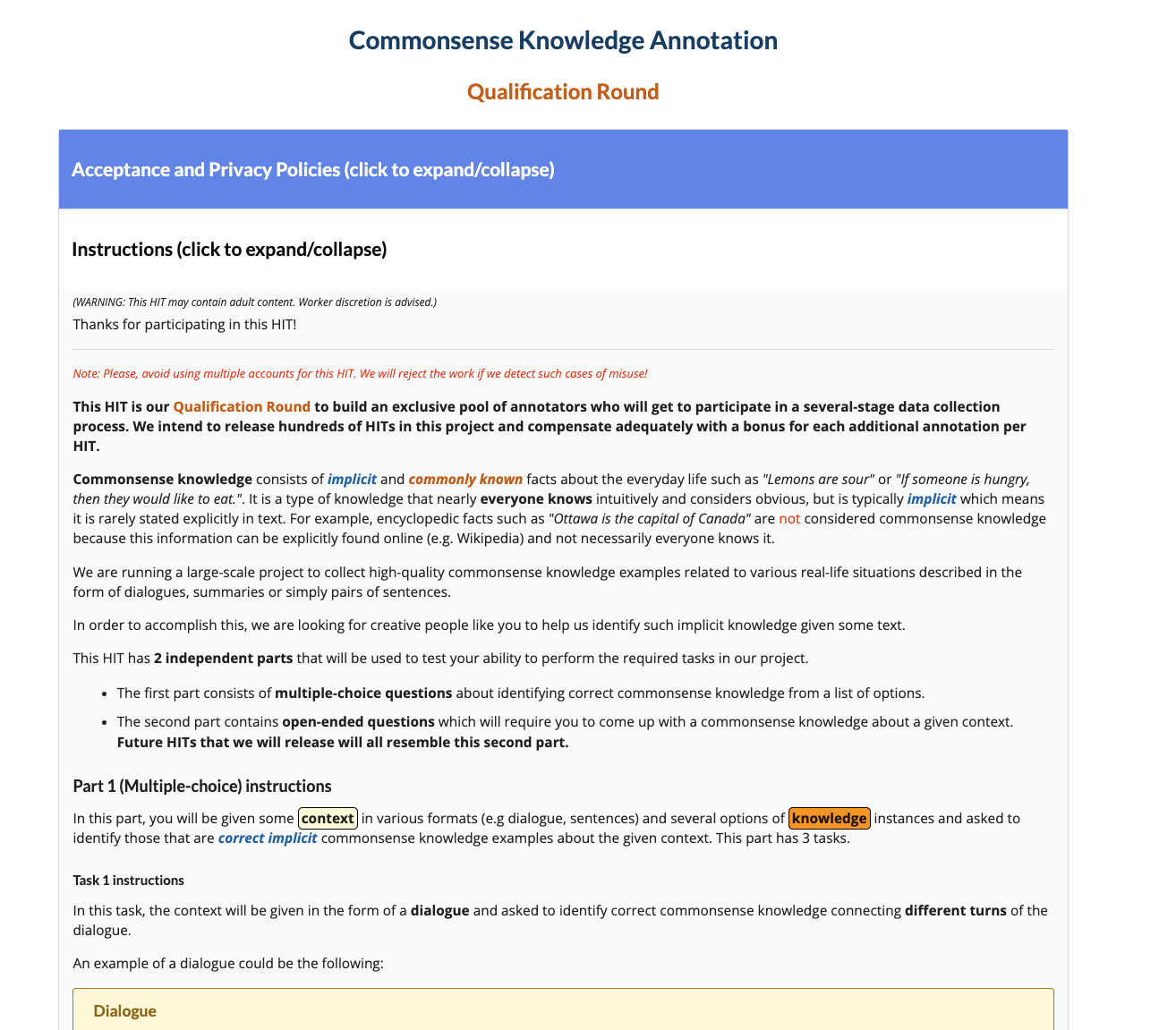}
    \caption{Mturk Instructions template for Qualification Stage}
    \label{fig:qual-instruct}
\end{figure*}

\begin{figure*}[h]
    \centering    
    \includegraphics[width=\textwidth]{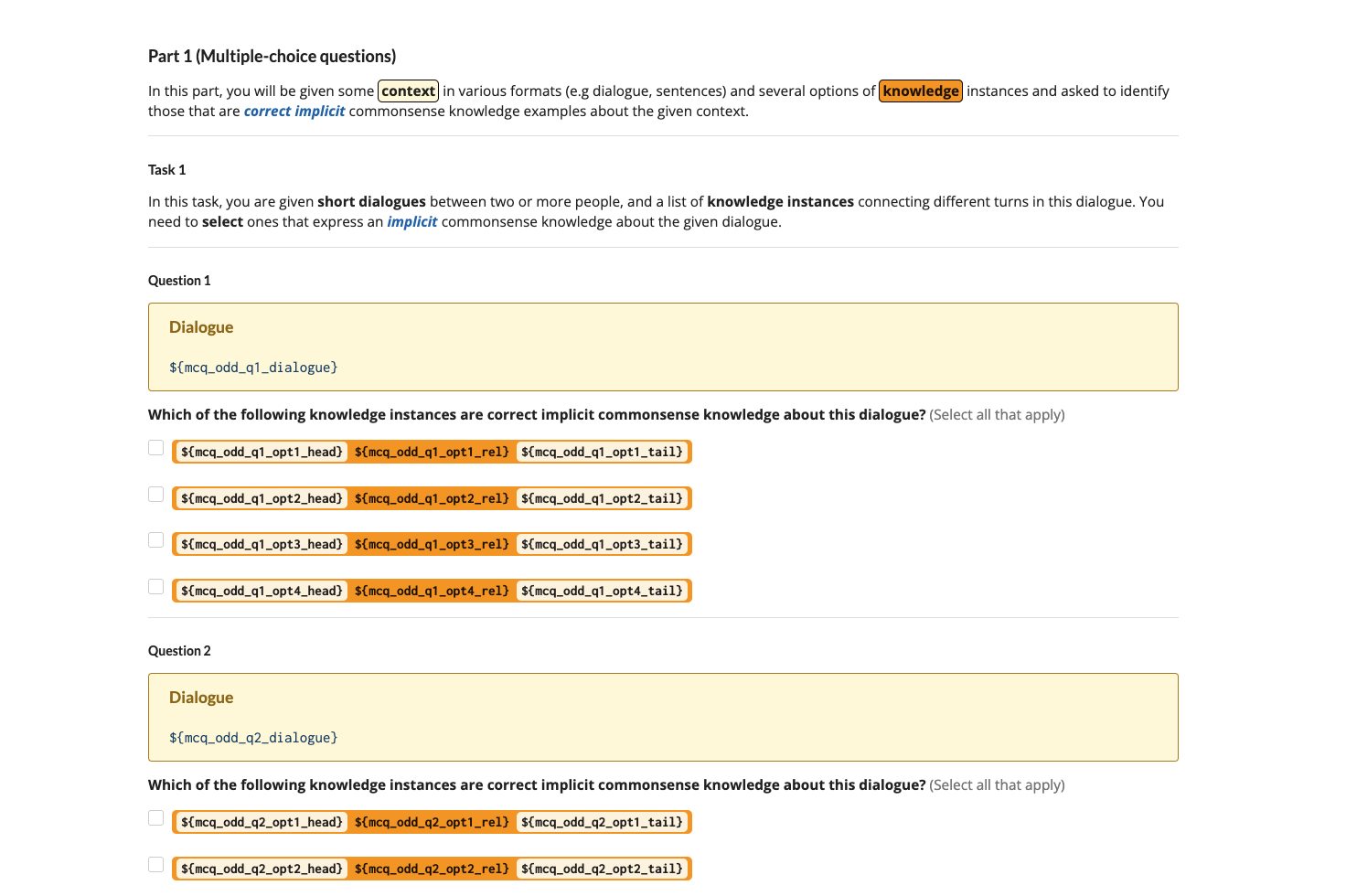}
    \caption{Mturk Task template for Qualification Stage Part 1 (MCQ)}
    \label{fig:qual-part1}
\end{figure*}

\begin{figure*}[h]
    \centering    
    \includegraphics[width=\textwidth]{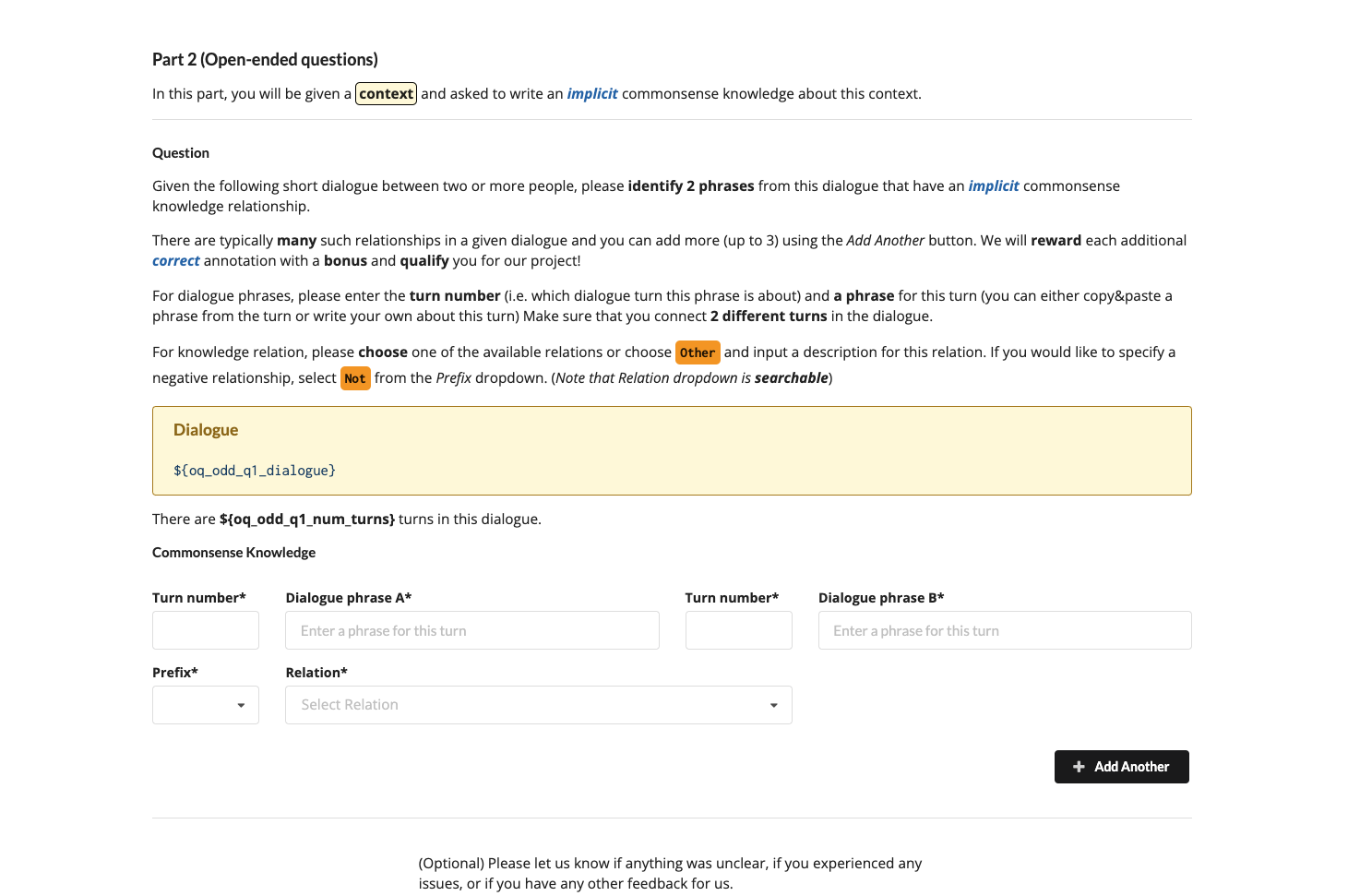}
    \caption{Mturk Task template for Qualification Stage Part 2 (Open-ended)}
    \label{fig:qual-part2}
\end{figure*}

This stage includes six multiple-choice questions (2 per task for dialogue, dialogue summarization and machine translation tasks) about identifying the correct implicit commonsense knowledge in a given context and one open-ended question that simulates the CKA stage for dialogue task. We compensate workers $\$2$ per HIT for this stage. Instruction and task templates can be found in Figures \ref{fig:qual-instruct}, \ref{fig:qual-part1}, \ref{fig:qual-part2}.

\subsection{Commonsense Knowledge Annotation and Validation}
\label{sec:app-cka}
\begin{figure*}[h]
    \centering    
    \includegraphics[width=\textwidth]{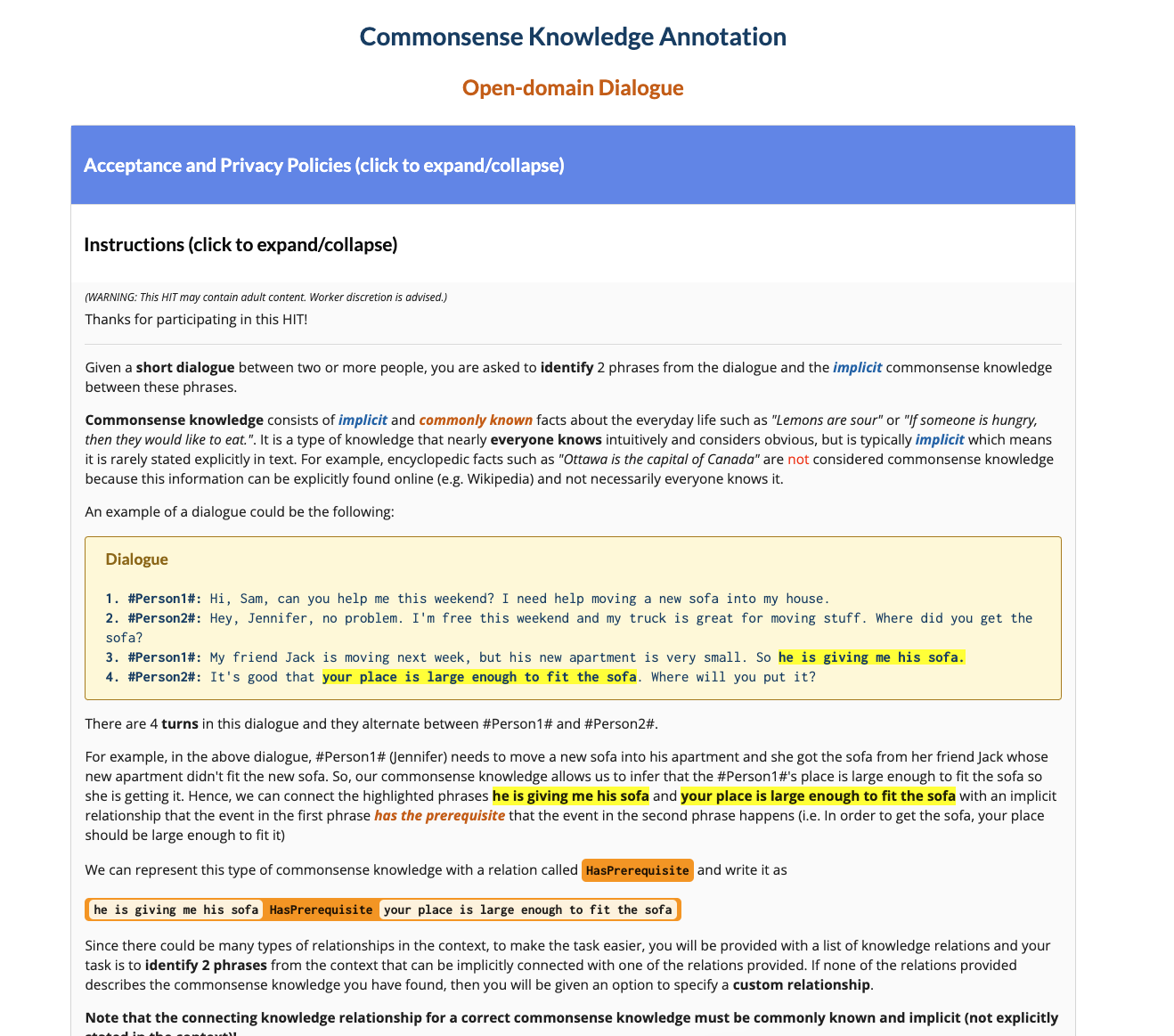}
    \caption{Mturk Instructions template for Dialogue CKA stage}
    \label{fig:odd-cka-instruct}
\end{figure*}

\begin{figure*}[h]
    \centering    
    \includegraphics[width=\textwidth]{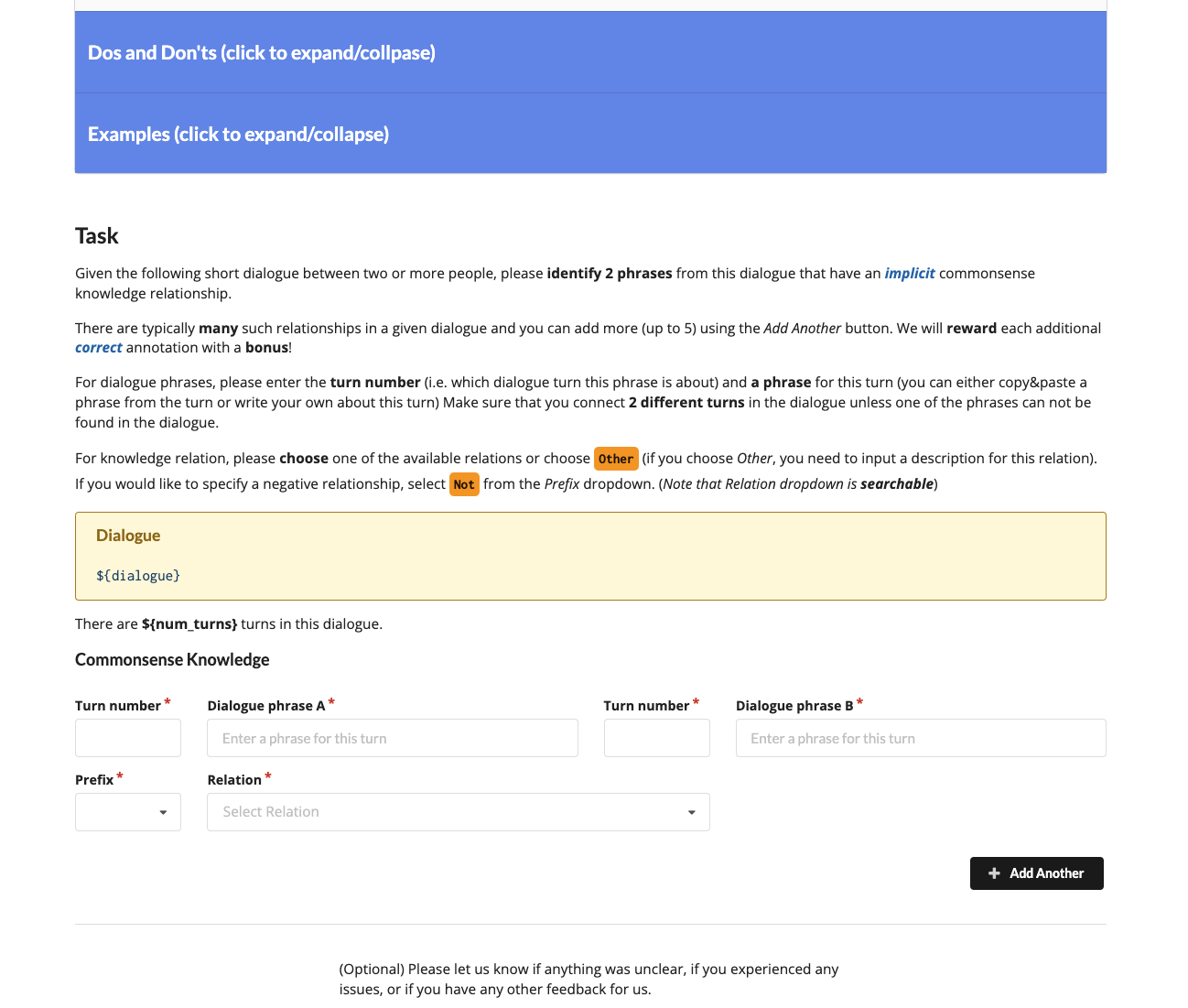}
    \caption{Mturk Task template for Dialogue CKA stage}
    \label{fig:odd-cka-task}
\end{figure*}

\begin{figure*}[h]
    \centering    
    \includegraphics[width=\textwidth]{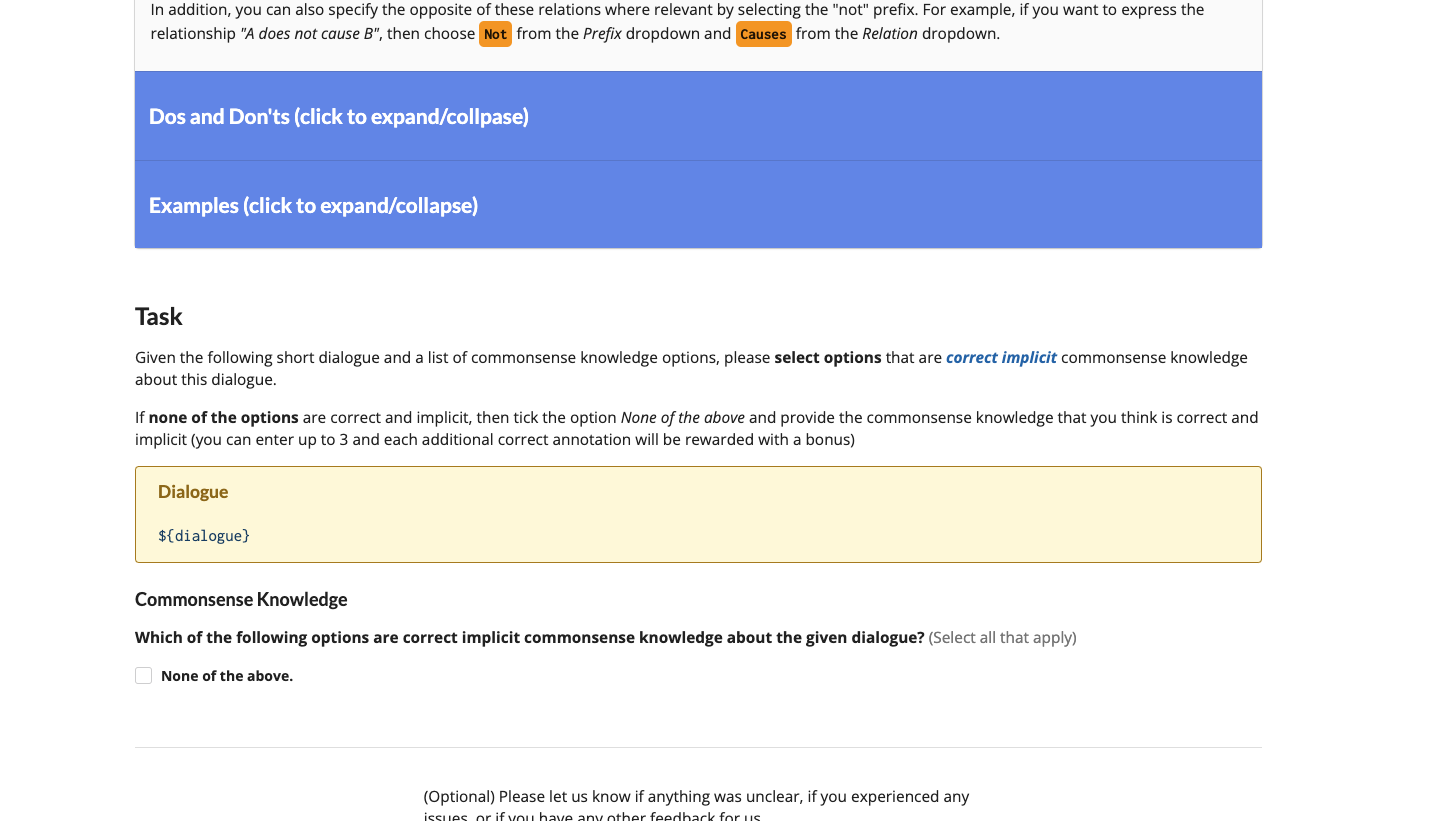}
    \caption{Mturk Instructions template for Dialogue CKV stage}
    \label{fig:odd-ckv}
\end{figure*}

\begin{figure*}[h]
    \centering    
    \includegraphics[width=\textwidth]{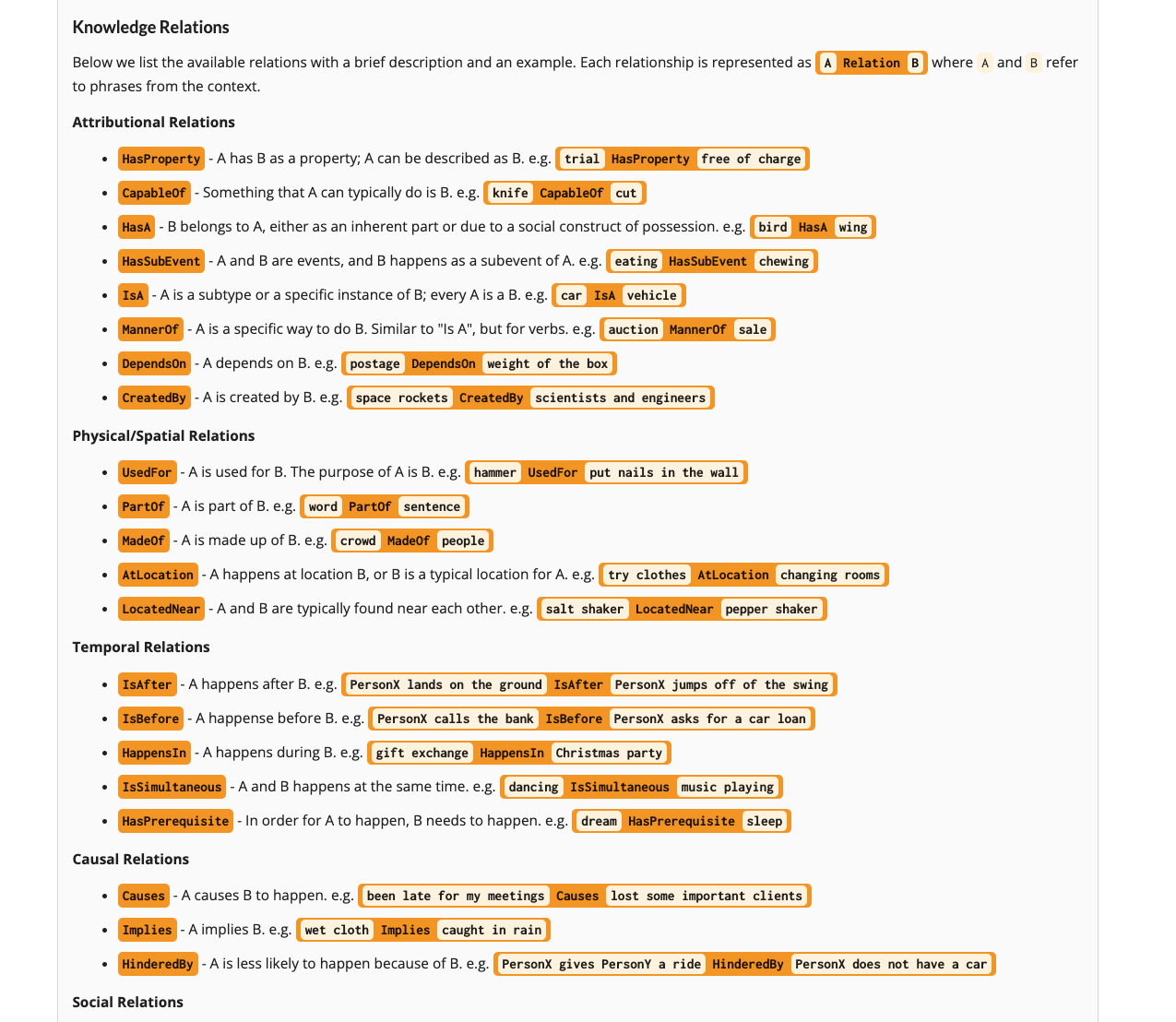}
    \caption{Mturk Knowledge Relations Section for Dialogue CKA stage}
    \label{fig:odd-cka-kg}
\end{figure*}

In the first stage of the pipeline, we ask annotators to identify commonsense knowledge in a given context for a task. They are asked to either select phrases from the context or write their own phrases about a context and connect these phrases with one of the commonsense knowledge relations provided (Table \ref{tab:commonsense_relations}). If none of the relations describe the commonsense knowledge they would like to express, they are given the option to specify a custom relationship (\ie \textit{Other} relation). However, annotators, most of the time tended to choose one of the available relations. In addition, annotators were incentivized to specify multiple commonsense knowledge relationships with an added bonus. Depending on the difficulty of the task, we pay workers $\$0.5-\$1$ per HIT in the annotation stage and $\$0.2-\$0.5$ per HIT in the validation stage. Instruction and task templates (for Dialogue task) can be found in Figures \ref{fig:odd-cka-instruct}, \ref{fig:odd-cka-kg}, \ref{fig:odd-cka-task}, \ref{fig:odd-ckv}. Templates for other tasks are similar.

\subsection{Winograd Schema Generation and Validation}
\label{sec:app-wsg}
\begin{figure*}[h]
    \centering    
    \includegraphics[width=\textwidth]{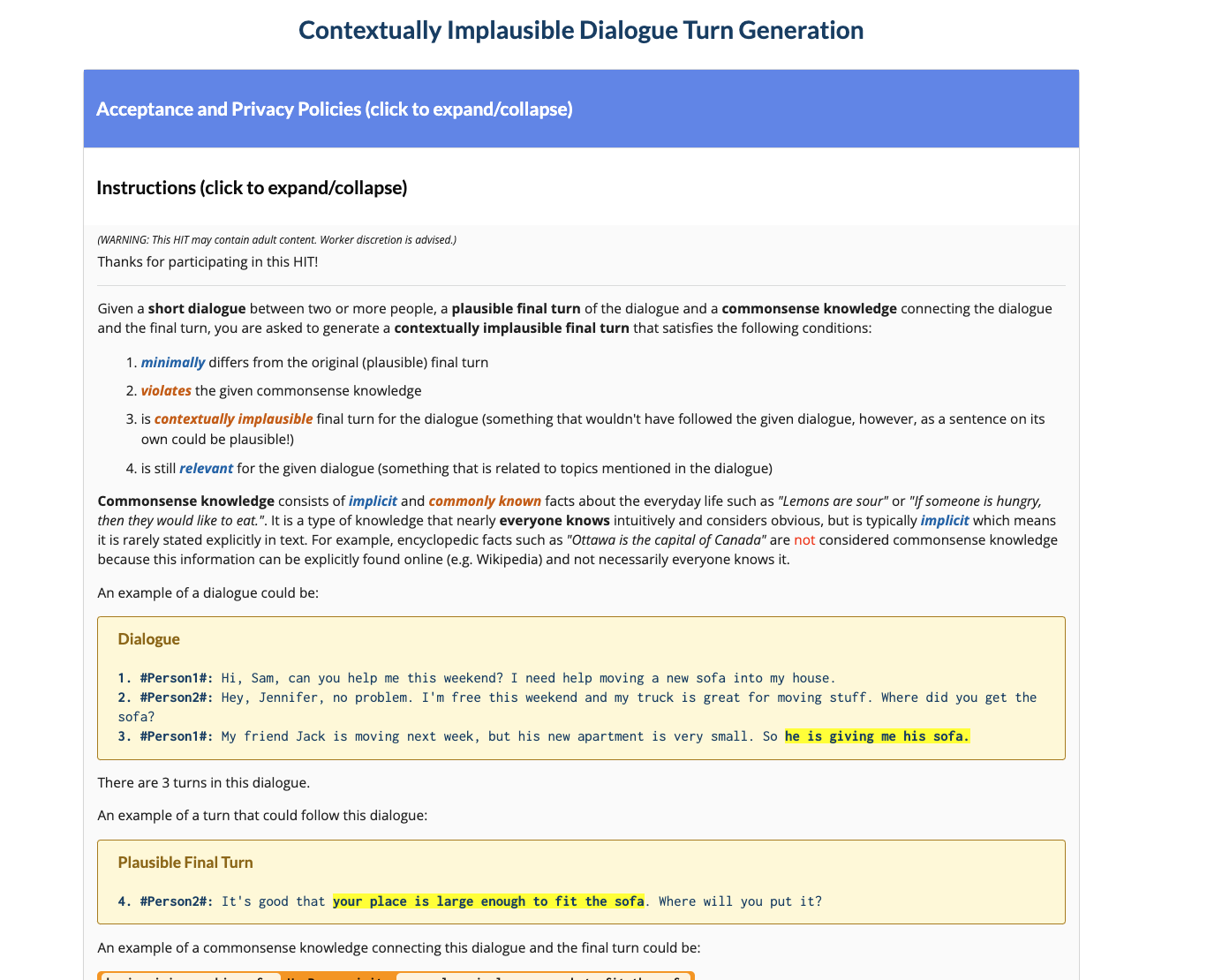}
    \caption{Mturk Instructions template for Dialogue WSG stage}
    \label{fig:odd-wsg-instruct}
\end{figure*}

\begin{figure*}[h]
    \centering    
    \includegraphics[width=\textwidth]{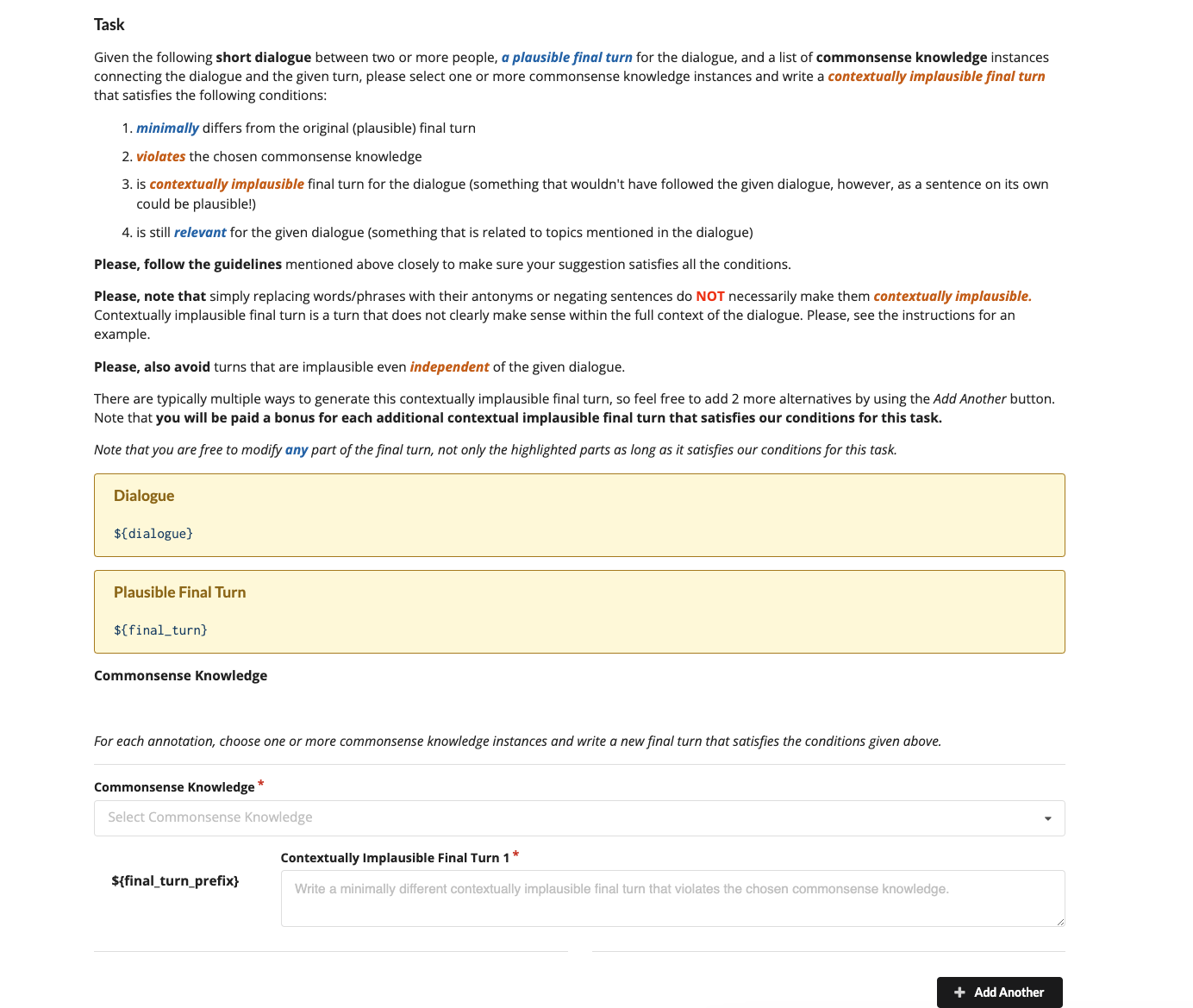}
    \caption{Mturk Task template for Dialogue WSG stage}
    \label{fig:odd-wsg-task}
\end{figure*}

\begin{figure*}[h]
    \centering    
    \includegraphics[width=\textwidth]{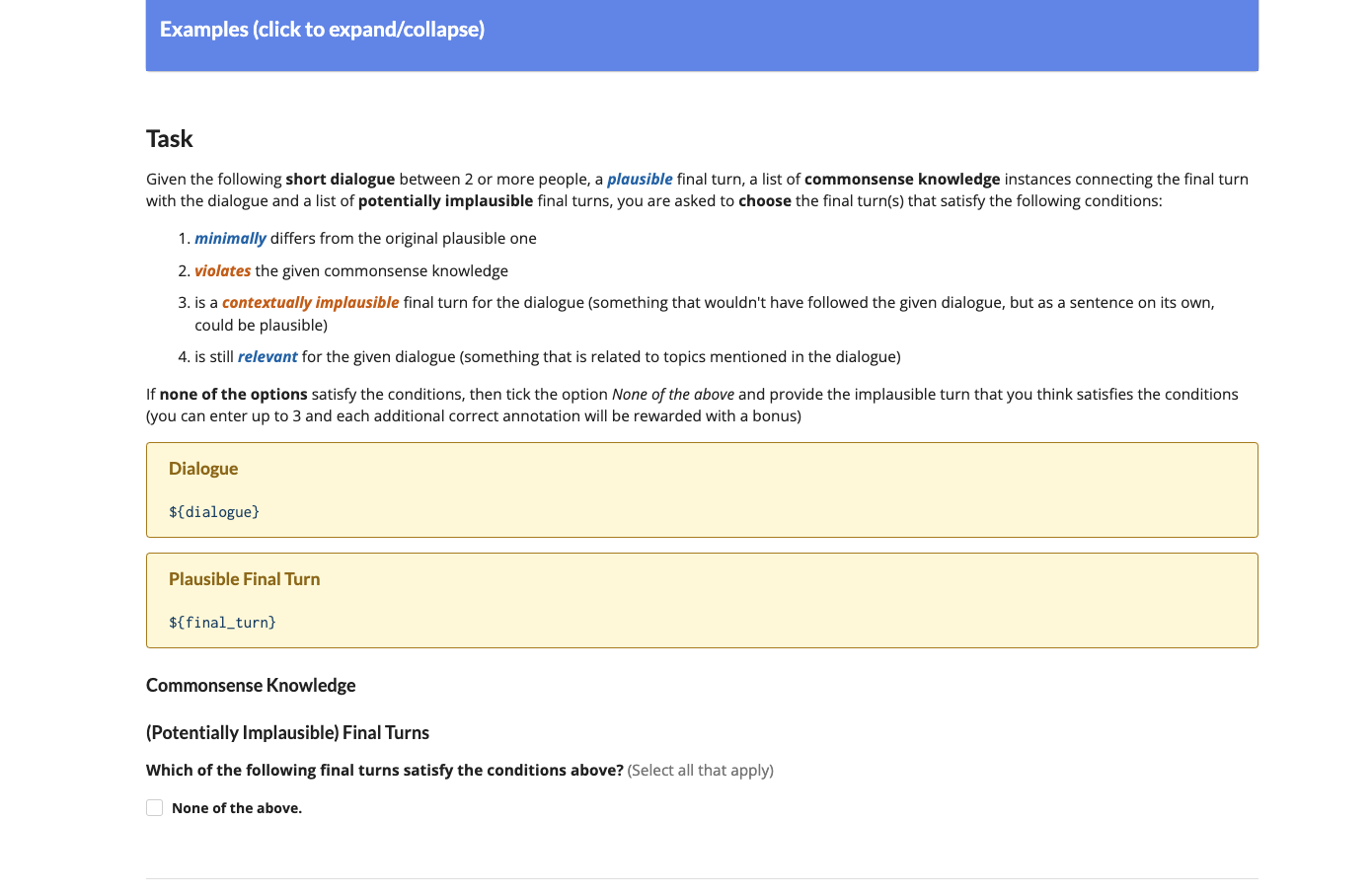}
    \caption{Mturk Instructions template for Dialogue WSV stage}
    \label{fig:odd-wsv-task}
\end{figure*}

The second main stage of our data collection pipeline is the generation of Winograd schemas using the commonsense knowledge annotated in the CKA stage. As mentioned before, for a given context and a piece of commonsense knowledge, we ask annotators to generate alternative targets that satisfy four conditions: (1) the new target is minimally different from the original one, (2) the new target violates the given commonsense knowledge, (3) new target is incorrect for the given task, and (4) new target is contextually relevant. The incorrectness is defined per task (See Appendix \ref{sec:app-tasks}). {As an example, we consider the following instance from the Safety Detection task:}
\begin{example}
    \textit{Scenario}: If you are being chased by a hungry animal\\
    \textit{Action}: get in a building and close the door\\
    \textit{Knowledge:} \\
    building \texttt{\textbf{HasProperty}} safe spot \\
    close the door \texttt{\textbf{CapableOf}} blocking animal
    \textit{Label:} safe\\
\end{example}
\noindent {Given the example above, our goal is to modify the target (\ie, action) such that it satisfies the conditions above. Here we provide some possible modifications and show whether the resulting target satisfies the conditions:
\begin{itemize}
    \item \textbf{find a spot where the animal can get to you}: This target satisfies all conditions except the (1) as it significantly modifies the original target.
    \item \textbf{get a gun and shoot the animal}: While this target is a good one, it does not really violate the given commonsense knowledge. It is also not minimally different from the original one.
    \item \textbf{get in a building and lock the door}: This target is not valid, because the action is still safe (in fact, even safer), so condition (3) is violated.
    \item \textbf{get in an alien spaceship and close this chapter of your life}: While this target seems to satisfy most conditions, it is largely out of context, so condition (4) is violated.
    \item \textbf{find a building and close your eyes}: This target satisfies all the conditions above.
\end{itemize}
}
\noindent Depending on the difficulty of the task, workers were paid $\$0.5-\$0.8$ per HIT in the generation stage and $\$0.2-\$0.5$ per HIT in the validation stage. Instructions and task templates (for Dialogue task) can be found in Figures \ref{fig:odd-wsg-instruct}, \ref{fig:odd-wsg-task}, \ref{fig:odd-wsv-task}. Templates for other tasks are similar.

\subsubsection{Instructions} 
\label{sec:app-instructions}

\paragraph{Minimal Change Rules}

During the WSG step of the annotation pipeline, the generated alternative sentence has to be deceptively close to (hard to differentiate from) the original sentence, but opposite of it in terms of commonsense knowledge or label. 
Thus, we asked annotators to follow a set of rules, that we report here (from the Safety Detection task):
\begin{itemize}
    \item You are allowed to change up to 5 words in the action. Note that you can also alternatively swap the existing words in place instead of replacing them as long as the resulting action satisfies the conditions above.
    \item You should avoid simply negating the word in the action unless that is the only way to achieve the goal. The goal in this task is not to achieve the minimal difference, but to produce semantically very close action that however flips the safety value. So, if you can creatively change 2 or 3 dependent words to achieve this, then go for it rather than changing one word such as adding/removing ``not''.
    \item Your change should directly target the given commonsense knowledge such that, semantically, the resulting action differs from the original one with respect to this knowledge.
\end{itemize}

\section{Tasks}
\label{sec:app-tasks}
For each task we have included in the benchmark, we define a common terminology that can be applied to other tasks in the future. \textit{Context} is defined as the unchanged part of the given example (\ie, the part that is not perturbed in the WSG stage) and the \textit{target} as the candidate for Winograd-style perturbation.
Note that the \textit{target} is not necessarily always the typical output of the model for a given task. For example, in classification tasks, the output of the model is binary while the \textit{target} is assigned to one of the inputs. For \textit{context} and \textit{target} assignments for each task, please refer to Table \ref{tab:benchmark_stats_2}.

\begin{table*}[t]
\small 
\resizebox{\linewidth}{!}{
    \centering
    \begin{tabular}{lcccccc}
    \toprule
   \textbf{Task} & \textbf{Context} & \textbf{Target} & \textbf{CKA/CKV} & \textbf{WSG/WSV} & \# \textbf{Contexts} & \# \textbf{Examples} \\ 
    \midrule
    Dialogue & dialogue & response & - & \checkmark & 1,169 & 3,548 \\
    Dialogue Summarization & dialogue & summary & \checkmark & \checkmark & 453 & 1,805 \\
    Machine Translation (zh-en) & sentence & translation & \checkmark & - & 600 & 1200 \\
    Machine Translation (en-de) & sentence & translation & \checkmark & - & 500 & 1000 \\
    Machine Translation (en-fr) & sentence & translation & \checkmark & - & 500 & 1000 \\
    Machine Translation (en-ru) & sentence & translation & \checkmark & - & 500 & 1000 \\
    Intent Detection & news headline & intent & \checkmark & \checkmark & 589 & 2,440 \\
    Stance Classification & belief/argument & argument/belief & - & \checkmark & 397 & 1,722 \\
    Safety Detection & scenario & action & \checkmark & \checkmark & 366 & 2,826 \\
   \midrule
   Total & & & & & 5,074 & 16,541 \\
   \bottomrule
\end{tabular}}
    \caption{Overview of the benchmark. CKA/CKV stands for Commonsense Knowledge Annotation and Validation stages, WSG/WSV stands for Winograd Schema Generation and Validation stages, respectively. Stages are skipped (-) for tasks that already have the necessary data available.}
    \label{tab:benchmark_stats_2}
\end{table*}

\subsection{Machine Translation (MT)}
Machine translation is one of the oldest sequence-to-sequence real-world NLP tasks where given input in the source language, a system is expected to output the translation in the target language.
In this task, we define the input in the source language as \textit{context} and the output translation as the candidate \textit{target} for perturbation. As both datasets we use include Winograd schemas, we skip the WSG and WSV stages for this task.  
In the CKA step for the Chinese-English dataset, annotators are given the target translations and asked to identify the commonsense knowledge violated in the incorrect translation. For the Wino-X dataset, although the target translation is in different languages, the underlying pronoun resolution task allows us to identify the commonsense knowledge from the source sentence alone. Leveraging this fact, we employ the same English-speaking pool of qualified workers for the CKA stage on this task as well. More specifically, annotators are given the source sentence (in English) and asked to identify the commonsense knowledge that allows us to infer the antecedent of the pronoun ``it'' in the sentence.
Here is an example of this task:
\begin{example}
\textbf{Example - Machine Translation:}\\
\textit{Sentence (English):} 
Bob would rather fill his emergency fund using his mobile instead of the bank because it was handy.\\
\textit{Translation (French):} 
Bob préférait remplir son fonds d'urgence en utilisant son mobile plutôt qu'avec la banque car elle était à portée de main.\\
\textit{Knowledge:} \\
mobile \texttt{\textbf{CapableOf}} store emergency funds\\
\textit{Label:} not correct (No)
\end{example}
We use the following prompt templates:
\begin{highlight}
\textbf{Prompt - Machine Translation: }\\
You are a helpful assistant for translation from \{source\_lang\} to \{target\_lang\}. Given a sentence in \{source\_lang\} and its translation in \{target\_lang\}, answer whether the translation is correct. Answer only Yes or No.\\
Example 1:\\
Sentence (\{source\_lang\}): \{sentence\}\\
Translation (\{target\_lang\}): \{translation\}\\
Answer:\{answer\} \\
Example 2:\\
...
\end{highlight}
\begin{highlight}
\textbf{Prompt - Machine Translation - CoT: }\\
You are a helpful assistant for translation from \{source\_lang\} to \{target\_lang\}. Given a sentence in \{source\_lang\} and its translation in \{target\_lang\}, answer whether the translation is correct. Let's work this out in a step-by-step way to be sure that we have the right answer. Then provide your final answer within the tags, <Answer>Yes/No</Answer>\\
Example 1:\\
Sentence (\{source\_lang\}): \{sentence\}\\
Translation (\{target\_lang\}): \{translation\}\\
Answer:\{answer\} \\
Example 2:\\
...
\end{highlight}
\begin{highlight}
\textbf{Prompt - Machine Translation + Commonsense Knowledge: }\\
You are a helpful assistant for translation from \{source\_lang\} to \{target\_lang\}. Given a sentence in \{source\_lang\}, its translation in \{target\_lang\} and a relevant knowledge, answer whether the translation is correct. Answer only Yes or No.\\
Example 1:\\
Sentence (\{source\_lang\}): \{sentence\}\\
Translation (\{target\_lang\}): \{translation\}\\
Knowledge: \{knowledge\}\\
Answer:\{answer\} \\
Example 2:\\
...
\end{highlight}
The template for CoT+Commonsense Knowledge is similar.
We use the following example as a one-shot example for CoT prompt of this task:
\begin{highlight}
\textbf{Prompt - Machine Translation - CoT Example: }\\
Sentence (English):\\
The song took longer to sing than the ballad because it was more words.\\
Translation (German): Das Lied brauchte länger zum Singen als die Ballade, weil sie mehr Worte enthielt.\\
Answer:\\
Let's break down the sentence and its translation:\\
- The song: Das Lied
- took longer: brauchte länger
- to sing: zum Singen
- than: als
- the ballad: die Ballade
- because: weil
- it: sie
- was more words: mehr Worte enthielt (literally: contained more words)\\
The translation is almost correct, but the pronoun for 'it' should be 'es' instead of 'sie'. This is because 'it' should refer to the thing that has more words and since 'song' took longer to sing than 'ballad', 'song' should contain more words, and the correct pronoun for 'song' is 'es', not 'sie'. So the translation is not correct.\\
<Answer>No</Answer>\\
\end{highlight}

\subsection{Dialogue (DG)}
The underlying task in Dialogue is to generate a response given a dialogue history.
In this task, we define the \textit{context} as the dialogue history and the \textit{target} as the response to this context. We skip the CKA and CKV stages for this task as the dataset we use comes with expert annotated commonsense knowledge annotations. To ensure the richness of the context and the knowledge, we filter out dialogue contexts with less than 4 turns and commonsense knowledge annotations that do not connect different turns in the dialogue.  
In the WSG stage, we ask the annotators to rewrite the final response of the given dialogue such that it satisfies our conditions for Winograd schemas mentioned above where the \textit{incorrectness} is defined as \textit{implausibility}. Since in an open-domain dialogue, several answers are possible for a given dialogue history, we aim for generating answers that violate some commonsense knowledge about the dialogue and hence, are implausible. However, since most of the commonsense knowledge in dialogues are \textit{contextual}, violating this knowledge does not automatically make the response implausible, hence we explicitly enforce a separate condition to ensure the implausibility. For example, given the following dialogue \textit{A: where will you have your birthday party? B: oh it is at my uncle's house}, the contextual commonsense knowledge can be the fact that \textit{(parties, AtLocation, uncle's house)}. Consequently, possible Winograd schema generated by violation of this knowledge could be \textit{B: oh it is at my friend's house}. However, this is not a correct Winograd schema for this task as it is a perfectly fine response to the dialogue. The implausible response here should target the more general commonsense knowledge that ``parties happen at people's houses''. In addition, we also ask annotators to avoid generating examples that are implausible independent of the dialogue context to make sure generations are not too easy for models to guess even in the absence of context.
Here is an example of this task:
\begin{example}
\textbf{Example - Dialogue:}\\
\textit{Dialogue}:
A: Good morning, sir. Is there anything I can do for you?\\
B: I would like to buy two bottles of brandy.\\
A: How about this one? It's the special local product.\\
B: Can I buy these tax free?\\
\textit{Response:} A: Yes . This is not a duty-free shop.
\textit{Knowledge:} \\
duty-free shop \texttt{\textbf{Implies}} tax free\\
\textit{Label:} Not plausible (No)
\end{example}
We use the following prompt templates:
\begin{highlight}
\textbf{Prompt - Dialogue: }\\
You are a helpful assistant for dialogue understanding. Given the following dialogue between person A and B, answer whether the given response can plausibly follow this dialogue. Answer only 'Yes' or 'No'.\\
Example 1:\\
Dialogue: \{dialogue\}\\
Response: \{response\}\\
Answer:\{answer\}\\
Example 2:\\
...
\end{highlight}
\begin{highlight}
\textbf{Prompt - Dialogue - CoT: }\\
You are a helpful assistant for dialogue understanding. Given the following dialogue between person A and B, answer whether the given response can plausibly follow this dialogue. Let's work this out in a step-by-step way to be sure that we have the right answer. Then provide your final answer within the tags, <Answer>Yes/No</Answer>.\\
Example 1:\\
Dialogue:\{dialogue\}\\
Response:\{response\}\\
Answer:\{answer\}\\
Example 2:\\
...
\end{highlight}
\begin{highlight}
\textbf{Prompt - Dialogue + Commonsense Knowledge: }\\
You are a helpful assistant for dialogue understanding. Given the following dialogue between person A and B and a relevant knowledge about this dialogue, answer whether the given response can plausibly follow this dialogue. Answer only 'Yes' or 'No'.\\
Example 1:\\
Dialogue: \{dialogue\}\\
Response: \{response\}\\
Knowledge: \{knowledge\}\\
Answer:\{answer\}\\
Example 2:\\
...
\end{highlight}
The template for CoT+Commonsense Knowledge is similar.
We use the following example as a one-shot sample for CoT prompt of this task:
\begin{highlight}
\textbf{Prompt - Dialogue - CoT Example:}\\
Dialogue:\\
A: ( Before Christmas Party ) Are you ready for the Christmas party tonight\\
B: Almost. I have to get dressed. It's a formal party and I have special party make up!\\
A: Use this lipstick and it will make your lips shine!\\
Response: \\
B: Great! Uh, remember that there's a rocket launch, too. We all have to bring a gift.\\
Answer:\\
Step 1: Identify the main topics in the dialogue.\\
- Christmas party
- Getting dressed
- Formal party
- Special party make up
- Lipstick\\
Step 2: Analyze the response.\\
- The response mentions a rocket launch, which is not related to the main topics in the dialogue.
- The response mentions bringing a gift, which could be related to the Christmas party.\\
Step 3: Determine if the response can plausibly follow the dialogue.\\
The mention of a rocket launch seems out of context and unrelated to the dialogue. In addition the second part of the response mentions an obligation to bring a gift which wouldn't follow the first part as rocket launch event typically does not require to bring a gift. A plausible event would be a gift exchange event. 
So the response does not plausibly follow the dialogue.
Final Answer: <Answer>No</Answer>
\end{highlight}

\subsection{Intent Detection (ID)}
In this task, we treat the text of the author as the \textit{context} and the intent as the \textit{target} for perturbation. We use the headline as our \textit{context} and the writer intent as the \textit{target} for the dataset we use for this task. The full pipeline is applied to this dataset and as a preprocessing step,  we filter out examples with too short headlines or intents.
Here is an example of sample for this task:
\begin{example}
\textbf{Example - Intent:}\\
\textit{Headline}: Hospitals on lockdown as first COVID vaccine patients start eating other patients.\\
\textit{Intent:} a hospital is on lockdown due to covid patients kissing other patients after getting the vaccine.\\
\textit{Knowledge:} \\
COVID vaccine \texttt{\textbf{Causes}} eating other patients\\
\end{example}
We use the following prompt templates:
\begin{highlight}
\textbf{Prompt - Intent: }\\
You are a helpful assistant for intent classification. Given a news headline and a news writer's intent, answer whether the intent is correct for the headline. Answer only Yes or No.\\
Example 1:\\
Headline:\{headline\}\\
Intent:\{intent\}\\
Answer:\{answer\}\\
Example 2:\\
...
\end{highlight}
\begin{highlight}
\textbf{Prompt - Intent - CoT: }\\
You are a helpful assistant for intent classification. Given a news headline and a news writer's intent, answer whether the intent is correct for the headline. Let's work this out in a step-by-step way to be sure that we have the right answer. Then provide your final answer within the tags, <Answer>Yes/No</Answer>.\\
Example 1:\\
Headline:\{headline\}\\
Intent:\{intent\}\\
Answer:\{answer\}\\
Example 2:\\
...
\end{highlight}
\begin{highlight}
\textbf{Prompt - Intent + Commonsense Knowledge: }\\
You are a helpful assistant for intent classification. Given a news headline, a news writer's intent and a relevant knowledge, answer whether the intent is correct for the headline. Answer only Yes or No.\\
Example 1:\\
Headline:\{headline\}\\
Intent:\{intent\}\\
Knowledge:\{knowledge\}\\
Answer:\{answer\}\\
Example 2:\\
...
\end{highlight}
The template for CoT+Commonsense Knowledge is similar.
We use the following example as one-shot sample for CoT prompt of this task:
\begin{highlight}
\textbf{Prompt - Intent - CoT Example: }\\
Headline:\\
Authorities will delay vaccines in Andalusia. They bought millions of syringes that will not work to distribute the COVID-19 vaccine \\
Intent:\\
the vaccine requires specific needles to apply \\
Answer:\\
Step 1: Analyze the headline\\
The headline states that authorities in Andalusia will delay vaccines because they bought millions of syringes that will not work to distribute the COVID-19 vaccine. This shows that there is a incompatibility between the bought syringes and syringes required for the vaccine.\\
Step 2: Analyze the intent\\
The intent states that the vaccine requires specific needles to apply. This means standard syringes might not be suitable.\\
Step 3: Compare the headline and intent\\
The headline implies that the syringes purchased are not suitable for distributing the COVID-19 vaccine, which aligns with the intent stating that specific needles are required to apply the vaccine.\\
So, the given intent is the correct one for this headline.
Final Answer: <Answer>Yes</Answer>
\end{highlight}

\subsection{Stance Classification (SC)}
Stance classification is a task where given a belief and an argument, the stance of the argument is predicted. Since the dataset we chose for this task is already annotated with commonsense knowledge, we skip the first two steps of the pipeline -- CKA and CKV. Similarly to the other selected tasks, we filter the examples with short sentences. 
to give more degrees of freedom to crowdsource workers for the WSG step. Moreover, in this task, the \textit{context} and the \textit{target} are dynamically chosen --- we treat both sentences (the belief and the argument) equally --- allowing workers to select the one to modify.
Here is an example of this task:
\begin{example}
\textbf{Example - Stance:}\\
\textit{Belief}: Cosmetic surgery should not be banned.\\
\textit{Argument}: Cosmetic surgery is not worth the risk\\
\textit{Knowledge:} \\
risky \texttt{\textbf{UsedFor}} human body\\
\textit{Label}: Counter (No)
\end{example}
We use the following prompt templates:
\begin{highlight}
\textbf{Prompt - Stance: }\\
You are a helpful assistant for stance classification. Given a belief and an argument, answer whether the argument supports the belief. Answer only Yes or No.\\
Example 1:\\
Belief: \{belief\}\\
Argument:\{argument\}\\
Answer:\{answer\}\\
Example 2:\\
...
\end{highlight}
\begin{highlight}
\textbf{Prompt - Stance - CoT: }\\
You are a helpful assistant for stance classification. Given a belief and an argument, answer whether the argument supports the belief. Let's work this out in a step-by-step way to be sure that we have the right answer. Then provide your final answer within the tags, <Answer>Yes/No</Answer>.\\
Example 1:\\
Belief: \{belief\}\\
Argument:\{argument\}\\
Answer:\{answer\}\\
Example 2:\\
...
\end{highlight}
\begin{highlight}
\textbf{Prompt - Stance + Commonsense Knowledge: }\\
You are a helpful assistant for stance classification. Given a belief, an argument and a relevant knowledge, answer whether the argument supports the belief. Answer only Yes or No.\\
Example 1:\\
Belief: \{belief\}\\
Argument:\{argument\}\\
Knowledge:\{knowledge\}\\
Answer:\{answer\}\\
Example 2:\\
...
\end{highlight}
The template for Cot+Commonsense Knowledge is similar.
We use the following example as a one-shot sample for CoT prompt of this task:
\begin{highlight}
\textbf{Prompt - Stance - CoT Example: }\\
Belief:\\
Cosmetic surgery should be allowed.\\
Argument:\\
Cosmetic surgery is not worth the risk.\\
Answer:\\
Step 1: Analyze the belief\\
According to the belief, cosmetic surgery should be allowed which might mean that it is not risky.\\
Step 2: Analyze the argument\\
The argument states that cosmetic surgery is not worth the risk, so it assumes that there are risks involved, but it is not worth to do while taking the risk.\\
Step 3: Compare the belief and argument\\
The belief supports cosmetic surgery, while the argument opposes it due to the risks involved.
Final Answer: <Answer>No</Answer>
\end{highlight}

\subsection{Safety Detection (SD)}
The underlying task is to determine the safe action given a real-life scenario and a list of actions comprised of safe and unsafe actions. In this task, we define the \textit{context} as the given scenario and \textit{target} as the possible action. 
We apply the full pipeline for this task where in the WSG stage, annotators are asked to either minimally modify the safe action to make it unsafe or vice-versa.
Here is an example of sample for this task:
\begin{example}
\textbf{Example - Safety:}\\
\textit{Scenario}: When heating up leftovers in the microwave\\
\textit{Action}: wait for the plate to cool off before you grab it\\
\textit{Knowledge:} \\
wait for the plate to cool off before you grab it \texttt{\textbf{xIntent}} avoid getting burnt\\
\textit{Label}: safe (Yes)
\end{example}
We use the following prompt templates:
\begin{highlight}
\textbf{Prompt - Safety: }\\
You are a helpful assistant for safety detection. Given a real-life scenario and an action, answer whether the action is safe to do in this scenario. Answer only Yes or No.\\
Example 1:\\
Scenario:\{scenario\}\\
Action:\{action\}\\
Answer:\{answer\}\\
Example 2:\\
...
\end{highlight}
\begin{highlight}
\textbf{Prompt - Safety - CoT: }\\
You are a helpful assistant for safety detection. Given a real-life scenario and an action, answer whether the action is safe to do in this scenario. Let's work this out in a step-by-step way to be sure that we have the right answer. Then provide your final answer within the tags, <Answer>Yes/No</Answer>\\
Example 1:\\
Scenario:\{scenario\}\\
Action:\{action\}\\
Answer:\{answer\}\\
Example 2:\\
...
\end{highlight}
\begin{highlight}
\textbf{Prompt - Safety + Commonsense Knowledge: }\\
You are a helpful assistant for safety detection. Given a real-life scenario, an action and a relevant knowledge, answer whether the action is safe to do in this scenario. Answer only Yes or No.\\
Example 1:\\
Scenario:\{scenario\}\\
Action:\{action\}\\
Knowledge:\{knowledge\}\\
Answer:\{answer\}\\
Example 2:\\
...
\end{highlight}
The template for CoT+Commonsense Knowledge is similar. 
We use the following example as a one-shot sample for CoT prompt of this task:
\begin{highlight}
\textbf{Prompt - Safety - CoT Example: }\\
Scenario:\\
If you're on a hike in the woods and you see a colorful mushroom.\\
Action:\\
do not eat it.\\
Answer:\\
Step 1: Identify the potential risks in the scenario\\
- Consuming an unknown and hence potentially harmful mushroom found in the woods.
Step 2: Evaluate the action\\
The action is to not eat the colorful mushroom.\\
Step 3: Determine if the action is safe in this scenario\\
By not eating the mushroom, the potential risk of consuming a poisonous or harmful mushroom is avoided. So, the action is safe to do in this scenario.\\
Final Answer: <Answer>Yes</Answer>
\end{highlight}

\subsection{Dialogue Summarization (DS)}
In dialogue summarization, given a dialogue context, the system is expected to output either an extractive or abstractive summary of the context. 
In this task, we designate the dialogue as the \textit{context} and the summary as the \textit{target} candidate for rewriting. The full data annotation pipeline is applied for this task where in the WSG stage, the incorrectness condition is satisfied if the rewritten summary does not correctly summarize the given dialogue. As a filtering phase, we discard examples with dialogues that have less than 4 turns or summaries that have less than 5 content words.
Here is an example of this task:
\begin{example}
\textbf{Example - Summarization:}\\
\textit{Dialogue}:\\
\#Person1\#: How may I help you?\\
\#Person2\#: I would like to return this book.\\
\#Person1\#: Is that all you need?\\
\#Person2\#: I also want to check out this video.\\
\#Person1\#: Do you have your library card?\\
\#Person2\#: Here it is.\\
\#Person1\#: If you damage the video, you will be fined.\\
\#Person2\#: I won't damage it.\\
\textit{Summary:}\\
\#Person1\# helps \#Person2\# to return a book and check out a video in the card-free, honor-system library.\\
\textit{Knowledge:} \\
check out a video in the library \texttt{\textbf{DependsOn}} have your library card\\
\textit{Label:} not correct (No)
\end{example}
We use the following prompt templates:
\begin{highlight}
\textbf{Prompt - Summarization:}\\
You are a helpful assistant for dialogue summarization. Given the following dialogue between \#Person1\# and \#Person2\#, answer whether the given summary correctly summarizes the dialogue. Answer only 'Yes' or 'No'.\\
Example 1:\\
Dialogue: \{dialogue\}\\
Summary:\{summary\}\\
Answer:\{answer\}\\
Example 2:\\
...
\end{highlight}
\begin{highlight}
\textbf{Prompt - Summarization - CoT:}\\
You are a helpful assistant for dialogue summarization. Given the following dialogue between \#Person1\# and \#Person2\#, answer whether the given summary correctly summarizes the dialogue. Let's work this out in a step-by-step way to be sure that we have the right answer. Then provide your final answer within the tags, <Answer>Yes/No</Answer>.\\
Example 1:\\
Dialogue:\{dialogue\}\\
Summary:\{summary\}\\
Answer:\{answer\}\\
Example 2:\\
...
\end{highlight}
\begin{highlight}
\textbf{Prompt - Summarization + Commonsense Knowledge:}\\
You are a helpful assistant for dialogue summarization. Given the following dialogue between \#Person1\# and \#Person2\# and a relevant knowledge, answer whether the given summary correctly summarizes the dialogue. Answer only 'Yes' or 'No'.\\
Example 1:\\
Dialogue: \{dialogue\}\\
Summary:\{summary\}\\
Knowledge:\{knowledge\}\\
Answer:\{answer\}\\
Example 2:\\
...
\end{highlight}
The template for CoT+Commonsense Knowledge is similar.
We use the following example as a one-shot sample for CoT prompt of this task:
\begin{highlight}
\textbf{Prompt - Summarization - CoT Example:}\\
Dialogue:\\
\#Person1\#: I'm going to New York for the first time, but I don't have a tour guide. Can you give me any suggestions?\\
\#Person2\#: There's a service called 'A friend in New York'. It's a personal tour guide service.\\
\#Person1\#: That's interesting. What does it do?\\
\#Person2\#: You give them your information by answering a questionnaire and they will create a perfect trip for you according to your budget.\\
\#Person1\#: Good. Where can I get the questionnaire?\\
\#Person2\#: You can easily download it from their website.\\
\#Person1\#: That's helpful! Thanks!\\
Summary:\\
\#Person1\# is going to New York for the first time. \#Person2\# suggests \#Person1\# use a personal tour guide service even though they won't know how to put together \#Person1\#'s trip plan.\\
Answer:\\
Step 1: Identify the main points in the dialogue.\\
- \#Person1\# is going to New York for the first time and needs suggestions.
- \#Person2\# suggests 'A friend in New York' service.
- The service creates a perfect trip based on a questionnaire.
- The questionnaire can be downloaded from their website.\\
Step 2: Compare the summary with the main points.\\
- The summary correctly mentions that \#Person1\# is going to New York for the first time.
- The summary mentions the personal tour guide service, but it incorrectly states that they won't know how to put together \#Person1\#'s trip plan because according the dialogue, the service can create a perfect trip based on the questionnaire.
Final Answer: <Answer>No</Answer>
\end{highlight}

\begin{table}[ht]
\scalebox{0.55}{
    \centering
    \begin{tabular}{lrrrrrr}
    \toprule
    \textbf{Model} & \multicolumn{6}{c}{{\textbf{CK Dimensions}}} \\
    \midrule
    & Attribution & Physical & Temporal & Causal & Social & Comparison \\
    \midrule
\textbf{Flan-Alpaca$^\clubsuit$} & 67.4 & 69.4 & 66.1 & 67.6 & 70.4 & 67.5 \\
\textbf{Flan-T5-11B$^\clubsuit$} & 75.5 & 76.6 & 76.5 & 75.9 & 77.1 & 79.2 \\
\textbf{LLaMa-33B$^\clubsuit$} & 42.2 & 42.4 & 44.2 & 42.0 & 42.4 & 43.5 \\
\textbf{Stable-Vicuna$^\clubsuit$} & 55.0 & 56.9 & 59.6 & 55.3 & 56.1 & 56.1 \\
\midrule
\textbf{BloomZ-7B} & 56.1 & 56.7 & 54.9 & 54.6 & 55.1 & 56.2 \\
\textbf{PaLM-1-540B} & 48.0 & 48.4 & 49.8 & 48.2 & 49.4 & 49.4 \\
\textbf{GPT-3.5} & 64.5 & 64.3 & 63.5 & 64.1 & 65.9 & 70.5 \\
\textbf{GPT-4} & 74.3 & 72.5 & 70.9 & 73.5 & 73.0 & 72.4 \\
\textbf{GPT-4-CoT} & 73.4 & 71.5 & 68.8 & 71.7 & 73.6 & 72.9 \\
   \bottomrule
    \end{tabular}}
    \caption{\textbf{Macro-F1} scores averaged across commonsense knowledge dimensions ($^\clubsuit$all tasks except for MT.). Temperature is set to $0.3$ and all results are averaged over three runs with different seeds.}
    \label{tab:app_results_commonsense_dimension}
\end{table}

\begin{table}[ht]
\centering
\scalebox{0.7}{
    \begin{tabular}{lrr}
    \toprule
    \textbf{Model} & \textbf{Oracle Knowledge} & \textbf{No Knowledge} \\
     \midrule
      \textbf{Flan-T5-11B$^\clubsuit$}  & 77.0 / 46.2 & 75.0 / 41.1\\ 
     \textbf{BloomZ-7B}  & 52.3 / 13.6 & 55.5 / 15.9\\ 
     \textbf{GPT-4}  & 74.5 / 47.5 & 72.9 / 45.4  \\
     \textbf{GPT-4-CoT}  & 76.5 / 52.8 &  72.2 / 46.3 \\
   \bottomrule
    \end{tabular}}
    \caption{\textbf{Macro-F1} / \textbf{Situational Accuracy} scores averaged over all tasks ($^\clubsuit$all tasks except MT), with and without providing commonsense knowledge in the prompt. Temperature is set to $0.3$ and all results are averaged over three runs with different seeds.}
    \label{tab:app_results_oracle}
\end{table}

\newpage
\section{Human Evaluation}
\label{sec:app-human-eval}
In this section, we provide more details on the human evaluation results. As we allow human evaluators to discuss cases of disagreement, the number of resolutions and the human performance before and after the discussion are of interest as well. In Table \ref{tab:further-human-results}, we report the percentage of resolved disagreements per task and the human results before and after discussion compared to the best-performant model which is GPT-4. Performance numbers for human scores before discussion are calculated by treating each annotator as a different prediction for each example and computing the performance over all predictions. If there is a clash and one annotator is correct and the other is not, then that example would receive a human score of $0.5$ for accuracy. After discussion, the annotators agree on the final prediction, so the agreed-upon label is the prediction. As we can see from the table, the human performance is already high before the disagreements are resolved, with GPT-4 only exceeding the individual human performance on stance classification. We note, however, that this is one of the tasks with the most initial disagreements. When discussed between both evaluators, however, they settle on labels and outperform GPT-4.
\paragraph{Analysis of disagreements}
{We observed different reasons for human disagreements, but the most common causes were: (i) instances requiring subtle factual knowledge that annotators may not recognize (see disagreement example below for Stance Classification), (ii) ambiguity and open-endedness (see disagreement example below for Safety Detection)}
\begin{example}
    \textbf{Disagreement Example - Stance:}\\
    \textit{Belief}: Entrapment is bad if done properly.\\
    \textit{Argument}: The bad parts of entrapment can be eliminated if the criminal can be given the opportunity to refuse.\\
    \textit{Answer:} Yes (supporting)\\
\end{example}
\noindent {Along with making commonsense inferences, the above example requires subtle factual knowledge about entrapment, namely that entrapment requires the criminal to not be aware that they are being entrapped. As a result, if the criminal is given the opportunity to refuse, then commonsense tells us that the entrapment is no longer being done properly since the criminal is aware. According to the belief statement, if the entrapment is no longer being done properly, then it is no longer bad (contrapositive of the belief statement). This is the statement made by the argument, which is why the belief is supported by the argument. Without the subtle factual knowledge about entrapment, however, this example is challenging for human raters.}
\begin{example}
    \textbf{Disagreement Example - Safety:}\\
    \textit{Scenario}: If your pets don't get along well\\
    \textit{Action}: let them play in a large room together\\
    \textit{Answer:} Yes (safe)\\
\end{example}
\noindent {The above example is ambiguous because the action could result in both a safe and unsafe scenario. The action statement specifies that the room is large, hinting that it provides enough space for the pets to be in the same room and that playing together may let them grow closer (a common way of introducing pets to each other). However, “pets don’t get along well” could also mean they attack each other and fight, and putting them in any room together might lead to an unsafe situation, regardless of their size. Without contextual information to clarify, the example remains ambiguous to human readers.}

\paragraph{Statistical Significance}
{We perform a Binomial Sign test on GPT-4 and Human evaluation results as our sample size is small. We find that for Machine Translation (en-ru, en-de and en-fr), Intent detection, Safety detection and Stance Classification tasks, results are statistically significant with a level of 0.01, for Dialogue and Summarization tasks with a level of 0.05. Only for Machine Translation (zh-en) task, we obtain a high p-value (0.57).}

\begin{table*}[t!]
\footnotesize
\scalebox{0.65}{
\centering
\begin{tabular}{lrrrrrrrrrrr}
\toprule 
   \multirow{2}{*}{\textbf{Models}}
 & \multicolumn{4}{c}{{\textbf{MT}}} 
 & \multicolumn{1}{c}{{\textbf{DG}}}
 & \multicolumn{1}{c}{{\textbf{DS}}}
 & \multicolumn{1}{c}{{\textbf{SC}}}
 & \multicolumn{1}{c}{{\textbf{SD}}}
 & \multicolumn{1}{c}{{\textbf{ID}}}
 & \multicolumn{1}{c}{{\textbf{CROW}}}
 & \multicolumn{1}{c}{{\textbf{CROW}}}  \\
 
 & \multicolumn{1}{c}{Zh-En}& \multicolumn{1}{c}{En-Fr}  & \multicolumn{1}{c}{En-De} & \multicolumn{1}{c}{En-Ru} & & & & & & \multicolumn{1}{c}{\textbf{Score (-MT)}} & \multicolumn{1}{c}{\textbf{Score}} \\
 \midrule
\textbf{GPT-4} & 75.9 / 57.9 & 54.5 / 21.5 & 54.4 / 20.5 & 54.1 / 19.7 & 72.4 / 46.5 & 89.6 / 75.3 & 79.6 / 54.7 & 89.7 / 51.9 & 84.0 / 57.2 & 83.1 / 57.1 & 72.7 / 45.0 \\
\midrule 
 \textbf{Human (before discussion)}   &  87.4 / 78.0  & 75.5 / 75.5  &  85.3 / 75.0 & 84.4 / 76.0  &  86.5 / 86.0 & 97.3 / 91.1 & 83.5 / 56.5 & 90.1 / 75.8 & 92.4 / 73.1 & 90.0 / 76.5  & 86.9 / 76.3 \\
 \textbf{Human (after discussion)}   &  87.9  /  78.0  & 83.0  /  82.9  &  89.9  /  82.0 & 89.9  /  86.0  &  87.0  /  86.9 & 98.9  /  96.4 & 88.1  /  69.6 & 97.8  /  93.9 & 93.9  /  80.7 & 93.1  /  85.5  & 90.7  /  84.0 \\
\midrule
\textbf{Resolutions} & 7\% & 18\% & 12\% & 15\% & 3\% & 5\% & 18\% & 19\% & 9\% \\
\bottomrule
\end{tabular}}
\caption{Human Evaluation results before and after discussion compared to GPT-4 and the percentage of resolved disagreements per task.}
\label{tab:further-human-results}
\end{table*}

\input{temp0.3-results-table}

\begin{table*}[h]
\footnotesize
\scalebox{0.69}{
\centering
\begin{tabular}{lrrrrrrrrrrr}
\toprule 
   \multirow{2}{*}{\textbf{Models}}
 & \multicolumn{4}{c}{{\textbf{MT}}} 
 & \multicolumn{1}{c}{{\textbf{DG}}}
 & \multicolumn{1}{c}{{\textbf{DS}}}
 & \multicolumn{1}{c}{{\textbf{SC}}}
 & \multicolumn{1}{c}{{\textbf{SD}}}
 & \multicolumn{1}{c}{{\textbf{ID}}}
 & \multicolumn{1}{c}{{\textbf{CROW}}}
 & \multicolumn{1}{c}{{\textbf{CROW}}}  \\
 
 & \multicolumn{1}{c}{Zh-En}& \multicolumn{1}{c}{En-Fr}  & \multicolumn{1}{c}{En-De} & \multicolumn{1}{c}{En-Ru} & & & & & & \multicolumn{1}{c}{\textbf{Score (-MT)}} & \multicolumn{1}{c}{\textbf{Score}} \\
 \midrule

\textbf{GPT-4 (temp=0.0)} & 75.9 / 57.9 & 54.5 / 21.5 & 54.4 / 20.5 & 54.1 / 19.7 & 72.4 / 46.5 & 89.6 / 75.3 & 79.6 / 54.7 & 89.7 / 51.9 & 84.0 / 57.2 & 83.1 / 57.1 & 72.7 / 45.0 \\
\textbf{GPT-4 (temp=0.1)} & 76.4 / 58.1 & 53.3 / 20.7 & 54.0 / 19.9 & 53.7 / 20.3 & 72.5 / 46.0 & 89.5 / 75.3 & 79.9 / 54.9 & 89.4 / 50.8 & 83.5 / 56.1 & 83.0 / 56.7 & 72.5 / 44.7 \\
\textbf{GPT-4 (temp=0.3)} & 75.6 / 56.6 & 54.5 / 22.1 & 54.6 / 20.3 & 53.8 / 20.5 & 72.0 / 45.6 & 90.5 / 77.3 & 81.5 / 57.3 & 89.2 / 50.2 & 84.7 / 59.0 & 83.6 / 57.9 & 72.9 / 45.4 \\

\midrule 
 \textbf{Human$^*$}   &  87.9  /  78.0  & 83.0  /  82.9  &  89.9  /  82.0 & 89.9  /  86.0  &  87.0  /  86.9 & 98.9  /  96.4 & 88.1  /  69.6 & 97.8  /  93.9 & 93.9  /  80.7 & 93.1  /  85.5  & 90.7  /  84.0 \\
\bottomrule
\end{tabular}}
\caption{\textbf{Macro-F1}  /  \textbf{Situational Accuracy} (\ie, results aggregated per \textit{context} instead of per \textit{sample}) for GPT-4 across \ourbenchmark{} tasks with varying temperature values.}
\label{tab:app_gpt4_temp_results}
\end{table*}

\begin{table*}[h]
\footnotesize
\scalebox{0.65}{
\centering
\begin{tabular}{lrrrrrrrrrrr}
\toprule 
   \multirow{2}{*}{\textbf{Models}}
 & \multicolumn{4}{c}{{\textbf{MT}}} 
 & \multicolumn{1}{c}{{\textbf{DG}}}
 & \multicolumn{1}{c}{{\textbf{DS}}}
 & \multicolumn{1}{c}{{\textbf{SC}}}
 & \multicolumn{1}{c}{{\textbf{SD}}}
 & \multicolumn{1}{c}{{\textbf{ID}}}
 & \multicolumn{1}{c}{{\textbf{CROW}}}
 & \multicolumn{1}{c}{{\textbf{CROW}}}  \\
 
 & \multicolumn{1}{c}{Zh-En}& \multicolumn{1}{c}{En-Fr}  & \multicolumn{1}{c}{En-De} & \multicolumn{1}{c}{En-Ru} & & & & & & \multicolumn{1}{c}{\textbf{Score (-MT)}} & \multicolumn{1}{c}{\textbf{Score}} \\
 \midrule

\textbf{GPT-4-CoT (temp=0.0)} & 71.6 / 52.2 & 64.7 / 42.6 & 57.1 / 34.2 & 57.3 / 30.0 & 55.3 / 22.8 & 88.6 / 70.6 & 84.3 / 60.7 & 87.8 / 47.3 & 84.0 / 57.0 & 80.0 / 51.7 & 72.3 / 46.4 \\
\textbf{GPT-4-CoT (temp=0.3, average)} & 71.3 / 51.1 & 64.4 / 42.0 & 57.4 / 34.1 & 56.6 / 28.9 & 55.2 / 22.8 & 88.8 / 71.3 & 83.7 / 60.8 & 88.1 / 47.5 & 83.9 / 57.3 & 80.0 / 51.9 & 72.2 / 46.2 \\
\textbf{GPT-4-CoT (temp=0.3, majority)} & 71.5 / 51.3 & 64.2 / 42.0 & 58.1 / 35.2 & 56.0 / 27.8 & 55.1 / 23.0 & 89.6 / 73.3 & 83.1 / 59.4 & 87.9 / 45.6 & 84.2 / 57.9 & 80.0 / 51.9 & 72.2 / 46.2 \\

\midrule 
 \textbf{Human$^*$}   &  87.9  /  78.0  & 83.0  /  82.9  &  89.9  /  82.0 & 89.9  /  86.0  &  87.0  /  86.9 & 98.9  /  96.4 & 88.1  /  69.6 & 97.8  /  93.9 & 93.9  /  80.7 & 93.1  /  85.5  & 90.7  /  84.0 \\
\bottomrule
\end{tabular}}
\caption{\textbf{Macro-F1}  /  \textbf{Situational Accuracy} (\ie, results aggregated per \textit{context} instead of per \textit{sample}) for GPT-4 with CoT across \ourbenchmark{} tasks in different scenarios. In the 'average' scenario, an average of five experiment results are reported. In the 'majority' scenario, similar to \cite{wang2023selfconsistency}, results based on the majority answer from five experiments are reported.}
\label{tab:app_gpt4_cot_results}
\end{table*}

\begin{figure*}[t]
    \centering    
    \includegraphics[width=0.9\textwidth]{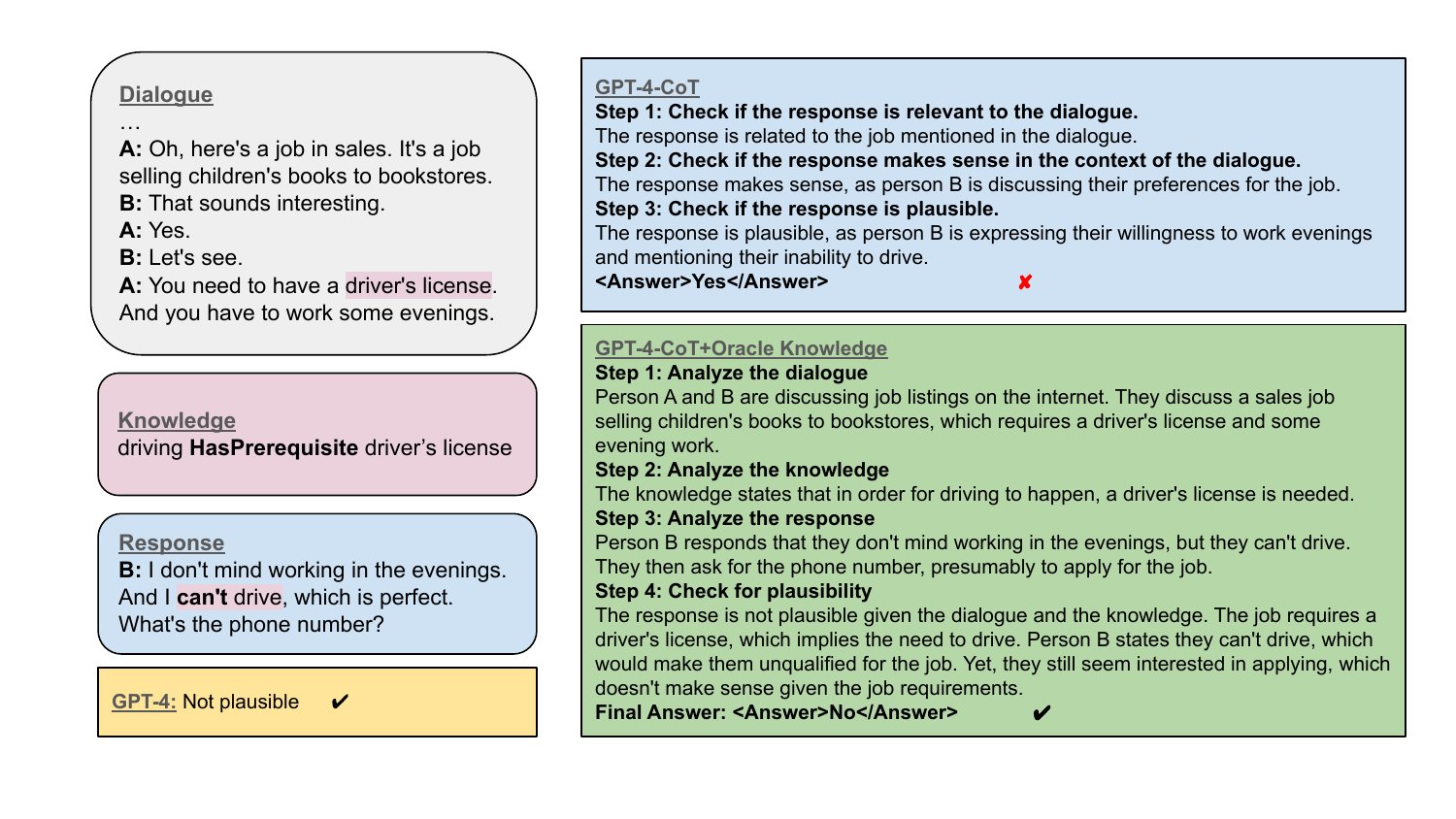}
    \caption{\textbf{Answers generated by GPT-4 in various settings for an example from the Dialogue task.} While chain-of-thought reasoning results in an incorrect answer, addition of the commonsense knowledge guides the model toward the correct answer.}
    \label{fig:app-analysis-ex}
\end{figure*}

\section{Additional Results}
\label{sec:app-results}
In this section, we provide the results of further experiments we did with varying hyperparameters. In Tables \ref{tab:app_main_results}, \ref{tab:app_results_commonsense_dimension} and \ref{tab:app_results_oracle}, we report the averaged results of three runs where the temperature parameter is set to $0.3$ respectively for all baselines, comparison over commonsense knowledge across tasks and with or without oracle knowledge. In Table \ref{tab:app_gpt4_temp_results}, we report the results of GPT-4 with varying temperature values compared to human performance. In Table \ref{tab:app_gpt4_cot_results}, we report the results of GPT-4 with chain-of-thought reasoning in various scenarios, including with self-consistency decoding \cite{wang2023selfconsistency}. Overall, in none of the scenarios we observe a particularly different performance of the models than what is reported in the main results in Table \ref{tab:main_results-greedy}.

%% file: temp0.3-results-table.tex
\begin{table*}[t!]
\footnotesize
\centering
\scalebox{0.72}{
\begin{tabular}{lrrrrrrrrrrr}
\toprule
   \multirow{2}{*}{\textbf{Models}}
 & \multicolumn{4}{c}{{\textbf{MT}}} 
 & \multicolumn{1}{c}{{\textbf{DG}}}
 & \multicolumn{1}{c}{{\textbf{DS}}}
 & \multicolumn{1}{c}{{\textbf{SC}}}
 & \multicolumn{1}{c}{{\textbf{SD}}}
 & \multicolumn{1}{c}{{\textbf{ID}}}
 & \multicolumn{1}{c}{{\textbf{CROW}}}
 & \multicolumn{1}{c}{{\textbf{CROW}}}  \\
 
 & \multicolumn{1}{c}{Zh-En}& \multicolumn{1}{c}{En-Fr}  & \multicolumn{1}{c}{En-De} & \multicolumn{1}{c}{En-Ru} & & & & & & \multicolumn{1}{c}{\textbf{Score (-MT)}} & \multicolumn{1}{c}{\textbf{Score}} \\
 \midrule

\textbf{Majority} & 33.3 / 0.0 & 33.3 / 0.0 & 33.3 / 0.0 & 33.3 / 0.0 & 40.1 / 0.0 & 42.8 / 0.0 & 33.6 / 0.0 & 36.5 / 0.0 & 41.3 / 0.0 & 38.9 / 0.0 & 36.4 / 0.0 \\
\textbf{Random} & 49.9 / 25.4 & 50.0 / 24.7 & 50.3 / 24.9 & 49.0 / 23.8 & 48.1 / 14.4 & 46.4 / 9.9 & 50.6 / 6.9 & 49.8 / 0.6 & 48.3 / 10.2 & 48.6 / 8.4 & 49.1 / 15.6 \\
\midrule
\textbf{LLaMa-7B} & 33.3 / 0.0 & -- & -- & -- & 41.5 / 1.3 & 44.3 / 1.4 & 33.7 / 0.0 & 29.9 / 0.0 & 41.3 / 0.0 & 38.1 / 0.5 & 37.3 / 0.5 \\
\textbf{LLaMa-13B} & 46.0 / 13.8 & -- & -- & -- & 50.2 / 10.9 & 45.1 / 2.1 & 34.4 / 0.7 & 30.6 / 0.1 & 43.8 / 1.8 & 40.8 / 3.1 & 41.7 / 4.9 \\
\textbf{LLaMa-33B} & 33.3 / 0.1 & -- & -- & -- & 52.5 / 4.9 & 49.6 / 5.6 & 33.7 / 0.2 & 30.1 / 0.0 & 41.3 / 0.0 & 41.4 / 2.1 & 40.1 / 1.8 \\
\textbf{Flan-T5-11B} & 57.0 / 26.8 & -- & -- & -- & 68.6 / 39.0 & 64.7 / 30.7 & 75.4 / 48.2 & 83.0 / 30.6 & 83.3 / 56.8 & 75.0 / 41.1 & 72.0 / 38.7 \\
\textbf{Alpaca} & 38.4 / 4.7 & -- & -- & -- & 54.8 / 16.7 & 56.5 / 12.3 & 41.0 / 6.1 & 43.8 / 2.4 & 52.0 / 9.4 & 49.6 / 9.4 & 47.8 / 8.6 \\
\textbf{Flan-Alpaca} & 60.2 / 26.5 & -- & -- & -- & 62.5 / 28.0 & 52.0 / 18.4 & 67.2 / 36.4 & 67.0 / 11.3 & 78.5 / 46.0 & 65.5 / 28.0 & 64.6 / 27.8 \\
\textbf{Vicuna} & 37.6 / 4.2 & -- & -- & -- & 60.4 / 21.4 & 61.9 / 18.9 & 45.1 / 11.7 & 42.0 / 1.6 & 57.2 / 15.0 & 53.3 / 13.7 & 50.7 / 12.1 \\
\textbf{Stable-Vicuna} & 60.5 / 30.9 & -- & -- & -- & 52.0 / 11.9 & 38.7 / 7.2 & 51.1 / 17.8 & 68.6 / 14.3 & 63.9 / 23.6 & 54.9 / 15.0 & 55.8 / 17.6 \\
\midrule
\textbf{mT0} & 55.1 / 20.2 & 37.7 / 3.3 & 35.6 / 1.5 & 33.9 / 0.3 & 44.8 / 4.8 & 64.0 / 23.7 & 55.2 / 14.6 & 50.8 / 2.8 & 49.5 / 6.2 & 52.8 / 10.4 & 47.4 / 8.6 \\
\textbf{BloomZ-7B} & 55.1 / 19.8 & 47.0 / 15.0 & 49.6 / 18.0 & 49.2 / 17.9 & 52.1 / 11.1 & 56.5 / 15.1 & 57.7 / 16.2 & 68.3 / 8.5 & 64.3 / 21.8 & 59.8 / 14.5 & 55.5 / 15.9 \\
\midrule
\textbf{PaLM 1} & 33.8 / 0.6 & 34.2 / 1.1 & 34.1 / 0.9 & 33.4 / 0.3 & 63.6 / 26.8 & 62.6 / 23.0 & 51.6 / 15.4 & 56.6 / 7.9 & 63.5 / 21.9 & 59.6 / 19.0 & 48.2 / 10.9 \\
\textbf{GPT-3} & 66.7 / 39.4 & 48.5 / 16.1 & 49.2 / 17.2 & 48.1 / 12.1 & 67.1 / 36.8 & 68.6 / 32.6 & 69.2 / 38.2 & 85.6 / 39.5 & 76.4 / 42.0 & 73.4 / 37.8 & 64.2 / 29.7 \\
\textbf{GPT-4} & 75.6 / 56.6 & 54.5 / 22.1 & 54.6 / 20.3 & 53.8 / 20.5 & 72.0 / 45.6 & 90.5 / 77.3 & 81.5 / 57.3 & 89.2 / 50.2 & 84.7 / 59.0 & 83.6 / 57.9 & 72.9 / 45.4 \\
\textbf{GPT-4-CoT} & 71.2 / 51.1 & 64.3 / 41.9 & 57.5 / 34.3 & 56.9 / 29.2 & 55.2 / 23.1 & 89.0 / 71.7 & 83.6 / 60.6 & 88.2 / 47.0 & 84.2 / 57.8 & 80.0 / 52.0 & 72.2 / 46.3 \\

\midrule 
\textbf{Human$^*$}   &  87.9  /  78.0  & 83.0  /  82.9  &  89.9  /  82.0 & 89.9  /  86.0  &  87.0  /  86.9 & 98.9  /  96.4 & 88.1  /  69.6 & 97.8  /  93.9 & 93.9  /  80.7 & 93.1  /  85.5  & 90.7  /  84.0 \\
\bottomrule
\end{tabular}}
\caption{\textbf{Macro-F1} / \textbf{Situational Accuracy} (\ie, results aggregated per \textit{context} instead of per \textit{sample}) for all examined models across \ourbenchmark{} tasks. All model results are averaged over three runs with different seeds. Temperature is set to 0.3 for all runs. $^*$Due to the cost of expert evaluation, our \textbf{Human} study is only evaluated on 100 instances per task.}
\label{tab:app_main_results}
\end{table*}

%% file: emnlp2023.bbl
\begin{thebibliography}{73}
\expandafter\ifx\csname natexlab\endcsname\relax\def\natexlab#1{#1}\fi

\bibitem[{Amidei et~al.(2018)Amidei, Piwek, and Willis}]{amidei-etal-2018-rethinking}
Jacopo Amidei, Paul Piwek, and Alistair Willis. 2018.
\newblock \href {https://aclanthology.org/C18-1281} {Rethinking the agreement in human evaluation tasks}.
\newblock In \emph{Proceedings of the 27th International Conference on Computational Linguistics}, pages 3318--3329, Santa Fe, New Mexico, USA. Association for Computational Linguistics.

\bibitem[{Bar-Hillel(1960)}]{BarHillel1960ADO}
Yehoshua Bar-Hillel. 1960.
\newblock A demonstration of the nonfeasibility of fully automatic high quality translation.

\bibitem[{Bhagavatula et~al.(2019)Bhagavatula, Bras, Malaviya, Sakaguchi, Holtzman, Rashkin, Downey, Yih, and Choi}]{Bhagavatula2019AbductiveCR}
Chandra Bhagavatula, Ronan~Le Bras, Chaitanya Malaviya, Keisuke Sakaguchi, Ari Holtzman, Hannah Rashkin, Doug Downey, Scott Yih, and Yejin Choi. 2019.
\newblock Abductive commonsense reasoning.
\newblock \emph{ArXiv}, abs/1908.05739.

\bibitem[{Bisk et~al.(2019)Bisk, Zellers, Bras, Gao, and Choi}]{Bisk2019PIQARA}
Yonatan Bisk, Rowan Zellers, Ronan~Le Bras, Jianfeng Gao, and Yejin Choi. 2019.
\newblock Piqa: Reasoning about physical commonsense in natural language.
\newblock In \emph{AAAI Conference on Artificial Intelligence}.

\bibitem[{Brown et~al.(2020)Brown, Mann, Ryder, Subbiah, Kaplan, Dhariwal, Neelakantan, Shyam, Sastry, Askell, Agarwal, Herbert-Voss, Krueger, Henighan, Child, Ramesh, Ziegler, Wu, Winter, Hesse, Chen, Sigler, Litwin, Gray, Chess, Clark, Berner, McCandlish, Radford, Sutskever, and Amodei}]{NEURIPS2020_gpt3}
Tom Brown, Benjamin Mann, Nick Ryder, Melanie Subbiah, Jared~D Kaplan, Prafulla Dhariwal, Arvind Neelakantan, Pranav Shyam, Girish Sastry, Amanda Askell, Sandhini Agarwal, Ariel Herbert-Voss, Gretchen Krueger, Tom Henighan, Rewon Child, Aditya Ramesh, Daniel Ziegler, Jeffrey Wu, Clemens Winter, Chris Hesse, Mark Chen, Eric Sigler, Mateusz Litwin, Scott Gray, Benjamin Chess, Jack Clark, Christopher Berner, Sam McCandlish, Alec Radford, Ilya Sutskever, and Dario Amodei. 2020.
\newblock \href {https://proceedings.neurips.cc/paper_files/paper/2020/file/1457c0d6bfcb4967418bfb8ac142f64a-Paper.pdf} {Language models are few-shot learners}.
\newblock In \emph{Advances in Neural Information Processing Systems}, volume~33, pages 1877--1901. Curran Associates, Inc.

\bibitem[{Chen et~al.(2019)Chen, D{'}Arcy, Liu, Fernandez, and Downey}]{chen-etal-2019-codah}
Michael Chen, Mike D{'}Arcy, Alisa Liu, Jared Fernandez, and Doug Downey. 2019.
\newblock \href {https://doi.org/10.18653/v1/W19-2008} {{CODAH}: An adversarially-authored question answering dataset for common sense}.
\newblock In \emph{Proceedings of the 3rd Workshop on Evaluating Vector Space Representations for {NLP}}, pages 63--69, Minneapolis, USA. Association for Computational Linguistics.

\bibitem[{Chen et~al.(2021)Chen, Liu, Chen, and Zhang}]{chen-etal-2021-dialogsum}
Yulong Chen, Yang Liu, Liang Chen, and Yue Zhang. 2021.
\newblock \href {https://doi.org/10.18653/v1/2021.findings-acl.449} {{D}ialog{S}um: {A} real-life scenario dialogue summarization dataset}.
\newblock In \emph{Findings of the Association for Computational Linguistics: ACL-IJCNLP 2021}, pages 5062--5074, Online. Association for Computational Linguistics.

\bibitem[{Chia et~al.(2023)Chia, Hong, Bing, and Poria}]{chia2023instructeval}
Yew~Ken Chia, Pengfei Hong, Lidong Bing, and Soujanya Poria. 2023.
\newblock Instructeval: Towards holistic evaluation of instruction-tuned large language models.
\newblock \emph{arXiv preprint arXiv:2306.04757}.

\bibitem[{Chiang et~al.(2023)Chiang, Li, Lin, Sheng, Wu, Zhang, Zheng, Zhuang, Zhuang, Gonzalez, Stoica, and Xing}]{vicuna2023}
Wei-Lin Chiang, Zhuohan Li, Zi~Lin, Ying Sheng, Zhanghao Wu, Hao Zhang, Lianmin Zheng, Siyuan Zhuang, Yonghao Zhuang, Joseph~E. Gonzalez, Ion Stoica, and Eric~P. Xing. 2023.
\newblock \href {https://lmsys.org/blog/2023-03-30-vicuna/} {Vicuna: An open-source chatbot impressing gpt-4 with 90\%* chatgpt quality}.

\bibitem[{Choi(2022)}]{Choi2022TheCC}
Yejin Choi. 2022.
\newblock The curious case of commonsense intelligence.
\newblock \emph{Daedalus}, 151:139--155.

\bibitem[{Chowdhery et~al.(2022)Chowdhery, Narang, Devlin, Bosma, Mishra, Roberts, Barham, Chung, Sutton, Gehrmann, Schuh, Shi, Tsvyashchenko, Maynez, Rao, Barnes, Tay, Shazeer, Prabhakaran, Reif, Du, Hutchinson, Pope, Bradbury, Austin, Isard, Gur-Ari, Yin, Duke, Levskaya, Ghemawat, Dev, Michalewski, Garcia, Misra, Robinson, Fedus, Zhou, Ippolito, Luan, Lim, Zoph, Spiridonov, Sepassi, Dohan, Agrawal, Omernick, Dai, Pillai, Pellat, Lewkowycz, Moreira, Child, Polozov, Lee, Zhou, Wang, Saeta, Diaz, Firat, Catasta, Wei, Meier-Hellstern, Eck, Dean, Petrov, and Fiedel}]{chowdhery2022palm}
Aakanksha Chowdhery, Sharan Narang, Jacob Devlin, Maarten Bosma, Gaurav Mishra, Adam Roberts, Paul Barham, Hyung~Won Chung, Charles Sutton, Sebastian Gehrmann, Parker Schuh, Kensen Shi, Sasha Tsvyashchenko, Joshua Maynez, Abhishek Rao, Parker Barnes, Yi~Tay, Noam Shazeer, Vinodkumar Prabhakaran, Emily Reif, Nan Du, Ben Hutchinson, Reiner Pope, James Bradbury, Jacob Austin, Michael Isard, Guy Gur-Ari, Pengcheng Yin, Toju Duke, Anselm Levskaya, Sanjay Ghemawat, Sunipa Dev, Henryk Michalewski, Xavier Garcia, Vedant Misra, Kevin Robinson, Liam Fedus, Denny Zhou, Daphne Ippolito, David Luan, Hyeontaek Lim, Barret Zoph, Alexander Spiridonov, Ryan Sepassi, David Dohan, Shivani Agrawal, Mark Omernick, Andrew~M. Dai, Thanumalayan~Sankaranarayana Pillai, Marie Pellat, Aitor Lewkowycz, Erica Moreira, Rewon Child, Oleksandr Polozov, Katherine Lee, Zongwei Zhou, Xuezhi Wang, Brennan Saeta, Mark Diaz, Orhan Firat, Michele Catasta, Jason Wei, Kathy Meier-Hellstern, Douglas Eck, Jeff Dean, Slav Petrov, and Noah Fiedel. 2022.
\newblock \href {http://arxiv.org/abs/2204.02311} {Palm: Scaling language modeling with pathways}.

\bibitem[{Chung et~al.(2022)Chung, Hou, Longpre, Zoph, Tay, Fedus, Li, Wang, Dehghani, Brahma, Webson, Gu, Dai, Suzgun, Chen, Chowdhery, Castro-Ros, Pellat, Robinson, Valter, Narang, Mishra, Yu, Zhao, Huang, Dai, Yu, Petrov, Chi, Dean, Devlin, Roberts, Zhou, Le, and Wei}]{chung2022scaling}
Hyung~Won Chung, Le~Hou, Shayne Longpre, Barret Zoph, Yi~Tay, William Fedus, Yunxuan Li, Xuezhi Wang, Mostafa Dehghani, Siddhartha Brahma, Albert Webson, Shixiang~Shane Gu, Zhuyun Dai, Mirac Suzgun, Xinyun Chen, Aakanksha Chowdhery, Alex Castro-Ros, Marie Pellat, Kevin Robinson, Dasha Valter, Sharan Narang, Gaurav Mishra, Adams Yu, Vincent Zhao, Yanping Huang, Andrew Dai, Hongkun Yu, Slav Petrov, Ed~H. Chi, Jeff Dean, Jacob Devlin, Adam Roberts, Denny Zhou, Quoc~V. Le, and Jason Wei. 2022.
\newblock \href {http://arxiv.org/abs/2210.11416} {Scaling instruction-finetuned language models}.

\bibitem[{Dalvi et~al.(2018)Dalvi, Huang, Tandon, Yih, and Clark}]{dalvi-etal-2018-tracking}
Bhavana Dalvi, Lifu Huang, Niket Tandon, Wen-tau Yih, and Peter Clark. 2018.
\newblock \href {https://doi.org/10.18653/v1/N18-1144} {Tracking state changes in procedural text: a challenge dataset and models for process paragraph comprehension}.
\newblock In \emph{Proceedings of the 2018 Conference of the North {A}merican Chapter of the Association for Computational Linguistics: Human Language Technologies, Volume 1 (Long Papers)}, pages 1595--1604, New Orleans, Louisiana. Association for Computational Linguistics.

\bibitem[{Davis(2023)}]{davis2023benchmarks}
Ernest Davis. 2023.
\newblock \href {http://arxiv.org/abs/2302.04752} {Benchmarks for automated commonsense reasoning: A survey}.

\bibitem[{Davis and Marcus(2015)}]{Davis2015CR}
Ernest Davis and Gary Marcus. 2015.
\newblock \href {https://doi.org/10.1145/2701413} {Commonsense reasoning and commonsense knowledge in artificial intelligence}.
\newblock \emph{Commun. ACM}, 58(9):92–103.

\bibitem[{Do and Pavlick(2021)}]{do-pavlick-2021-rotten}
Nam Do and Ellie Pavlick. 2021.
\newblock \href {https://doi.org/10.18653/v1/2021.findings-acl.181} {Are rotten apples edible? challenging commonsense inference ability with exceptions}.
\newblock In \emph{Findings of the Association for Computational Linguistics: ACL-IJCNLP 2021}, pages 2061--2073, Online. Association for Computational Linguistics.

\bibitem[{Eisenschlos et~al.(2023)Eisenschlos, Cole, Liu, and Cohen}]{eisenschlos-etal-2023-winodict}
Julian~Martin Eisenschlos, Jeremy~R. Cole, Fangyu Liu, and William~W. Cohen. 2023.
\newblock \href {https://aclanthology.org/2023.eacl-main.7} {{W}ino{D}ict: Probing language models for in-context word acquisition}.
\newblock In \emph{Proceedings of the 17th Conference of the European Chapter of the Association for Computational Linguistics}, pages 94--102, Dubrovnik, Croatia. Association for Computational Linguistics.

\bibitem[{Emelin and Sennrich(2021)}]{emelin-sennrich-2021-wino}
Denis Emelin and Rico Sennrich. 2021.
\newblock \href {https://doi.org/10.18653/v1/2021.emnlp-main.670} {Wino-{X}: Multilingual {W}inograd schemas for commonsense reasoning and coreference resolution}.
\newblock In \emph{Proceedings of the 2021 Conference on Empirical Methods in Natural Language Processing}, pages 8517--8532, Online and Punta Cana, Dominican Republic. Association for Computational Linguistics.

\bibitem[{Gabriel et~al.(2022)Gabriel, Hallinan, Sap, Nguyen, Roesner, Choi, and Choi}]{gabriel-etal-2022-misinfo}
Saadia Gabriel, Skyler Hallinan, Maarten Sap, Pemi Nguyen, Franziska Roesner, Eunsol Choi, and Yejin Choi. 2022.
\newblock \href {https://doi.org/10.18653/v1/2022.acl-long.222} {Misinfo reaction frames: Reasoning about readers{'} reactions to news headlines}.
\newblock In \emph{Proceedings of the 60th Annual Meeting of the Association for Computational Linguistics (Volume 1: Long Papers)}, pages 3108--3127, Dublin, Ireland. Association for Computational Linguistics.

\bibitem[{Gao et~al.(2022)Gao, Hwang, Kanno, Wakaki, Mitsufuji, and Bosselut}]{gao-etal-2022-comfact}
Silin Gao, Jena~D. Hwang, Saya Kanno, Hiromi Wakaki, Yuki Mitsufuji, and Antoine Bosselut. 2022.
\newblock \href {https://aclanthology.org/2022.findings-emnlp.120} {{C}om{F}act: A benchmark for linking contextual commonsense knowledge}.
\newblock In \emph{Findings of the Association for Computational Linguistics: EMNLP 2022}, pages 1656--1675, Abu Dhabi, United Arab Emirates. Association for Computational Linguistics.

\bibitem[{Geva et~al.(2019)Geva, Goldberg, and Berant}]{geva-etal-2019-modeling}
Mor Geva, Yoav Goldberg, and Jonathan Berant. 2019.
\newblock \href {https://doi.org/10.18653/v1/D19-1107} {Are we modeling the task or the annotator? an investigation of annotator bias in natural language understanding datasets}.
\newblock In \emph{Proceedings of the 2019 Conference on Empirical Methods in Natural Language Processing and the 9th International Joint Conference on Natural Language Processing (EMNLP-IJCNLP)}, pages 1161--1166, Hong Kong, China. Association for Computational Linguistics.

\bibitem[{Ghosal et~al.(2021)Ghosal, Hong, Shen, Majumder, Mihalcea, and Poria}]{ghosal-etal-2021-cider}
Deepanway Ghosal, Pengfei Hong, Siqi Shen, Navonil Majumder, Rada Mihalcea, and Soujanya Poria. 2021.
\newblock \href {https://aclanthology.org/2021.sigdial-1.33} {{CIDER}: Commonsense inference for dialogue explanation and reasoning}.
\newblock In \emph{Proceedings of the 22nd Annual Meeting of the Special Interest Group on Discourse and Dialogue}, pages 301--313, Singapore and Online. Association for Computational Linguistics.

\bibitem[{Ghosal et~al.(2022)Ghosal, Shen, Majumder, Mihalcea, and Poria}]{ghosal-etal-2022-cicero}
Deepanway Ghosal, Siqi Shen, Navonil Majumder, Rada Mihalcea, and Soujanya Poria. 2022.
\newblock \href {https://doi.org/10.18653/v1/2022.acl-long.344} {{CICERO}: A dataset for contextualized commonsense inference in dialogues}.
\newblock In \emph{Proceedings of the 60th Annual Meeting of the Association for Computational Linguistics (Volume 1: Long Papers)}, pages 5010--5028, Dublin, Ireland. Association for Computational Linguistics.

\bibitem[{He et~al.(2020)He, Wang, Xiong, and Liu}]{he-etal-2020-box}
Jie He, Tao Wang, Deyi Xiong, and Qun Liu. 2020.
\newblock \href {https://doi.org/10.18653/v1/2020.findings-emnlp.327} {The box is in the pen: Evaluating commonsense reasoning in neural machine translation}.
\newblock In \emph{Findings of the Association for Computational Linguistics: EMNLP 2020}, pages 3662--3672, Online. Association for Computational Linguistics.

\bibitem[{Huang et~al.(2019)Huang, Le~Bras, Bhagavatula, and Choi}]{huang-etal-2019-cosmos}
Lifu Huang, Ronan Le~Bras, Chandra Bhagavatula, and Yejin Choi. 2019.
\newblock \href {https://doi.org/10.18653/v1/D19-1243} {Cosmos {QA}: Machine reading comprehension with contextual commonsense reasoning}.
\newblock In \emph{Proceedings of the 2019 Conference on Empirical Methods in Natural Language Processing and the 9th International Joint Conference on Natural Language Processing (EMNLP-IJCNLP)}, pages 2391--2401, Hong Kong, China. Association for Computational Linguistics.

\bibitem[{Hwang et~al.(2020)Hwang, Bhagavatula, Bras, Da, Sakaguchi, Bosselut, and Choi}]{Hwang2020COMETATOMIC2O}
Jena~D. Hwang, Chandra Bhagavatula, Ronan~Le Bras, Jeff Da, Keisuke Sakaguchi, Antoine Bosselut, and Yejin Choi. 2020.
\newblock Comet-atomic 2020: On symbolic and neural commonsense knowledge graphs.
\newblock In \emph{AAAI Conference on Artificial Intelligence}.

\bibitem[{Ilievski et~al.(2021)Ilievski, Oltramari, Ma, Zhang, McGuinness, and Szekely}]{ilievski2021dimensions}
Filip Ilievski, Alessandro Oltramari, Kaixin Ma, Bin Zhang, Deborah~L. McGuinness, and Pedro Szekely. 2021.
\newblock \href {http://arxiv.org/abs/2101.04640} {Dimensions of commonsense knowledge}.

\bibitem[{Kim et~al.(2022)Kim, Joo, Chae, Kim, won Hwang, and Yeo}]{Kim2022MindTG}
Seungone Kim, Sehrang Joo, Hyungjoo Chae, Chaehyeong Kim, Seung won Hwang, and Jinyoung Yeo. 2022.
\newblock Mind the gap! injecting commonsense knowledge for abstractive dialogue summarization.
\newblock In \emph{International Conference on Computational Linguistics}.

\bibitem[{Levesque et~al.(2011)Levesque, Davis, and Morgenstern}]{Levesque2011TheWS}
Hector~J. Levesque, Ernest Davis, and L.~Morgenstern. 2011.
\newblock The winograd schema challenge.
\newblock In \emph{International Conference on Principles of Knowledge Representation and Reasoning}.

\bibitem[{Levy et~al.(2022)Levy, Allaway, Subbiah, Chilton, Patton, McKeown, and Wang}]{levy-etal-2022-safetext}
Sharon Levy, Emily Allaway, Melanie Subbiah, Lydia Chilton, Desmond Patton, Kathleen McKeown, and William~Yang Wang. 2022.
\newblock \href {https://aclanthology.org/2022.emnlp-main.154} {{S}afe{T}ext: A benchmark for exploring physical safety in language models}.
\newblock In \emph{Proceedings of the 2022 Conference on Empirical Methods in Natural Language Processing}, pages 2407--2421, Abu Dhabi, United Arab Emirates. Association for Computational Linguistics.

\bibitem[{Lin et~al.(2020{\natexlab{a}})Lin, Lee, Khanna, and Ren}]{lin-etal-2020-birds}
Bill~Yuchen Lin, Seyeon Lee, Rahul Khanna, and Xiang Ren. 2020{\natexlab{a}}.
\newblock \href {https://doi.org/10.18653/v1/2020.emnlp-main.557} {{B}irds have four legs?! {N}umer{S}ense: {P}robing {N}umerical {C}ommonsense {K}nowledge of {P}re-{T}rained {L}anguage {M}odels}.
\newblock In \emph{Proceedings of the 2020 Conference on Empirical Methods in Natural Language Processing (EMNLP)}, pages 6862--6868, Online. Association for Computational Linguistics.

\bibitem[{Lin et~al.(2020{\natexlab{b}})Lin, Zhou, Shen, Zhou, Bhagavatula, Choi, and Ren}]{lin-etal-2020-commongen}
Bill~Yuchen Lin, Wangchunshu Zhou, Ming Shen, Pei Zhou, Chandra Bhagavatula, Yejin Choi, and Xiang Ren. 2020{\natexlab{b}}.
\newblock \href {https://doi.org/10.18653/v1/2020.findings-emnlp.165} {{C}ommon{G}en: A constrained text generation challenge for generative commonsense reasoning}.
\newblock In \emph{Findings of the Association for Computational Linguistics: EMNLP 2020}, pages 1823--1840, Online. Association for Computational Linguistics.

\bibitem[{Lourie et~al.(2021)Lourie, Bras, Bhagavatula, and Choi}]{Lourie2021UNICORNOR}
Nicholas Lourie, Ronan~Le Bras, Chandra Bhagavatula, and Yejin Choi. 2021.
\newblock Unicorn on rainbow: A universal commonsense reasoning model on a new multitask benchmark.
\newblock \emph{ArXiv}, abs/2103.13009.

\bibitem[{McCarthy(1960)}]{McCarthy1960ProgramsWC}
John McCarthy. 1960.
\newblock Programs with common sense.

\bibitem[{Muennighoff et~al.(2023)Muennighoff, Wang, Sutawika, Roberts, Biderman, Scao, Bari, Shen, Yong, Schoelkopf, Tang, Radev, Aji, Almubarak, Albanie, Alyafeai, Webson, Raff, and Raffel}]{muennighoff2022crosslingual}
Niklas Muennighoff, Thomas Wang, Lintang Sutawika, Adam Roberts, Stella~Rose Biderman, Teven~Le Scao, M~Saiful Bari, Sheng Shen, Zheng-Xin Yong, Hailey Schoelkopf, Xiangru Tang, Dragomir~R. Radev, Alham~Fikri Aji, Khalid Almubarak, Samuel Albanie, Zaid Alyafeai, Albert Webson, Edward Raff, and Colin Raffel. 2023.
\newblock \href {https://api.semanticscholar.org/CorpusID:253264914} {Crosslingual generalization through multitask finetuning}.
\newblock In \emph{Annual Meeting of the Association for Computational Linguistics}.

\bibitem[{Nangia et~al.(2021)Nangia, Sugawara, Trivedi, Warstadt, Vania, and Bowman}]{nangia-etal-2021-ingredients}
Nikita Nangia, Saku Sugawara, Harsh Trivedi, Alex Warstadt, Clara Vania, and Samuel~R. Bowman. 2021.
\newblock \href {https://doi.org/10.18653/v1/2021.acl-long.98} {What ingredients make for an effective crowdsourcing protocol for difficult {NLU} data collection tasks?}
\newblock In \emph{Proceedings of the 59th Annual Meeting of the Association for Computational Linguistics and the 11th International Joint Conference on Natural Language Processing (Volume 1: Long Papers)}, pages 1221--1235, Online. Association for Computational Linguistics.

\bibitem[{Oortwijn et~al.(2021)Oortwijn, Ossenkoppele, and Betti}]{oortwijn-etal-2021-interrater}
Yvette Oortwijn, Thijs Ossenkoppele, and Arianna Betti. 2021.
\newblock \href {https://aclanthology.org/2021.humeval-1.15} {Interrater disagreement resolution: A systematic procedure to reach consensus in annotation tasks}.
\newblock In \emph{Proceedings of the Workshop on Human Evaluation of NLP Systems (HumEval)}, pages 131--141, Online. Association for Computational Linguistics.

\bibitem[{OpenAI(2023)}]{openai2023gpt4}
OpenAI. 2023.
\newblock \href {http://arxiv.org/abs/2303.08774} {Gpt-4 technical report}.

\bibitem[{Peng et~al.(2023)Peng, Li, He, Galley, and Gao}]{peng2023instruction}
Baolin Peng, Chunyuan Li, Pengcheng He, Michel Galley, and Jianfeng Gao. 2023.
\newblock Instruction tuning with gpt-4.
\newblock \emph{arXiv preprint arXiv:2304.03277}.

\bibitem[{Qin et~al.(2021)Qin, Gupta, Upadhyay, He, Choi, and Faruqui}]{qin-etal-2021-timedial}
Lianhui Qin, Aditya Gupta, Shyam Upadhyay, Luheng He, Yejin Choi, and Manaal Faruqui. 2021.
\newblock \href {https://doi.org/10.18653/v1/2021.acl-long.549} {{TIMEDIAL}: Temporal commonsense reasoning in dialog}.
\newblock In \emph{Proceedings of the 59th Annual Meeting of the Association for Computational Linguistics and the 11th International Joint Conference on Natural Language Processing (Volume 1: Long Papers)}, pages 7066--7076, Online. Association for Computational Linguistics.

\bibitem[{Raffel et~al.(2020)Raffel, Shazeer, Roberts, Lee, Narang, Matena, Zhou, Li, and Liu}]{raffel2020exploring}
Colin Raffel, Noam Shazeer, Adam Roberts, Katherine Lee, Sharan Narang, Michael Matena, Yanqi Zhou, Wei Li, and Peter~J Liu. 2020.
\newblock Exploring the limits of transfer learning with a unified text-to-text transformer.
\newblock \emph{The Journal of Machine Learning Research}, 21(1):5485--5551.

\bibitem[{Rashkin et~al.(2018{\natexlab{a}})Rashkin, Bosselut, Sap, Knight, and Choi}]{rashkin-etal-2018-modeling}
Hannah Rashkin, Antoine Bosselut, Maarten Sap, Kevin Knight, and Yejin Choi. 2018{\natexlab{a}}.
\newblock \href {https://doi.org/10.18653/v1/P18-1213} {Modeling naive psychology of characters in simple commonsense stories}.
\newblock In \emph{Proceedings of the 56th Annual Meeting of the Association for Computational Linguistics (Volume 1: Long Papers)}, pages 2289--2299, Melbourne, Australia. Association for Computational Linguistics.

\bibitem[{Rashkin et~al.(2018{\natexlab{b}})Rashkin, Sap, Allaway, Smith, and Choi}]{rashkin-etal-2018-event2mind}
Hannah Rashkin, Maarten Sap, Emily Allaway, Noah~A. Smith, and Yejin Choi. 2018{\natexlab{b}}.
\newblock \href {https://doi.org/10.18653/v1/P18-1043} {{E}vent2{M}ind: Commonsense inference on events, intents, and reactions}.
\newblock In \emph{Proceedings of the 56th Annual Meeting of the Association for Computational Linguistics (Volume 1: Long Papers)}, pages 463--473, Melbourne, Australia. Association for Computational Linguistics.

\bibitem[{Reddy et~al.(2019)Reddy, Chen, and Manning}]{reddy-etal-2019-coqa}
Siva Reddy, Danqi Chen, and Christopher~D. Manning. 2019.
\newblock \href {https://doi.org/10.1162/tacl_a_00266} {{C}o{QA}: A conversational question answering challenge}.
\newblock \emph{Transactions of the Association for Computational Linguistics}, 7:249--266.

\bibitem[{Richardson and Heck(2023)}]{richardson2023commonsense}
Christopher Richardson and Larry Heck. 2023.
\newblock \href {http://arxiv.org/abs/2302.07926} {Commonsense reasoning for conversational ai: A survey of the state of the art}.

\bibitem[{Roemmele et~al.(2011)Roemmele, Bejan, and Gordon}]{copaGordon2011}
Melissa Roemmele, Cosmin~Adrian Bejan, and Andrew~S. Gordon. 2011.
\newblock Choice of plausible alternatives: An evaluation of commonsense causal reasoning.

\bibitem[{Rudinger et~al.(2018)Rudinger, Naradowsky, Leonard, and Durme}]{Rudinger2018GenderBI}
Rachel Rudinger, Jason Naradowsky, Brian Leonard, and Benjamin~Van Durme. 2018.
\newblock Gender bias in coreference resolution.
\newblock In \emph{North American Chapter of the Association for Computational Linguistics}.

\bibitem[{Saha et~al.(2021)Saha, Yadav, Bauer, and Bansal}]{saha-etal-2021-explagraphs}
Swarnadeep Saha, Prateek Yadav, Lisa Bauer, and Mohit Bansal. 2021.
\newblock \href {https://doi.org/10.18653/v1/2021.emnlp-main.609} {{E}xpla{G}raphs: An explanation graph generation task for structured commonsense reasoning}.
\newblock In \emph{Proceedings of the 2021 Conference on Empirical Methods in Natural Language Processing}, pages 7716--7740, Online and Punta Cana, Dominican Republic. Association for Computational Linguistics.

\bibitem[{Sakaguchi et~al.(2019)Sakaguchi, Bras, Bhagavatula, and Choi}]{Sakaguchi2019WINOGRANDEAA}
Keisuke Sakaguchi, Ronan~Le Bras, Chandra Bhagavatula, and Yejin Choi. 2019.
\newblock Winogrande: An adversarial winograd schema challenge at scale.
\newblock \emph{ArXiv}, abs/1907.10641.

\bibitem[{Sap et~al.(2019{\natexlab{a}})Sap, Bras, Allaway, Bhagavatula, Lourie, Rashkin, Roof, Smith, and Choi}]{Sap2019ATOMICAA}
Maarten Sap, Ronan~Le Bras, Emily Allaway, Chandra Bhagavatula, Nicholas Lourie, Hannah Rashkin, Brendan Roof, Noah~A. Smith, and Yejin Choi. 2019{\natexlab{a}}.
\newblock Atomic: An atlas of machine commonsense for if-then reasoning.
\newblock In \emph{AAAI Conference on Artificial Intelligence}.

\bibitem[{Sap et~al.(2019{\natexlab{b}})Sap, Rashkin, Chen, Le~Bras, and Choi}]{sap-etal-2019-social}
Maarten Sap, Hannah Rashkin, Derek Chen, Ronan Le~Bras, and Yejin Choi. 2019{\natexlab{b}}.
\newblock \href {https://doi.org/10.18653/v1/D19-1454} {Social {IQ}a: Commonsense reasoning about social interactions}.
\newblock In \emph{Proceedings of the 2019 Conference on Empirical Methods in Natural Language Processing and the 9th International Joint Conference on Natural Language Processing (EMNLP-IJCNLP)}, pages 4463--4473, Hong Kong, China. Association for Computational Linguistics.

\bibitem[{Scao et~al.(2022)Scao, Fan, Akiki, Pavlick, Ili{\'c}, Hesslow, Castagn{\'e}, Luccioni, Yvon, Gall{\'e} et~al.}]{scao2022bloom}
Teven~Le Scao, Angela Fan, Christopher Akiki, Ellie Pavlick, Suzana Ili{\'c}, Daniel Hesslow, Roman Castagn{\'e}, Alexandra~Sasha Luccioni, Fran{\c{c}}ois Yvon, Matthias Gall{\'e}, et~al. 2022.
\newblock Bloom: A 176b-parameter open-access multilingual language model.
\newblock \emph{arXiv preprint arXiv:2211.05100}.

\bibitem[{Singh et~al.(2021)Singh, Wen, Hou, Alipoormolabashi, Wu, Ma, and Peng}]{singh-etal-2021-com2sense}
Shikhar Singh, Nuan Wen, Yu~Hou, Pegah Alipoormolabashi, Te-lin Wu, Xuezhe Ma, and Nanyun Peng. 2021.
\newblock \href {https://doi.org/10.18653/v1/2021.findings-acl.78} {{COM}2{SENSE}: A commonsense reasoning benchmark with complementary sentences}.
\newblock In \emph{Findings of the Association for Computational Linguistics: ACL-IJCNLP 2021}, pages 883--898, Online. Association for Computational Linguistics.

\bibitem[{Speer et~al.(2016)Speer, Chin, and Havasi}]{Speer2016ConceptNet5A}
Robyn Speer, Joshua Chin, and Catherine Havasi. 2016.
\newblock Conceptnet 5.5: An open multilingual graph of general knowledge.
\newblock \emph{ArXiv}, abs/1612.03975.

\bibitem[{Srivastava et~al.(2022)Srivastava, Rastogi, Rao, Shoeb, Abid, Fisch, Brown, Santoro, Gupta, Garriga-Alonso, Kluska, Lewkowycz, Agarwal, Power, Ray, Warstadt, Kocurek, Safaya, Tazarv, Xiang, Parrish, Nie, Hussain, Askell, Dsouza, Slone, Rahane, Iyer, Andreassen, Madotto, Santilli, Stuhlmüller, Dai, La, Lampinen, Zou, Jiang, Chen, Vuong, Gupta, Gottardi, Norelli, Venkatesh, Gholamidavoodi, Tabassum, Menezes, Kirubarajan, Mullokandov, Sabharwal, Herrick, Efrat, Erdem, Karakaş, Roberts, Loe, Zoph, Bojanowski, Özyurt, Hedayatnia, Neyshabur, Inden, Stein, Ekmekci, Lin, Howald, Diao, Dour, Stinson, Argueta, Ramírez, Singh, Rathkopf, Meng, Baral, Wu, Callison-Burch, Waites, Voigt, Manning, Potts, Ramirez, Rivera, Siro, Raffel, Ashcraft, Garbacea, Sileo, Garrette, Hendrycks, Kilman, Roth, Freeman, Khashabi, Levy, González, Perszyk, Hernandez, Chen, Ippolito, Gilboa, Dohan, Drakard, Jurgens, Datta, Ganguli, Emelin, Kleyko, Yuret, Chen, Tam, Hupkes, Misra, Buzan, Mollo, Yang, Lee, Shutova, Cubuk, Segal,
  Hagerman, Barnes, Donoway, Pavlick, Rodola, Lam, Chu, Tang, Erdem, Chang, Chi, Dyer, Jerzak, Kim, Manyasi, Zheltonozhskii, Xia, Siar, Martínez-Plumed, Happé, Chollet, Rong, Mishra, Winata, de~Melo, Kruszewski, Parascandolo, Mariani, Wang, Jaimovitch-López, Betz, Gur-Ari, Galijasevic, Kim, Rashkin, Hajishirzi, Mehta, Bogar, Shevlin, Schütze, Yakura, Zhang, Wong, Ng, Noble, Jumelet, Geissinger, Kernion, Hilton, Lee, Fisac, Simon, Koppel, Zheng, Zou, Kocoń, Thompson, Kaplan, Radom, Sohl-Dickstein, Phang, Wei, Yosinski, Novikova, Bosscher, Marsh, Kim, Taal, Engel, Alabi, Xu, Song, Tang, Waweru, Burden, Miller, Balis, Berant, Frohberg, Rozen, Hernandez-Orallo, Boudeman, Jones, Tenenbaum, Rule, Chua, Kanclerz, Livescu, Krauth, Gopalakrishnan, Ignatyeva, Markert, Dhole, Gimpel, Omondi, Mathewson, Chiafullo, Shkaruta, Shridhar, McDonell, Richardson, Reynolds, Gao, Zhang, Dugan, Qin, Contreras-Ochando, Morency, Moschella, Lam, Noble, Schmidt, He, Colón, Metz, Şenel, Bosma, Sap, ter Hoeve, Farooqi, Faruqui,
  Mazeika, Baturan, Marelli, Maru, Quintana, Tolkiehn, Giulianelli, Lewis, Potthast, Leavitt, Hagen, Schubert, Baitemirova, Arnaud, McElrath, Yee, Cohen, Gu, Ivanitskiy, Starritt, Strube, Swędrowski, Bevilacqua, Yasunaga, Kale, Cain, Xu, Suzgun, Tiwari, Bansal, Aminnaseri, Geva, Gheini, T, Peng, Chi, Lee, Krakover, Cameron, Roberts, Doiron, Nangia, Deckers, Muennighoff, Keskar, Iyer, Constant, Fiedel, Wen, Zhang, Agha, Elbaghdadi, Levy, Evans, Casares, Doshi, Fung, Liang, Vicol, Alipoormolabashi, Liao, Liang, Chang, Eckersley, Htut, Hwang, Miłkowski, Patil, Pezeshkpour, Oli, Mei, Lyu, Chen, Banjade, Rudolph, Gabriel, Habacker, Delgado, Millière, Garg, Barnes, Saurous, Arakawa, Raymaekers, Frank, Sikand, Novak, Sitelew, LeBras, Liu, Jacobs, Zhang, Salakhutdinov, Chi, Lee, Stovall, Teehan, Yang, Singh, Mohammad, Anand, Dillavou, Shleifer, Wiseman, Gruetter, Bowman, Schoenholz, Han, Kwatra, Rous, Ghazarian, Ghosh, Casey, Bischoff, Gehrmann, Schuster, Sadeghi, Hamdan, Zhou, Srivastava, Shi, Singh, Asaadi, Gu,
  Pachchigar, Toshniwal, Upadhyay, Shyamolima, Debnath, Shakeri, Thormeyer, Melzi, Reddy, Makini, Lee, Torene, Hatwar, Dehaene, Divic, Ermon, Biderman, Lin, Prasad, Piantadosi, Shieber, Misherghi, Kiritchenko, Mishra, Linzen, Schuster, Li, Yu, Ali, Hashimoto, Wu, Desbordes, Rothschild, Phan, Wang, Nkinyili, Schick, Kornev, Telleen-Lawton, Tunduny, Gerstenberg, Chang, Neeraj, Khot, Shultz, Shaham, Misra, Demberg, Nyamai, Raunak, Ramasesh, Prabhu, Padmakumar, Srikumar, Fedus, Saunders, Zhang, Vossen, Ren, Tong, Zhao, Wu, Shen, Yaghoobzadeh, Lakretz, Song, Bahri, Choi, Yang, Hao, Chen, Belinkov, Hou, Hou, Bai, Seid, Zhao, Wang, Wang, Wang, and Wu}]{srivastava2022imitation}
Aarohi Srivastava, Abhinav Rastogi, Abhishek Rao, Abu Awal~Md Shoeb, Abubakar Abid, Adam Fisch, Adam~R. Brown, Adam Santoro, Aditya Gupta, Adrià Garriga-Alonso, Agnieszka Kluska, Aitor Lewkowycz, Akshat Agarwal, Alethea Power, Alex Ray, Alex Warstadt, Alexander~W. Kocurek, Ali Safaya, Ali Tazarv, Alice Xiang, Alicia Parrish, Allen Nie, Aman Hussain, Amanda Askell, Amanda Dsouza, Ambrose Slone, Ameet Rahane, Anantharaman~S. Iyer, Anders Andreassen, Andrea Madotto, Andrea Santilli, Andreas Stuhlmüller, Andrew Dai, Andrew La, Andrew Lampinen, Andy Zou, Angela Jiang, Angelica Chen, Anh Vuong, Animesh Gupta, Anna Gottardi, Antonio Norelli, Anu Venkatesh, Arash Gholamidavoodi, Arfa Tabassum, Arul Menezes, Arun Kirubarajan, Asher Mullokandov, Ashish Sabharwal, Austin Herrick, Avia Efrat, Aykut Erdem, Ayla Karakaş, B.~Ryan Roberts, Bao~Sheng Loe, Barret Zoph, Bartłomiej Bojanowski, Batuhan Özyurt, Behnam Hedayatnia, Behnam Neyshabur, Benjamin Inden, Benno Stein, Berk Ekmekci, Bill~Yuchen Lin, Blake Howald,
  Cameron Diao, Cameron Dour, Catherine Stinson, Cedrick Argueta, César~Ferri Ramírez, Chandan Singh, Charles Rathkopf, Chenlin Meng, Chitta Baral, Chiyu Wu, Chris Callison-Burch, Chris Waites, Christian Voigt, Christopher~D. Manning, Christopher Potts, Cindy Ramirez, Clara~E. Rivera, Clemencia Siro, Colin Raffel, Courtney Ashcraft, Cristina Garbacea, Damien Sileo, Dan Garrette, Dan Hendrycks, Dan Kilman, Dan Roth, Daniel Freeman, Daniel Khashabi, Daniel Levy, Daniel~Moseguí González, Danielle Perszyk, Danny Hernandez, Danqi Chen, Daphne Ippolito, Dar Gilboa, David Dohan, David Drakard, David Jurgens, Debajyoti Datta, Deep Ganguli, Denis Emelin, Denis Kleyko, Deniz Yuret, Derek Chen, Derek Tam, Dieuwke Hupkes, Diganta Misra, Dilyar Buzan, Dimitri~Coelho Mollo, Diyi Yang, Dong-Ho Lee, Ekaterina Shutova, Ekin~Dogus Cubuk, Elad Segal, Eleanor Hagerman, Elizabeth Barnes, Elizabeth Donoway, Ellie Pavlick, Emanuele Rodola, Emma Lam, Eric Chu, Eric Tang, Erkut Erdem, Ernie Chang, Ethan~A. Chi, Ethan Dyer, Ethan
  Jerzak, Ethan Kim, Eunice~Engefu Manyasi, Evgenii Zheltonozhskii, Fanyue Xia, Fatemeh Siar, Fernando Martínez-Plumed, Francesca Happé, Francois Chollet, Frieda Rong, Gaurav Mishra, Genta~Indra Winata, Gerard de~Melo, Germán Kruszewski, Giambattista Parascandolo, Giorgio Mariani, Gloria Wang, Gonzalo Jaimovitch-López, Gregor Betz, Guy Gur-Ari, Hana Galijasevic, Hannah Kim, Hannah Rashkin, Hannaneh Hajishirzi, Harsh Mehta, Hayden Bogar, Henry Shevlin, Hinrich Schütze, Hiromu Yakura, Hongming Zhang, Hugh~Mee Wong, Ian Ng, Isaac Noble, Jaap Jumelet, Jack Geissinger, Jackson Kernion, Jacob Hilton, Jaehoon Lee, Jaime~Fernández Fisac, James~B. Simon, James Koppel, James Zheng, James Zou, Jan Kocoń, Jana Thompson, Jared Kaplan, Jarema Radom, Jascha Sohl-Dickstein, Jason Phang, Jason Wei, Jason Yosinski, Jekaterina Novikova, Jelle Bosscher, Jennifer Marsh, Jeremy Kim, Jeroen Taal, Jesse Engel, Jesujoba Alabi, Jiacheng Xu, Jiaming Song, Jillian Tang, Joan Waweru, John Burden, John Miller, John~U. Balis,
  Jonathan Berant, Jörg Frohberg, Jos Rozen, Jose Hernandez-Orallo, Joseph Boudeman, Joseph Jones, Joshua~B. Tenenbaum, Joshua~S. Rule, Joyce Chua, Kamil Kanclerz, Karen Livescu, Karl Krauth, Karthik Gopalakrishnan, Katerina Ignatyeva, Katja Markert, Kaustubh~D. Dhole, Kevin Gimpel, Kevin Omondi, Kory Mathewson, Kristen Chiafullo, Ksenia Shkaruta, Kumar Shridhar, Kyle McDonell, Kyle Richardson, Laria Reynolds, Leo Gao, Li~Zhang, Liam Dugan, Lianhui Qin, Lidia Contreras-Ochando, Louis-Philippe Morency, Luca Moschella, Lucas Lam, Lucy Noble, Ludwig Schmidt, Luheng He, Luis~Oliveros Colón, Luke Metz, Lütfi~Kerem Şenel, Maarten Bosma, Maarten Sap, Maartje ter Hoeve, Maheen Farooqi, Manaal Faruqui, Mantas Mazeika, Marco Baturan, Marco Marelli, Marco Maru, Maria Jose~Ramírez Quintana, Marie Tolkiehn, Mario Giulianelli, Martha Lewis, Martin Potthast, Matthew~L. Leavitt, Matthias Hagen, Mátyás Schubert, Medina~Orduna Baitemirova, Melody Arnaud, Melvin McElrath, Michael~A. Yee, Michael Cohen, Michael Gu,
  Michael Ivanitskiy, Michael Starritt, Michael Strube, Michał Swędrowski, Michele Bevilacqua, Michihiro Yasunaga, Mihir Kale, Mike Cain, Mimee Xu, Mirac Suzgun, Mo~Tiwari, Mohit Bansal, Moin Aminnaseri, Mor Geva, Mozhdeh Gheini, Mukund~Varma T, Nanyun Peng, Nathan Chi, Nayeon Lee, Neta Gur-Ari Krakover, Nicholas Cameron, Nicholas Roberts, Nick Doiron, Nikita Nangia, Niklas Deckers, Niklas Muennighoff, Nitish~Shirish Keskar, Niveditha~S. Iyer, Noah Constant, Noah Fiedel, Nuan Wen, Oliver Zhang, Omar Agha, Omar Elbaghdadi, Omer Levy, Owain Evans, Pablo Antonio~Moreno Casares, Parth Doshi, Pascale Fung, Paul~Pu Liang, Paul Vicol, Pegah Alipoormolabashi, Peiyuan Liao, Percy Liang, Peter Chang, Peter Eckersley, Phu~Mon Htut, Pinyu Hwang, Piotr Miłkowski, Piyush Patil, Pouya Pezeshkpour, Priti Oli, Qiaozhu Mei, Qing Lyu, Qinlang Chen, Rabin Banjade, Rachel~Etta Rudolph, Raefer Gabriel, Rahel Habacker, Ramón~Risco Delgado, Raphaël Millière, Rhythm Garg, Richard Barnes, Rif~A. Saurous, Riku Arakawa, Robbe
  Raymaekers, Robert Frank, Rohan Sikand, Roman Novak, Roman Sitelew, Ronan LeBras, Rosanne Liu, Rowan Jacobs, Rui Zhang, Ruslan Salakhutdinov, Ryan Chi, Ryan Lee, Ryan Stovall, Ryan Teehan, Rylan Yang, Sahib Singh, Saif~M. Mohammad, Sajant Anand, Sam Dillavou, Sam Shleifer, Sam Wiseman, Samuel Gruetter, Samuel~R. Bowman, Samuel~S. Schoenholz, Sanghyun Han, Sanjeev Kwatra, Sarah~A. Rous, Sarik Ghazarian, Sayan Ghosh, Sean Casey, Sebastian Bischoff, Sebastian Gehrmann, Sebastian Schuster, Sepideh Sadeghi, Shadi Hamdan, Sharon Zhou, Shashank Srivastava, Sherry Shi, Shikhar Singh, Shima Asaadi, Shixiang~Shane Gu, Shubh Pachchigar, Shubham Toshniwal, Shyam Upadhyay, Shyamolima, Debnath, Siamak Shakeri, Simon Thormeyer, Simone Melzi, Siva Reddy, Sneha~Priscilla Makini, Soo-Hwan Lee, Spencer Torene, Sriharsha Hatwar, Stanislas Dehaene, Stefan Divic, Stefano Ermon, Stella Biderman, Stephanie Lin, Stephen Prasad, Steven~T. Piantadosi, Stuart~M. Shieber, Summer Misherghi, Svetlana Kiritchenko, Swaroop Mishra, Tal
  Linzen, Tal Schuster, Tao Li, Tao Yu, Tariq Ali, Tatsu Hashimoto, Te-Lin Wu, Théo Desbordes, Theodore Rothschild, Thomas Phan, Tianle Wang, Tiberius Nkinyili, Timo Schick, Timofei Kornev, Timothy Telleen-Lawton, Titus Tunduny, Tobias Gerstenberg, Trenton Chang, Trishala Neeraj, Tushar Khot, Tyler Shultz, Uri Shaham, Vedant Misra, Vera Demberg, Victoria Nyamai, Vikas Raunak, Vinay Ramasesh, Vinay~Uday Prabhu, Vishakh Padmakumar, Vivek Srikumar, William Fedus, William Saunders, William Zhang, Wout Vossen, Xiang Ren, Xiaoyu Tong, Xinran Zhao, Xinyi Wu, Xudong Shen, Yadollah Yaghoobzadeh, Yair Lakretz, Yangqiu Song, Yasaman Bahri, Yejin Choi, Yichi Yang, Yiding Hao, Yifu Chen, Yonatan Belinkov, Yu~Hou, Yufang Hou, Yuntao Bai, Zachary Seid, Zhuoye Zhao, Zijian Wang, Zijie~J. Wang, Zirui Wang, and Ziyi Wu. 2022.
\newblock \href {http://arxiv.org/abs/2206.04615} {Beyond the imitation game: Quantifying and extrapolating the capabilities of language models}.

\bibitem[{Storks and Chai(2021)}]{storks-chai-2021-beyond-tip}
Shane Storks and Joyce Chai. 2021.
\newblock \href {https://doi.org/10.18653/v1/2021.findings-emnlp.272} {Beyond the tip of the iceberg: Assessing coherence of text classifiers}.
\newblock In \emph{Findings of the Association for Computational Linguistics: EMNLP 2021}, pages 3169--3177, Punta Cana, Dominican Republic. Association for Computational Linguistics.

\bibitem[{Storks et~al.(2019)Storks, Gao, and Chai}]{Storks2019RecentAI}
Shane Storks, Qiaozi Gao, and Joyce~Yue Chai. 2019.
\newblock Recent advances in natural language inference: A survey of benchmarks, resources, and approaches.
\newblock \emph{arXiv: Computation and Language}.

\bibitem[{Storks et~al.(2021)Storks, Gao, Zhang, and Chai}]{storks-etal-2021-tiered-reasoning}
Shane Storks, Qiaozi Gao, Yichi Zhang, and Joyce Chai. 2021.
\newblock \href {https://doi.org/10.18653/v1/2021.findings-emnlp.422} {Tiered reasoning for intuitive physics: Toward verifiable commonsense language understanding}.
\newblock In \emph{Findings of the Association for Computational Linguistics: EMNLP 2021}, pages 4902--4918, Punta Cana, Dominican Republic. Association for Computational Linguistics.

\bibitem[{Talmor et~al.(2019)Talmor, Herzig, Lourie, and Berant}]{talmor-etal-2019-commonsenseqa}
Alon Talmor, Jonathan Herzig, Nicholas Lourie, and Jonathan Berant. 2019.
\newblock \href {https://doi.org/10.18653/v1/N19-1421} {{C}ommonsense{QA}: A question answering challenge targeting commonsense knowledge}.
\newblock In \emph{Proceedings of the 2019 Conference of the North {A}merican Chapter of the Association for Computational Linguistics: Human Language Technologies, Volume 1 (Long and Short Papers)}, pages 4149--4158, Minneapolis, Minnesota. Association for Computational Linguistics.

\bibitem[{Tamari et~al.(2022)Tamari, Richardson, Kahlon, Sar-shalom, Liu, Tsarfaty, and Shahaf}]{tamari-etal-2022-dyna}
Ronen Tamari, Kyle Richardson, Noam Kahlon, Aviad Sar-shalom, Nelson~F. Liu, Reut Tsarfaty, and Dafna Shahaf. 2022.
\newblock \href {https://doi.org/10.18653/v1/2022.starsem-1.9} {{D}yna-b{A}b{I}: unlocking b{A}b{I}{'}s potential with dynamic synthetic benchmarking}.
\newblock In \emph{Proceedings of the 11th Joint Conference on Lexical and Computational Semantics}, pages 101--122, Seattle, Washington. Association for Computational Linguistics.

\bibitem[{Taori et~al.(2023)Taori, Gulrajani, Zhang, Dubois, Li, Guestrin, Liang, and Hashimoto}]{alpaca}
Rohan Taori, Ishaan Gulrajani, Tianyi Zhang, Yann Dubois, Xuechen Li, Carlos Guestrin, Percy Liang, and Tatsunori~B. Hashimoto. 2023.
\newblock Stanford alpaca: An instruction-following llama model.
\newblock \url{https://github.com/tatsu-lab/stanford_alpaca}.

\bibitem[{Thrush et~al.(2022)Thrush, Jiang, Bartolo, Singh, Williams, Kiela, and Ross}]{Thrush2022WinogroundPV}
Tristan Thrush, Ryan Jiang, Max Bartolo, Amanpreet Singh, Adina Williams, Douwe Kiela, and Candace Ross. 2022.
\newblock Winoground: Probing vision and language models for visio-linguistic compositionality.
\newblock \emph{2022 IEEE/CVF Conference on Computer Vision and Pattern Recognition (CVPR)}, pages 5228--5238.

\bibitem[{Touvron et~al.(2023)Touvron, Lavril, Izacard, Martinet, Lachaux, Lacroix, Rozière, Goyal, Hambro, Azhar, Rodriguez, Joulin, Grave, and Lample}]{touvron2023llama}
Hugo Touvron, Thibaut Lavril, Gautier Izacard, Xavier Martinet, Marie-Anne Lachaux, Timothée Lacroix, Baptiste Rozière, Naman Goyal, Eric Hambro, Faisal Azhar, Aurelien Rodriguez, Armand Joulin, Edouard Grave, and Guillaume Lample. 2023.
\newblock \href {http://arxiv.org/abs/2302.13971} {Llama: Open and efficient foundation language models}.

\bibitem[{Wang et~al.(2019{\natexlab{a}})Wang, Pruksachatkun, Nangia, Singh, Michael, Hill, Levy, and Bowman}]{Wang2019SuperGLUEAS}
Alex Wang, Yada Pruksachatkun, Nikita Nangia, Amanpreet Singh, Julian Michael, Felix Hill, Omer Levy, and Samuel~R. Bowman. 2019{\natexlab{a}}.
\newblock Superglue: A stickier benchmark for general-purpose language understanding systems.
\newblock In \emph{Neural Information Processing Systems}.

\bibitem[{Wang et~al.(2019{\natexlab{b}})Wang, Liang, Zhang, Li, and Gao}]{wang-etal-2019-make}
Cunxiang Wang, Shuailong Liang, Yue Zhang, Xiaonan Li, and Tian Gao. 2019{\natexlab{b}}.
\newblock \href {https://doi.org/10.18653/v1/P19-1393} {Does it make sense? and why? a pilot study for sense making and explanation}.
\newblock In \emph{Proceedings of the 57th Annual Meeting of the Association for Computational Linguistics}, pages 4020--4026, Florence, Italy. Association for Computational Linguistics.

\bibitem[{Wang et~al.(2023)Wang, Wei, Schuurmans, Le, Chi, Narang, Chowdhery, and Zhou}]{wang2023selfconsistency}
Xuezhi Wang, Jason Wei, Dale Schuurmans, Quoc~V Le, Ed~H. Chi, Sharan Narang, Aakanksha Chowdhery, and Denny Zhou. 2023.
\newblock \href {https://openreview.net/forum?id=1PL1NIMMrw} {Self-consistency improves chain of thought reasoning in language models}.
\newblock In \emph{The Eleventh International Conference on Learning Representations}.

\bibitem[{Winograd(1974)}]{Winograd1974UnderstandingNL}
Terry Winograd. 1974.
\newblock Understanding natural language.

\bibitem[{Xue et~al.(2020)Xue, Constant, Roberts, Kale, Al-Rfou, Siddhant, Barua, and Raffel}]{xue2020mt5}
Linting Xue, Noah Constant, Adam Roberts, Mihir Kale, Rami Al-Rfou, Aditya Siddhant, Aditya Barua, and Colin Raffel. 2020.
\newblock mt5: A massively multilingual pre-trained text-to-text transformer.
\newblock \emph{arXiv preprint arXiv:2010.11934}.

\bibitem[{Zellers et~al.(2018)Zellers, Bisk, Schwartz, and Choi}]{Zellers2018SWAGAL}
Rowan Zellers, Yonatan Bisk, Roy Schwartz, and Yejin Choi. 2018.
\newblock Swag: A large-scale adversarial dataset for grounded commonsense inference.
\newblock In \emph{Conference on Empirical Methods in Natural Language Processing}.

\bibitem[{Zellers et~al.(2019)Zellers, Holtzman, Bisk, Farhadi, and Choi}]{zellers-etal-2019-hellaswag}
Rowan Zellers, Ari Holtzman, Yonatan Bisk, Ali Farhadi, and Yejin Choi. 2019.
\newblock \href {https://doi.org/10.18653/v1/P19-1472} {{H}ella{S}wag: Can a machine really finish your sentence?}
\newblock In \emph{Proceedings of the 57th Annual Meeting of the Association for Computational Linguistics}, pages 4791--4800, Florence, Italy. Association for Computational Linguistics.

\bibitem[{Zhang et~al.(2018)Zhang, Liu, Liu, Gao, Duh, and Durme}]{Zhang2018ReCoRDBT}
Sheng Zhang, Xiaodong Liu, Jingjing Liu, Jianfeng Gao, Kevin Duh, and Benjamin~Van Durme. 2018.
\newblock Record: Bridging the gap between human and machine commonsense reading comprehension.
\newblock \emph{ArXiv}, abs/1810.12885.

\bibitem[{Zhou et~al.(2019)Zhou, Khashabi, Ning, and Roth}]{zhou-etal-2019-going}
Ben Zhou, Daniel Khashabi, Qiang Ning, and Dan Roth. 2019.
\newblock \href {https://doi.org/10.18653/v1/D19-1332} {{``}going on a vacation{''} takes longer than {``}going for a walk{''}: A study of temporal commonsense understanding}.
\newblock In \emph{Proceedings of the 2019 Conference on Empirical Methods in Natural Language Processing and the 9th International Joint Conference on Natural Language Processing (EMNLP-IJCNLP)}, pages 3363--3369, Hong Kong, China. Association for Computational Linguistics.

\bibitem[{Zhou et~al.(2021)Zhou, Gopalakrishnan, Hedayatnia, Kim, Pujara, Ren, Liu, and Hakkani-Tur}]{zhou-etal-2021-commonsense}
Pei Zhou, Karthik Gopalakrishnan, Behnam Hedayatnia, Seokhwan Kim, Jay Pujara, Xiang Ren, Yang Liu, and Dilek Hakkani-Tur. 2021.
\newblock \href {https://aclanthology.org/2021.sigdial-1.13} {Commonsense-focused dialogues for response generation: An empirical study}.
\newblock In \emph{Proceedings of the 22nd Annual Meeting of the Special Interest Group on Discourse and Dialogue}, pages 121--132, Singapore and Online. Association for Computational Linguistics.

\end{thebibliography}
